\documentclass[12pt,a4paper,table]{report}
\usepackage{graphicx}
\usepackage{mathptmx}
\usepackage{amssymb}
\usepackage{amsmath}
\usepackage{glossaries}
\usepackage{multirow}
\usepackage{CJKutf8} 
\usepackage[table,xcdraw]{xcolor}
\usepackage{caption}
\usepackage{subcaption}
\usepackage{multicol}
\usepackage{float}
\usepackage{array}

\usepackage{hyperref}

\newcommand{\etal}{\textit{et al}. }
\newcommand{\ie}{\textit{i}.\textit{e}., }
\newcommand{\eg}{\textit{e}.\textit{g}. }

\usepackage{glossaries}
\makeglossaries

\newglossaryentry{BOW}
{
    name=BOW,
    description={Bag-of-words}
}

\newglossaryentry{COLIEE}
{
    name=COLIEE,
    description={Competition on Legal Information Extraction/Entailment}
}

\newglossaryentry{CBOW}
{
    name=CBOW,
    description={Continous Bag-of-words}
}

\newglossaryentry{CNN}
{
    name=CNN,
    description={Convolutional Neural Network}
}

\newglossaryentry{deep legal}
{
    name=Deep legal,
    description={Deep learning in processing legal text}
}

\newglossaryentry{FC}
{
    name=FC,
    description={Fully Connected}
}

\newglossaryentry{MLP}
{
    name=MLP,
    description={Multilayer Perceptrons}
}

\newglossaryentry{MLM}
{
    name=MLM,
    description={Masked Language Model}
}

\newglossaryentry{nlp}
{
    name=NLP,
    description={Natural Language Processing}
}

\newglossaryentry{NSP}
{
    name=NSP,
    description={Next Sentence Prediction}
}

\newglossaryentry{NFSP}
{
    name=NFSP,
    description={Next Foreign Sentence Prediction}
}

\newglossaryentry{NMSP}
{
    name=NMSP,
    description={Neighbor Multilingual Sentence Prediction}
}

\newglossaryentry{NLI}
{
    name=NLI,
    description={Natural Language Inference}
}

\newglossaryentry{OOV}
{
    name=OOV,
    description={Out of Vocabulary}
}

\newglossaryentry{TPU}
{
    name=TPU,
    description={Tensor Processing Unit}
}

\newglossaryentry{TRE}
{
    name=TRE,
    description={Transferred-Requisite-Effective}
}

\newglossaryentry{TF-IDF}
{
    name=TF-IDF,
    description={Term Frequency — Inverse Document Frequency}
}

\newglossaryentry{TE}
{
    name=TE,
    description={Textual Entailment}
}

\begin{document}
\thispagestyle{empty}
\begin{center}
Doctoral Dissertation\\% Erase either one of them
\vfill
Toward Improving Attentive Neural Networks in Legal Text Processing\\
\vfill
Nguyen Ha Thanh\\
\vfill
Supervisor:  Nguyen Le Minh\\
\vfill
Graduate School of Advanced Science and Technology\\ % Or, School of Information Science
Japan Advanced Institute of Science and Technology\\
% (ZZZZZ)\\ % intended degree
\vfill
March, 2022\\ % conferment month/year
\vfill
\end{center}

\clearpage
\chapter*{Acknowledgements}
\thispagestyle{empty}%no page number
I hereby declare that this dissertation is written by myself, from the observations and experiments I made while studying at JAIST as a doctoral student.
Even so, to obtain the results described in this dissertation, I am indebted to many people.

First and foremost, I would like to express my best gratitude to Professor Nguyen Le Minh, who has imparted to me many valuable research experiences. He made me broaden my mind and believe in the possibility of making the impossible possible.
He is the inspiration for me to strive in my research career.

I would like to thank Professor Satoshi Tojo, Associate Professor Kiyoaki Shirai, Professor Ken Satoh, and Professor Shinobu Hasegawa for their guidance and constructive suggestions to improve my research.

I also want to express gratitude to Associate Professor Nguyen Viet Ha, Professor Nguyen Dinh Duc, Professor Nguyen Thanh Thuy, and other teachers at VNU-UET and VNU-LS, who have inspired and built me up a strong research background before studying at JAIST.

I would like to thank my co-authors, colleagues and fraternities for the opportunity to work with them. In addition, I also highly appreciate authors whose works are cited in this dissertation. This dissertation would never have been done without the foundations they created.

I am grateful to Ministry of Education, Culture, Sports, Science and
Technology (MEXT) and other research funds for supporting the necessary expenses for my research activities in Japan. I would also like to thank JAIST for providing me with excellent living and research facilities.

Finally and most sincerely, I would like to thank my parents, family and friends, who have always supported and given me great motivation to overcome difficult challenges to obtain initial results in scientific research.

\begin{flushright}
\textit{Nguyen Ha Thanh}

\textit{Japan, 2022}
\end{flushright}

\clearpage
\begin{abstract}
In recent years, thanks to breakthroughs in neural network techniques especially attentive deep learning models, natural language processing has made many impressive achievements.
However, automated legal word processing is still a difficult branch of natural language processing.
Legal sentences are often long and contain complicated legal terminologies.
Hence, models that work well on general documents still face challenges in dealing with legal documents. 
We have verified the existence of this problem with our experiments in this work.
In this dissertation, we selectively present the main achievements in improving attentive neural networks in automatic legal document processing.
Language models tend to grow larger and larger, though, without expert knowledge, these models can still fail in domain adaptation, especially for specialized fields like law.

This dissertation has three main tasks to achieve the goal of improving attentive models in legal document processing.
First, we survey and verify the factors affecting the performance of the models when operating on a specific domain such as law.
This investigation is to provide clearer insights to improve models in this domain.
Second, as pretrained language models are recently the most well-known attentive approaches in natural language processing, we provide methods to create language models specific to the legal domain, producing state-of-the-art results on reliable datasets.
These models are built on features from the data of legal documents, with the goal of overcoming the challenges found in our previous survey.
Third, besides the approach to let the model learn completely from raw data, we propose and prove the effectiveness of using different knowledge sources to inject into the model in different ways to adjust their output.
This approach not only increases explainability but also allows humans to control pretrained language models and take advantage of the knowledge resources available during the development of the field such as vocabulary, grammar, logic and law.

\vspace{1cm}
\textbf{\textit{Keywords:}} Legal Text Processing, Attentive Neural Network, Deep Legal, Pretrained Language Model, Knowledge Injection.
\end{abstract}

\tableofcontents\thispagestyle{empty}
\listoffigures\thispagestyle{empty}
\listoftables\thispagestyle{empty}
\printglossaries\thispagestyle{empty}

\setcounter{page}{0}
\chapter{Introduction}
\section{Introduction}
Automated processing of legal documents is an urgent need in today's information society. 
Besides the convenience of social media, our actions on these platforms may involve or result in many legal effects.
Legal questions about freedom of speech raised around Twitter\footnote{https://twitter.com}'s banning of former American President Donald Trump on their platform \cite{roberts2019trump} 
or Tesla\footnote{https://www.tesla.com} has to hire employees to control the legal risk of their chairman Elon Musk's statements are good examples attesting to this phenomenon.
However, for social and technical reasons, the quality of automatic law processing systems has not yet met the needs of society.

In terms of social reasons, computer science has made significant results only in recent years, while the law is a field that has been attached to people for centuries since the formation of countries. 
The law exists parallel to the development of mankind and for a long time, without any requirement of technology. 
Besides, both law and computer science are specialized academic disciplines that do not have much in common. 
Therefore, it may take a long time to get breakthroughs in the application of computer science to law.

For technical reasons, the sentences are often long and have a complex semantic structure. 
It is even difficult for a human to understand the exact meaning of a legal sentence on the first reading.
There must be an interpreting role of the Court in the common law system in countries such as the United Kingdom, the United States of America, Canada; and guiding documents in the civil law system like in Germany, Japan, Vietnam.
Besides, legal documents are written in natural language, a means of communication that is not designed for correctness. 
The ambiguity in the natural language could be an obstacle for any intelligent system, even for human beings. 
Especially in languages with multi-layered meanings (such as Chinese, Japanese, Vietnamese), understanding the exact meaning through sentences is a more difficult problem.
In addition, the vocabulary used in the legal domain does not completely coincide with the words that people use to communicate every day. 
Therefore, it can be considered as a special sublanguage in our language.

Along with the growth of hardware computation capacity, deep learning and especially attentive models have proven their power in many different tasks in  natural language processing. 
Delicate tasks such as speech recognition, question answering, and language generation are all well performed by systems using this approach. 
Given such achievements, we can expect the possibility of using deep learning models in dealing with more complex linguistic tasks in the legal domain. 
In this dissertation, we selectively report our research results in improving the performance and explainability of deep learning especially attentive models  in processing legal text (we use the term \gls{deep legal} processing in short) . Since legal language is different from daily language, we need an appropriate approach for this kind of data. Besides the enhancement in performance, the dissertation also provides the informative characteristics of deep legal processing for the readers.

Transfer learning and pretrained attentive models are robust approaches in domain adaptation.
However, in a specialized domain like law, without an understanding of the domain and the data, it is difficult for these models to yield good results. 
Therefore, a detailed investigation of the possibilities and methods of applying deep learning to legal text processing is useful information for the development of automation in this field. 
The three main research questions would be answered in this dissertation include:

\begin{enumerate}
    \item What factors impact the performance of end-to-end deep learning models trained with mere provisioning data perform legal document processing tasks? 
    \item Pretrained language models have become one of the powerful approaches in deep learning. What characteristics in the legal text can be used to implement successful instances of these models?
    \item How to make use of available knowledge sources to inject into the deep learning models to have a better performance? Which kinds of knowledge are available?
\end{enumerate}

To answer these questions, we make hypotheses and test them in specific problems. For each problem, we propose methods, conduct experiments, observe, analyze experimental results and draw conclusions.

% This is not a work that represents an attempt to replace lawyers by machines. 
% This replacement is theoretically impossible because the law is a tool to protect people, built on people's views on morality and justice. 
% Therefore, no matter how powerful, machine systems are only a supporting factor for humans to make final decisions. 
% Even so, this work attempts to demonstrate that, with the aid of machines, human decisions can become accurate, reduce bias, and be aware of biases and prejudices that exist in their legal thoughts and decisions.

\section{Motivations}

\subsection{Factor Analysis for Deep Legal Systems}

The first motivation of this study is to understand the factors that influence a deep legal system and to propose appropriate improvements based on these understandings.
% Our first research direction is feature analysis for deep legal.
Conducting the works introduced in this dissertation, we focus on improving both the performance and the apparentness of deep legal models.
Deep learning models are often considered black boxes, as long as there is enough data, they will achieve the desired effect.
Even so, the assumption of enough data is hard to be satisfied in all areas of daily life.
% On the contrary, data is never enough for different human needs.
Therefore, analyzing the characteristics of deep legal helps us to use data more effectively.
% Our second research direction is performance of deep legal, 
This dissertation also conveys information about what tasks and under what conditions can deep learning models perform well in the legal domain.
This work can also be seen as an effort to increase the explainability of deep legal models, which is crucial to bring these models to real-life applications.

Understanding the factors that can affect systems in a domain is an important requirement for good designs.
Data characteristics in the legal domain are fragmented data, long legal sentences, and many specialized terms.
Therefore, we choose to investigate in detail factors such as the amount of data, the way the data is represented, and the architecture of the model working on the data.
For the data amount factor, we experiment on a problem with limited data, propose solutions to increase the data and compare the results in the new setting.
To understand the impact of data representation, we propose a method to evaluate different embedding methods in both the general and legal domains.
About the model architecture, we compare the performance of different architectures on the same problem. 
The experimental results show an interesting superiority of an attentive \gls{CNN} network compared with a pre-trained cumbersome language model with the vanilla architecture.

\subsection{Pretrained Language Models for Deep Legal Processing}
Our second motivation is to verify the ability of the pretrained language model in the legal domain.
In recent years, pretrained language models have gained popularity and made many breakthroughs in various problems in natural language processing. 
Following this trend, we design pretrained language models for deep legal tasks. 
Besides performance, an important factor to evaluate models, we focus on philosophy when designing them. 
The models introduced are the result of observations drawn from the factors affecting deep legal models obtained from our investigation.
Pretrained language models often contain biases that exist in the training data, so often perform poorly on a very distinct domain.
Fortunately, for the legal field, we can take advantage of the data properties of this domain to train or adjust the weights of these models.

From observing the importance of data representation in the legal field, we propose a pretrained language model named BERTLaw, which is trained from scratch using a large amount of legal data.
Besides achieving good results in our experiment, this model also helps us confirm the importance of data representation. 
Having a good data representation is a prerequisite for a strong deep legal system.
In addition to BERTLaw, we introduce Paralaw and Paraformer, models based on pretrained language models that overcome the issues of data amount and model architecture limitations.

\subsection{Knowledge Injection for Deep Legal Models}
Our third motivation is to perform and leverage legal and linguistic knowledge sources to improve the performance of deep legal models.
Deep learning models can learn from data and demonstrate their effectiveness on a wide range of tasks.
However, relying solely on data has three disadvantages.
First, the quality of the model is based on the quality of the data, or in other words garbage in garbage out. This can be dangerous when lay users are too dependent on data.
Second, humans will be less likely to participate in the decision-making process. This can lead to the abuse of power by intelligent systems.
Third, these systems are considered black boxes and debugging them is extremely difficult.
Therefore, we investigate and propose approaches to inject knowledge into deep learning models to guide the learning and generation processes  of these models.

For linguistic knowledge, we introduce HYDRA, an architecture that allows to train the transformer model's attention heads separately and then graft them onto the original body.
This approach leads to cost-effectiveness in training as well as storage.
For legal knowledge, we experiment with knowledge of logical parts of legal sentences. We use a special mechanism to inject this knowledge into the different layers of the transformer model.
Finally, with the model of language generation in the legal field, we propose a method that uses the knowledge of fairness to regulate the output of this system.
These findings are the basis for using other types of knowledge resources to improve future deep legal models.
\section{Contributions}

The dissertation contributes three main values: performance improvement, methodology, philosophy.
First, the systems proposed in this dissertation all have better performance than existing baselines.
Some of them achieve state-of-the-art results on reliable datasets.
Second, the performance improvements of the systems are all based on methods designed from observations of experimental results.
We not only explain the proposed methods in each chapter but also outline the journey to build them.
Third, the conclusions and discussions in the sections of this dissertation have philosophical value in the design of deep legal models.

\begin{figure}[ht]
  \centering
\includegraphics[width=.8\linewidth]{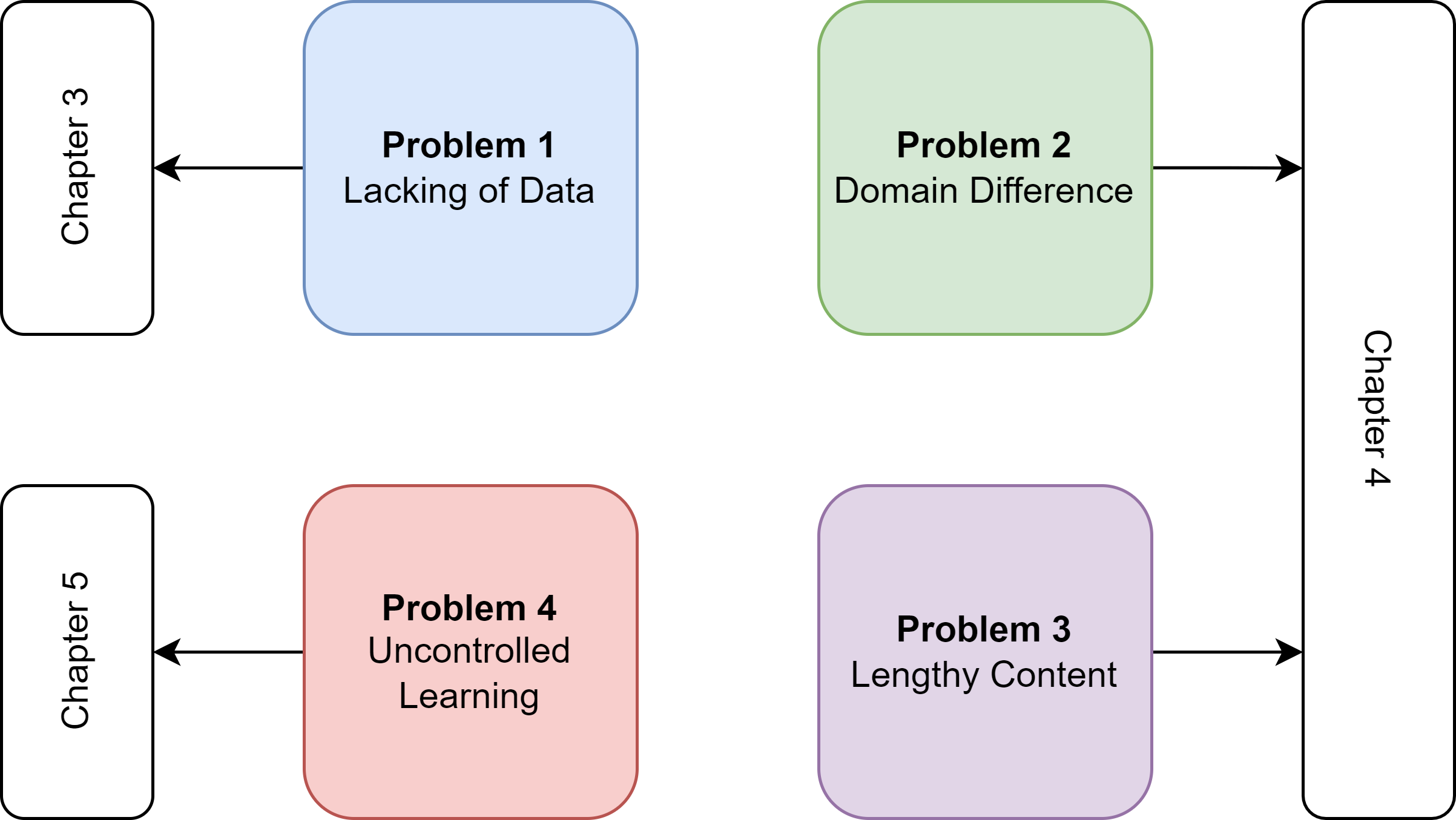}
  \caption{The main problems which are mentioned, analyzed and solved in the dissertation.}
  \label{fig:dissertation_problems}
\end{figure}

The main contribution of the dissertation includes discovery and settling 4 common problems for deep learning systems in the law domain, namely \textit{lacking of data}, \textit{domain difference}, \textit{lengthy content}, and \textit{uncontrolled learning} as demonstrated in Figure \ref{fig:dissertation_problems}.
 Besides the non-architecture solutions, the models proposed in this dissertation all take advantage of the attention mechanism.
 The dissertation also shows that, without appropriate approaches, the power of attentive models can be wasted. This is especially demonstrated in the sections on Attentive CNN, pretrained language models, and Paraformer.

To this end, we provide qualitative and quantitative information about attentive neural networks in legal text processing. 
We propose different approaches to utilize the characteristics of legal text and supplementary knowledge to improve not only the performance of these models but also their explainability. 
Besides, we propose methods to customize the attentive architecture in neural networks for better designs. 
With the detailed explanation about different levels of intervention in the attentive network to inject expert knowledge, this dissertation could also be a good technical reference document for people who may concern.

As a collection of works toward improving attentive neural networks in legal text processing, the details about sub-contributions of this dissertation are listed in Table \ref{tab:contributions}.
The attentive models are investigated and proposed with our improvements from Section 3.4 to Section 5.4. 

\begin{table}
\small
\center
\begin{tabular}{|c|p{11cm}|}
\hline
\textbf{Section}         & \textbf{Contributions}\\ \hline\hline
3.2                      & \textit{LVC}, \textit{LECA} metrics, and a visualization method to understand about distribution of legal word vectors in the different legal embeddings. The measurement and visualization results suggest the problem of \textit{domain difference}s in pretrained language models.\\ \hline
3.3                     & \textit{Lawfulness classification}, a new way to consider the problem of textual-entailment-based question-answering for legal text, can boost the performance of the same model by increasing the data amount. The experimental results indicate the existence of the \textit{lacking of data} problem in deep learning approaches for some legal tasks in this domain.\\ \hline
3.4                     & \textit{Attentive CNN}, a simple convolutional neural network with attention mechanism. This model overperforms a robust pretrained language model (\textit{XLM-RoBERTa}), which suggests the existence of lengthy input problem in the legal domain. \\ \hline\hline
%--------------------------
4.2                     & \textit{BERTLaw}, a model with the vocabulary and the model's weights are constructed and trained from scratch using a large legal corpus. This model achieves the state-of-the-art result in COLIEE 2020's Task 4. \\ \hline
4.3                      & \textit{NFSP} and \textit{NMSP}, novel pretraining tasks based on the characteristic of translation alignment in legal data. Our model pretrained with \textit{NFSP} task achieves state-of-the-art result in COLIEE 2021's Task 5.\\ \hline
4.4                      & \textit{Paraformer}, a novel pretrained language model that could handle much longer input text than the base model. This characteristic is important to obtain good predictions for lengthy legal inputs.\\ \hline\hline
%---------------------------
5.2                     & \textit{HYDRA}, an architecture-friendly and extensible method to improve the effectiveness of the transformer-based language models by pretraining and appending new knowledge-guided heads to their architecture. \\ \hline
5.3                      & \textit{TRE} framework, a logic-structure knowledge injection approach for pretrained Transformer models. Our experiments indicate that a suitable strategy to inject legal logical knowledge can boost the performance of deep legal models.\\ \hline
5.4                     & \textit{BART2S}, a regulated generator-discriminator generative framework for generating terms of services automatically using the generative models.
We also propose a novel tuning process to adjust the fairness of the generated content based on the knowledge of fairness learned by the discriminator. \\ \hline
\end{tabular}
\caption{Sub-contributions of the dissertation.}\label{tab:contributions}
\end{table}

The impact of this research may contribute to both scientific and practical meaning. The dissertation provides the whole picture of deep learning in legal text processing and related aspect in its content. In addition, embedding methods, training tasks, and architectural designs, which are the most important factors of every deep learning model, will be presented in this dissertation. From a practical viewpoint, the outcome of this research may contribute to bringing the most advanced techniques in deep learning to the legal domain. This document can be useful for researchers who seek for explainability of the deep learning model in the legal domain but not only using it as a black box. Explainability is a prerequisite for the deep legal system to be approved to operate in the real life.

\section{Dissertation Outline}

This dissertation is conducted with the purpose of analyzing and improving current state-of-the-art techniques in legal document processing using deep learning models.
Firstly, we analyze different aspects of applying end-to-end deep learning models to a legal processing problem.
By doing so, we obtain a clear insight to design effective models for each particular condition.
Secondly, we propose novel ways to pretrain language models in the legal domain to improve their performance.
Thirdly, we design approaches in using expert knowledge to support the models to have better learning and prediction in the legal domain.

\begin{figure}[ht]
  \centering
\includegraphics[width=.9\linewidth]{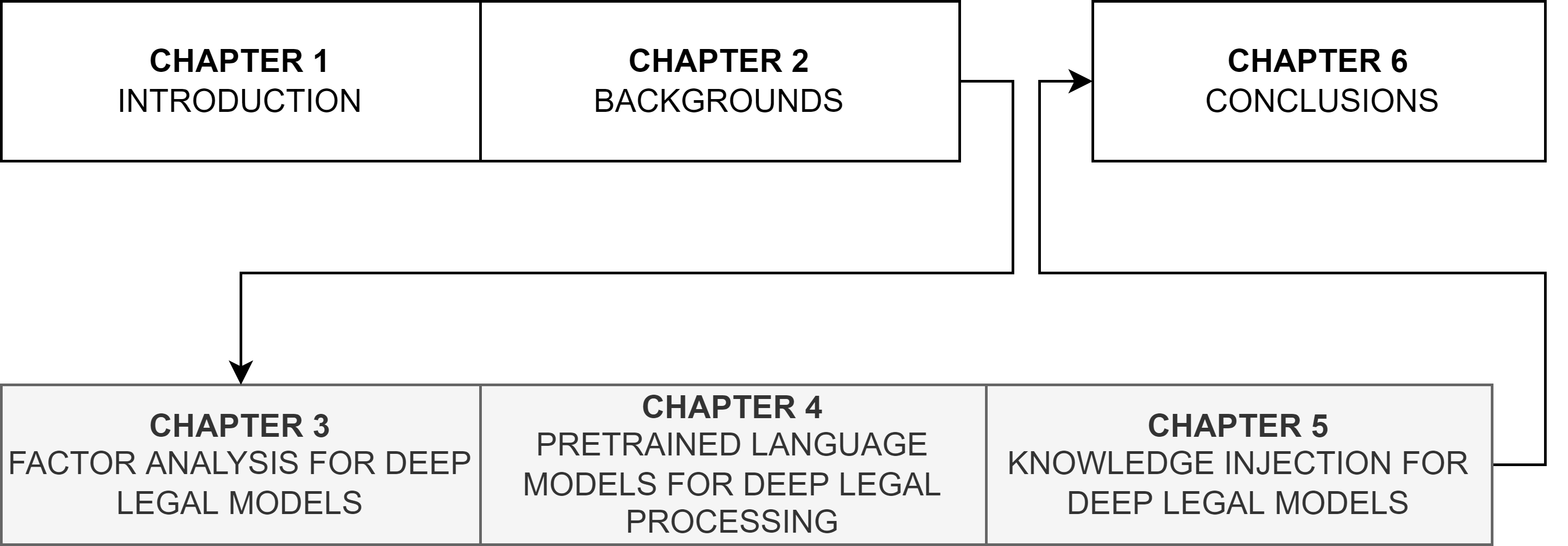}
  \caption{Outline of the dissertation.}
  \label{fig:dissertation_outline}
\end{figure}

The outline of the dissertation is presented in Figure \ref{fig:dissertation_outline}. Firstly, we want to confirm the ability of deep learning models in performing a legal task, which often requires expert knowledge. We analyze the impact of data representation, data amount and the architecture of deep learning models. This confirmation is the first step in exploring the knowledge about deep legal processing. After that, we discover further about which characteristics of legal data can be utilized to pretrain strong legal language models, a family of multi-head attentive networks achieving many good results in natural language processing recently. The legal embedding, legal multilingual capacity and legal structure representation are addressed when answering this question. Lastly, we investigate the possibility of injecting knowledge into the neural networks to gain the performance and explainability of the models in this domain. Linguistic knowledge, legal knowledge and self-learned knowledge are investigated to answer this question. 

Before answering the research questions, we dedicate Chapter 1 to introduce the research goals, the challenges as well as the motivations when we conduct this research and Chapter 2 to present the basic knowledge of deep learning, attention mechanism, and multi-head attentive models \cite{vaswani2017attention}. These technologies are highly influential when this dissertation was written. This knowledge not only provides the foundation for the reader to access the next chapters but also contributes to clarifying the context of the research. It is possible that these technologies will become obsolete and superseded in the future. However, the philosophy and the methodology of the dissertation still retain their reference values. Besides, we also present the characteristics of legal documents, the difference of legal documents from the daily text, challenges, and advantages when processing legal documents by deep learning.
	
The first research question is answered in Chapter 3. 
% In this chapter, we do not simply give a yes/no answer, but provide detailed information about this possibility. 
We examine in detail the factors that can affect deep learning models such as data representation, data amount, and model architecture. During our research on deep learning architectures, we found very simple architectures like SCNN \cite{nguyen2019swarm}, which have a small number of parameters that can still outperform other models. Interestingly, we also found the simple combination of CNN \cite{lecun1999object} architecture and attention mechanism \cite{kien2020answering} can give better results than bulky models in some specific cases. This chapter will answer the question of under what conditions can the end-to-end model perform well in a legal text processing task.

The next research question is answered in Chapter 4. In recent years, language models have been a powerful approach in deep learning. Pretrained with a huge amount of data, these models are capable of understanding the language and performing excellently on tasks in benchmark data. Models such as BERT \cite{devlin2018bert}, GPT-3 \cite{brown2020language}, and BART \cite{lewis2019bart} make for a breakthrough in NLP compared to traditional NLP methods. These models take advantage of the idea of transfer learning, learning one task can improve the results of another. Many studies show that combining and interleaving tasks can improve the efficiency of the model. In our research, we present novel ways to pretrain language models. 
In the legal domain, our proposed model like BERTLaw \cite{nguyen2020jnlp}, ParaLaw \cite{nguyen2021paralaw} proved their effectiveness in the \gls{COLIEE} 2020 and COLIEE 2021 competitions with the standard data set provided by the organizers. With an end-to-end model, if garbage in, then garbage out, so having an appropriate training method is important to build a quality deep learning model.

The final research question is answered in Chapter 5. Besides the traditional training and pretrain-finetune paradigms, there exists the third approach, knowledge injection \cite{nguyen2021knowledge}. This approach is the use of expert knowledge to support learning models and decision-making. Instead of feeding the model with data so that it learns relationships on its own, we can directly feed expert knowledge into the model in the form of signals. This method helps to solve the problem of sparse, noisy data and leverages expert knowledge in training deep learning models. This expert knowledge can be in the form of linguistic features or semantic features. Through our experiments, we demonstrate that injecting this expert knowledge into the neural network will increase the performance of the model. Besides, this approach also helps to increase the accountability and debuggability in deep learning models.

The ultimate goal of the dissertation is to present our work on the road to improving attentive neural networks in legal text processing. 
The contents in Chapters 3 and 4 are our results and observations from our COLIEE participation.
Chapter 5 presents preliminary investigations in an attempt to enhance the explainability of the attentive neural networks, which are considered black boxes.
Although this work is done with meticulousness, there may be blind spots in the experiments and bias in the interpretation of the results. 
Therefore, in each work, we do not only report the performance as a single number but have a deeper analysis of the experimental results. 
At the end of each chapter, we summarize the main points of that chapter and the related discussions.
Our final discussions and conclusions are presented in Chapter 6. 
This chapter enables the readers to understand our contributions as a coherent work towards improving attentive models in legal text processing.
Last but not least, we outline future directions that can widen the scope and elevate this research to real-world applications.

\chapter{Backgrounds}
\label{chap:background}
\section{Premiliary about Deep learning}

\subsection{Linear Neural Networks}
A few years ago, deep learning is a blue-sky discipline, not many researchers and engineers are interested in it because the hardware condition does not allow them to build anything meaningful with this approach.
However, recently, deep learning has become the most powerful and well-known technology in many different fields of AI, computer vision and natural language processing.
Deep learning is also the backbone technology in this dissertation containing multiple sub-studies in applying deep learning to legal text processing.
For that reason, we dedicate this section to the preliminary of deep learning.

\begin{figure}[ht]
  \centering
\includegraphics[width=.8\linewidth]{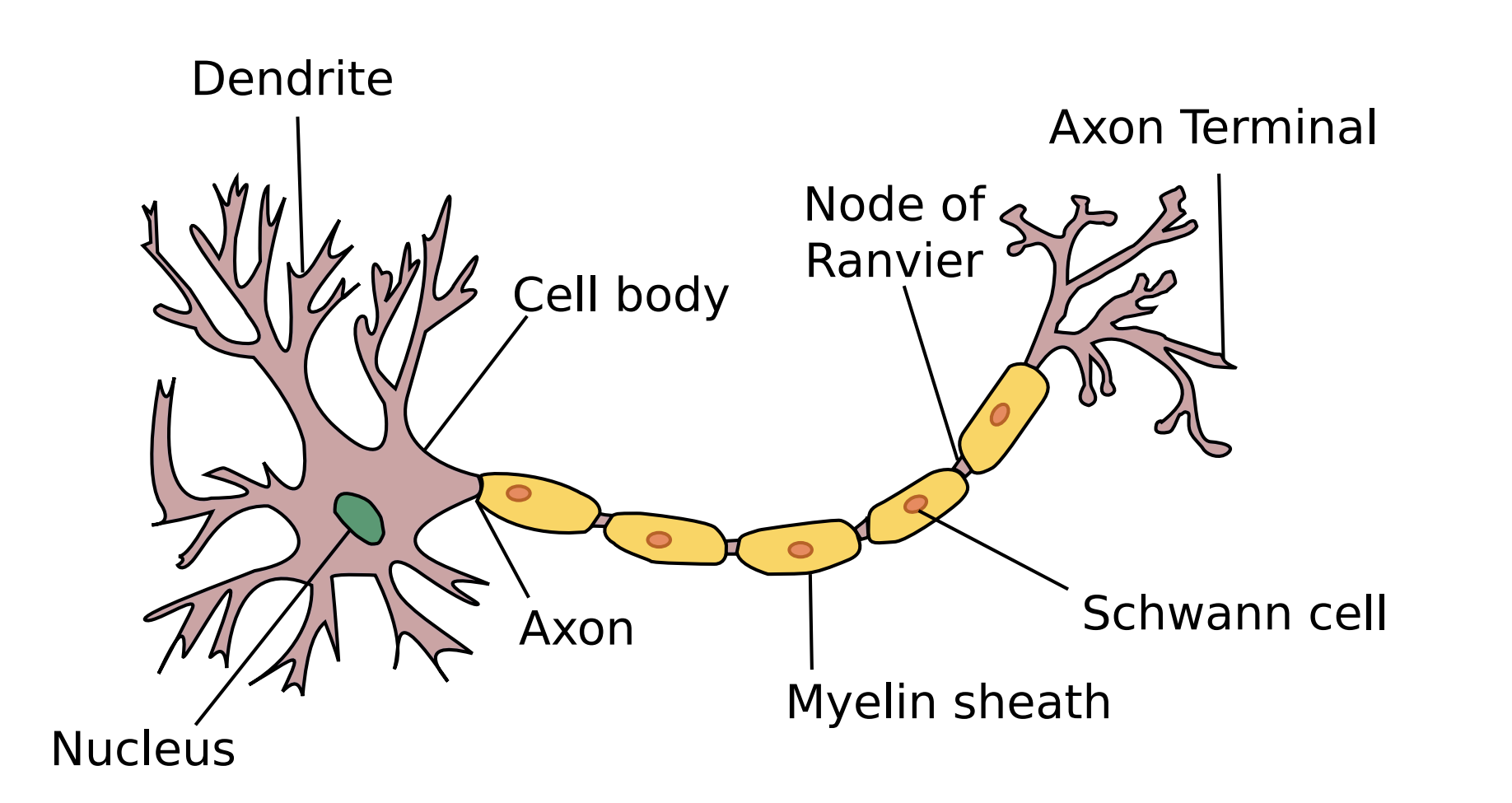}
  \caption{Structure of a real neuron.}
  \label{fig:real_neuron}
\end{figure}

\begin{figure}[ht]
  \centering
\includegraphics[width=.8\linewidth]{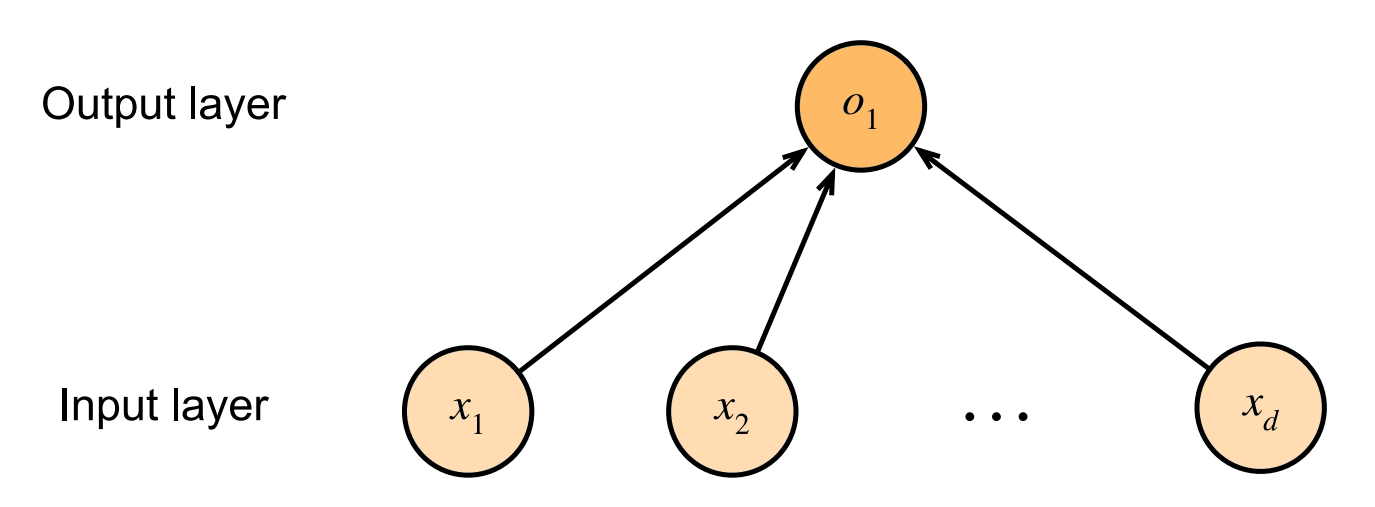}
  \caption{Demonstration of a linear regression model.}
  \label{fig:linear_regression}
\end{figure}

Deep learning is a technique that uses an artificial neural network, which is an architecture inspired by the real nervous system of animals.
Figure \ref{fig:real_neuron} demonstrates the structure of a real neuron and Figure \ref{fig:linear_regression} shows a linear regression model, which is the simplest imitation of this computational unit.
One single neuron receives signals from the others as inputs and computes its own signal.
Satisfying certain conditions, this signal is passed to another neuron.

With $x_1$, $x_2$, ..., $x_d$ be the input signal, $w_1$, $w_2$, ..., $w_d$, $b$ be the parameters of this model, the prediction $\hat{y}$ value can be calculated as Equation \ref{eq:linear_regression_pred}.

\begin{equation}
    \label{eq:linear_regression_pred}
    \hat{y}=w_{1} x_{1}+\ldots+w_{d} x_{d}+b
\end{equation}

In another representation, let $\mathbf{x}$ be the input vector, $\mathbf{w}$ be the parameter vector, we have:
\begin{equation}
    \label{eq:linear_regression_vector}
    \hat{y}=\mathbf{w}^{\top} \mathbf{x}+b
\end{equation}

Training the model in a dataset contain $n$ sample, we need to minimize the loss value given by the Equation \ref{eq:linear_regression_loss} with the optimal parameters given by the Equation \ref{eq:linear_regression_root}:

\begin{align}
    L(\mathbf{w}, b)&=\frac{1}{n} \sum_{i=1}^{n} \frac{1}{2}\left(\mathbf{w}^{\top} \mathbf{x}^{(i)}+b-y^{(i)}\right)^{2} \label{eq:linear_regression_loss} \\
    \mathbf{w}^{*}, b^{*}&=\underset{\mathbf{w}, b}{\operatorname{argmin}} L(\mathbf{w}, b) \label{eq:linear_regression_root}
\end{align}
    
As the simplest form of neuron imitation, linear regression can only predict points on a straight line.
For nonlinear functions, this model cannot be fit, so its application is limited.
To simulate the world, we need more complex functions than straight-line equations.

\subsection{Multilayer Perceptrons}
\begin{figure}[ht]
  \centering
\includegraphics[width=.8\linewidth]{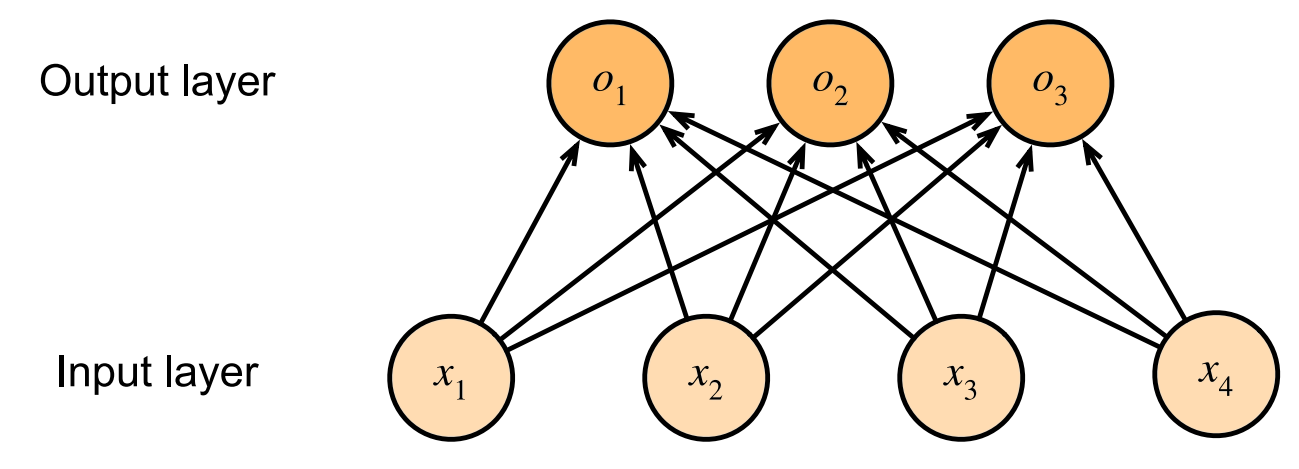}
  \caption{Demonstration of a softmax regression model.}
  \label{fig:softmax_regression}
\end{figure}

Apparently, the linear regression model can not perform well on classification problems. When we add more output nodes into the network, we obtain a softmax regression model. This model map a vector into a probability distribution, which can be used for a classification problem.
The Figure \ref{fig:softmax_regression} demonstrates a softmax regression network.
The prediction value $\mathbf{\hat{y}}$ is calculated from the output $o_i$ as in the Equation \ref{eq:softmax_regression_pred}

\begin{equation}
    \label{eq:softmax_regression_pred}
    \hat{\mathbf{y}}=\operatorname{softmax}(\mathbf{o}) \text { where } \quad \hat{y}_{j}=\frac{\exp \left(o_{j}\right)}{\sum_{k} \exp \left(o_{k}\right)}
\end{equation}

\begin{figure}[ht]
  \centering
\includegraphics[width=.8\linewidth]{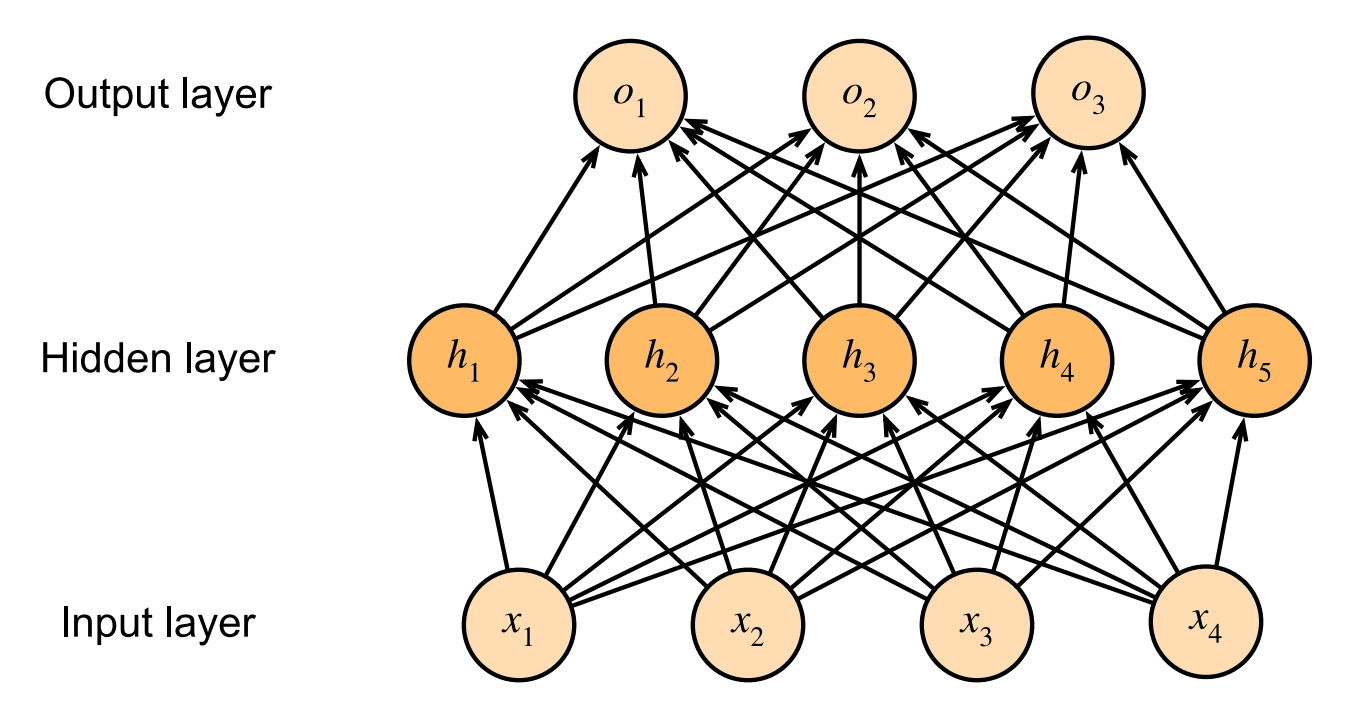}
  \caption{Demonstration of Multilayer Perceptrons.}
  \label{fig:mlp}
\end{figure}

The first architecture which is truly considered a deep learning model is multilayers perceptron (\gls{MLP}).
As demonstrated in Figure \ref{fig:mlp}, this model contains at least one hidden layer between the input layer and the softmax layer.
With this design, the model can learn both the representations in the hidden layers and the mapping function from the hidden signal to the output.
It is proved that the multilayer perceptrons can approximate any function with arbitrary accuracy \cite{hornik1989multilayer}.

\subsection{Convolutional Neural Networks}

Although the multilayer perceptrons can approximate any function, this architecture requires many parameters when the size of input increases.
Over more, the more parameters, the higher chance the model suffers overfitting issue.
Convolutional neural network (CNN) is invented to constrain the network's parameters to reduce the number of them and avoid overfitting.

\begin{figure}[ht]
  \centering
\includegraphics[width=.8\linewidth]{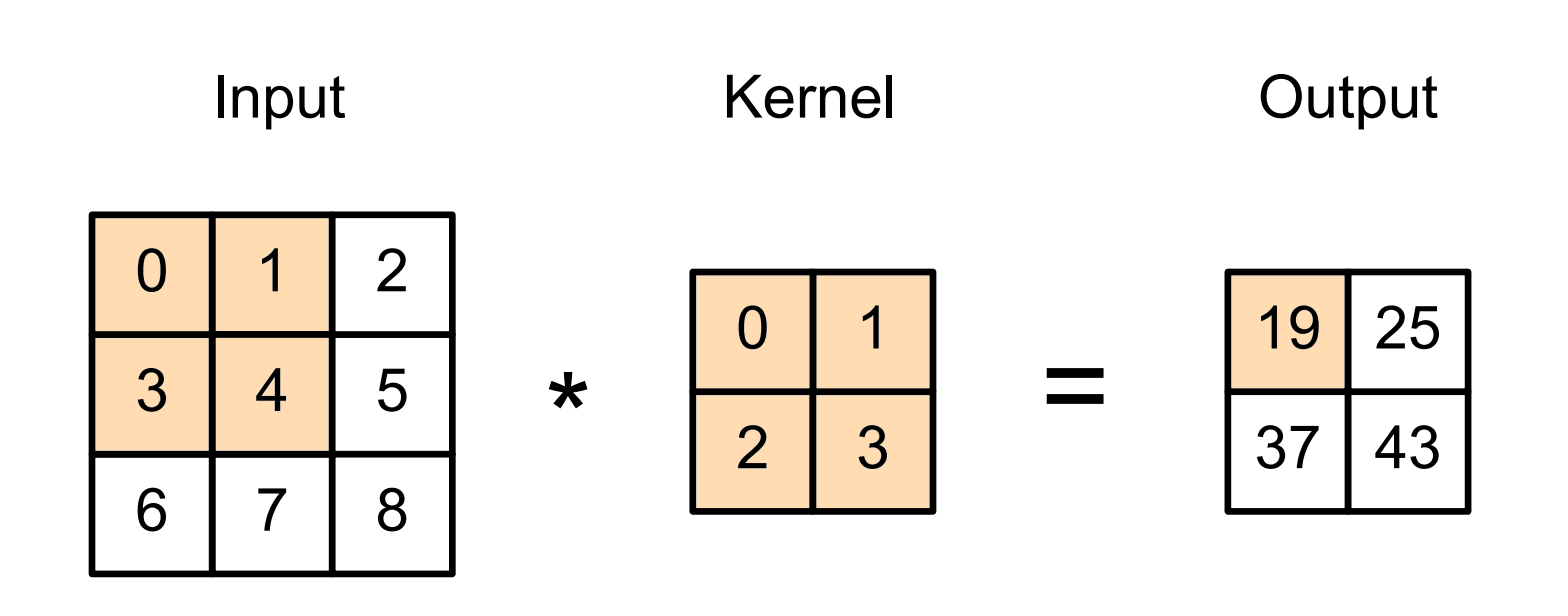}
  \caption{The convolution function in a CNN.}
  \label{fig:convolution}
\end{figure}

\begin{figure}[ht]
  \centering
\includegraphics[width=.8\linewidth]{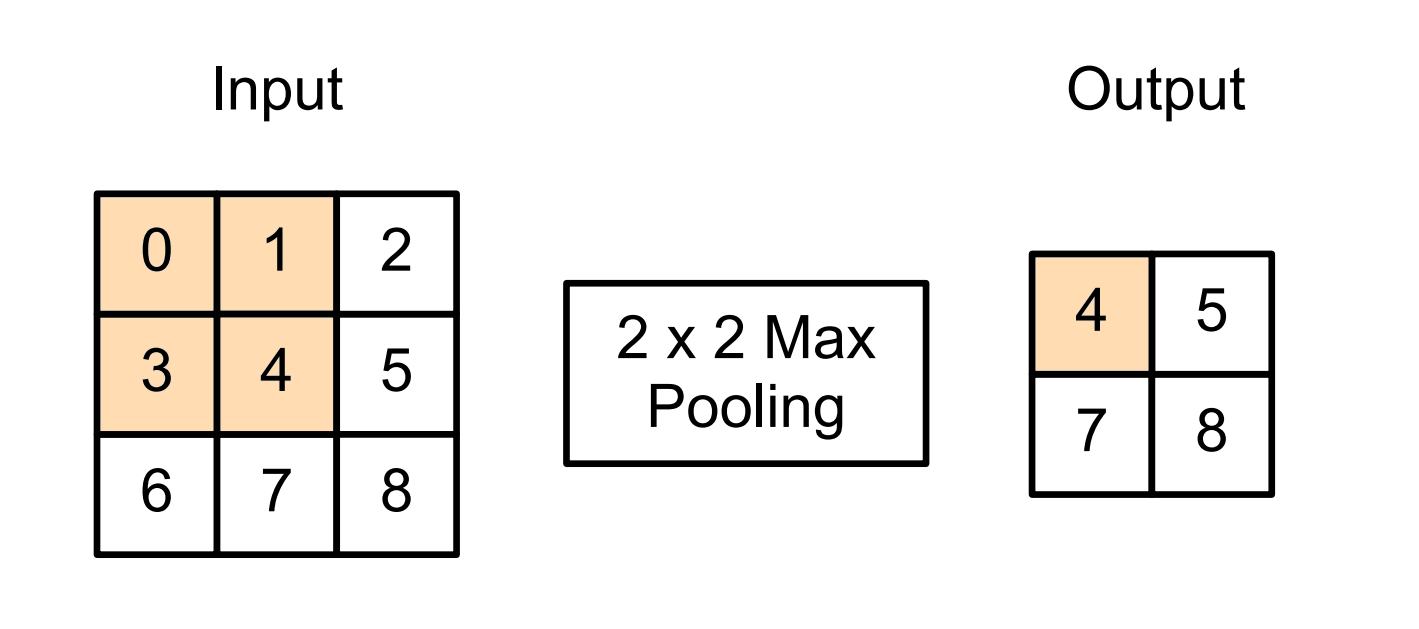}
  \caption{The pooling function in a CNN.}
  \label{fig:pooling}
\end{figure}

Figure \ref{fig:convolution} and \ref{fig:pooling} demonstrate the two main functions in a Convolutional Neural Network (\ie convolution function and pooling function).
Applying these functions to the inputs, the model can produce cross-correlation features in the next layer.
As a result, the number of parameters in the next layer is reduced and the model needs to extract the most important feature to optimize the loss value of the whole network.

\subsection{Recurrent Neural Networks}
The real-world data is often in the sequential format (\eg documents, movies, stock market, events, conversations).
As a result, it is required to have an appropriate architecture to model the sequence of data.
Intuitively, the information from the previous signal in the sequence is necessary to predict the consequent signal.

\begin{figure}[ht]
  \centering
\includegraphics[width=.8\linewidth]{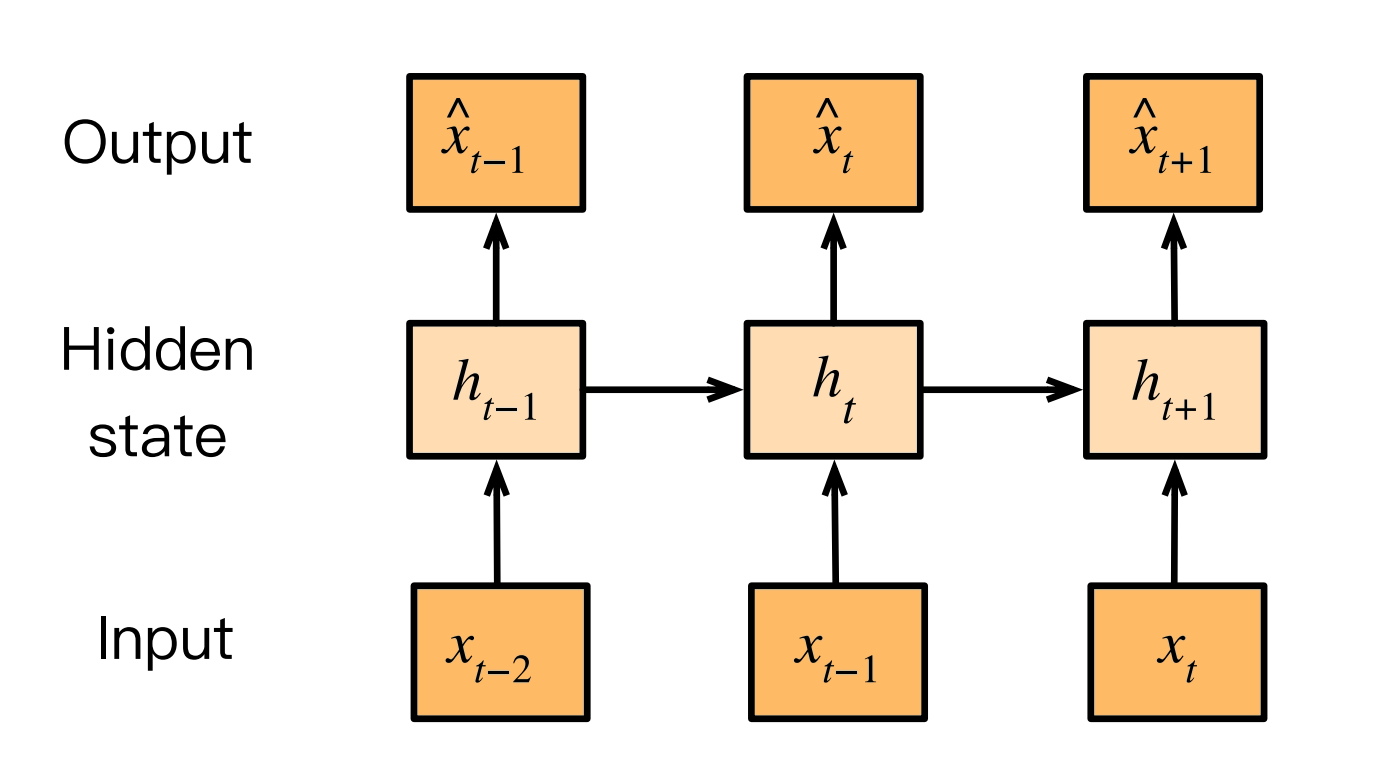}
  \caption{A simple diagram of the recurrent neural network.}
  \label{fig:rnn}
\end{figure}

Figure \ref{fig:rnn} shows a simple diagram of the recurrent neural network architecture. 
The general idea of this architecture is that a hidden state at time step $t$ is calculated from both the input and the hidden state at the previous timestep $t-1$ as in Equation \ref{eq:rnn}.
From the diagram, we can see that the computational route in a recurrent neural network is longer than in a feed-forward model.
This characteristic may lead to a technical problem namely \textit{vanishing gradient}.
This problem is solved by modern architectures with gating mechanisms.

\begin{equation}
    \label{eq:rnn}
    h_{t}=f\left(x_{t}, h_{t-1}\right)
\end{equation}

\begin{figure}[ht]
  \centering
\includegraphics[width=.8\linewidth]{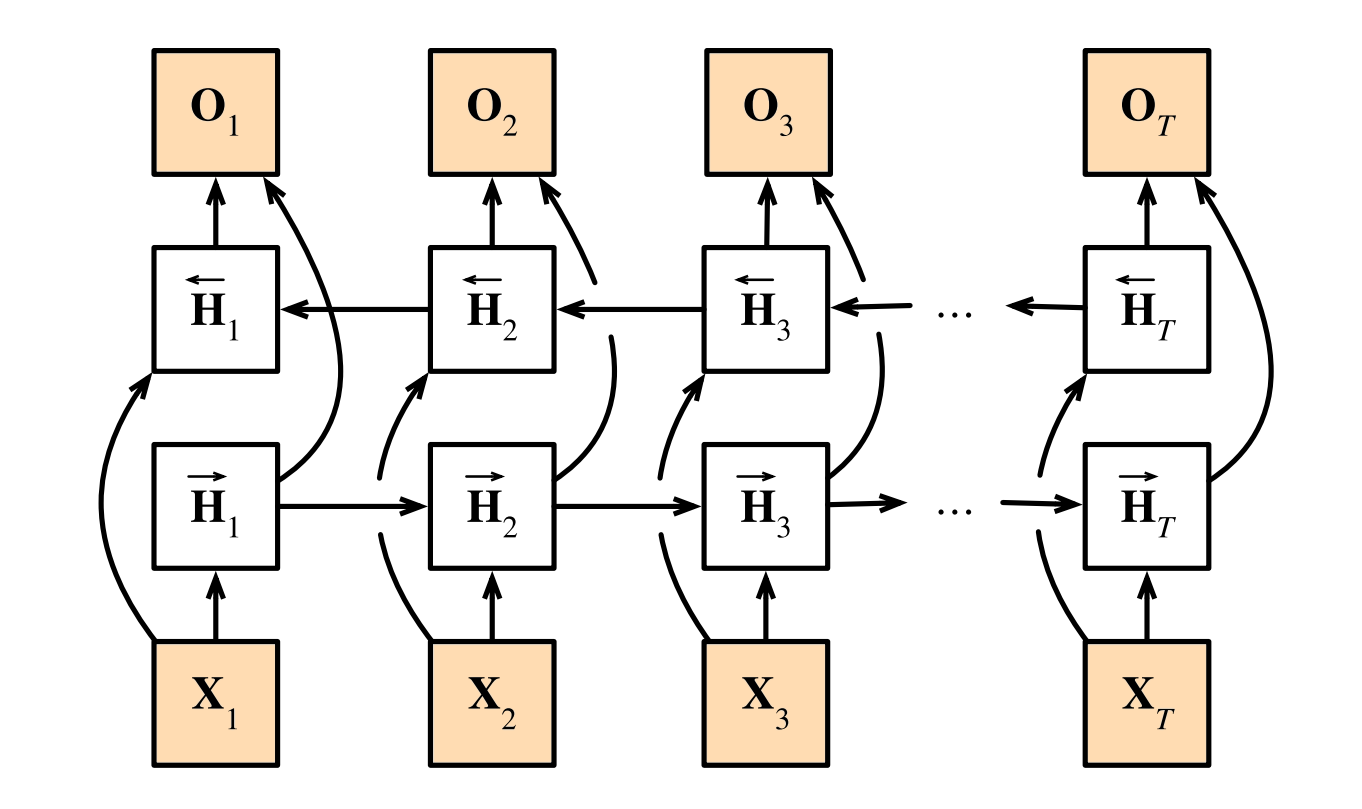}
  \caption{A recurrent neural network in a bi-directional paradigm.}
  \label{fig:bi_rnn}
\end{figure}

The recurrent neural network can also be designed in a bi-directional paradigm.
As demonstrated in Figure \ref{fig:bi_rnn}, the input of a hidden node comes from both directions.
This paradigm enables each node to access more information from the neighbor nodes in the sequence.

\section{Premiliary about Attentive Models}
\subsection{Attention Mechanism}
\label{sec:attention_mechanism}
In real life, information is abundant but meaningful information is not that abundant.
In information processing, we do not create new information but extract meaningful information from the original one.
Information often comes with noises, which is meaningless signals in the environment.
The definition of noise depends on the specific task we conduct and the aspect we want to learn in the data.
Attention is the mechanism that helps a model (\ie human model or machine model) to focus on the important parts of information.

\begin{figure}[ht]
  \centering
\includegraphics[width=.8\linewidth]{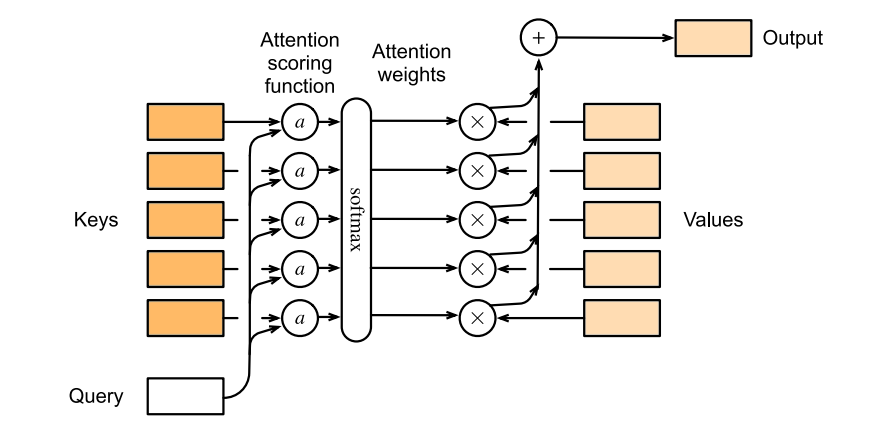}
  \caption{The attention mechanism is essentially a weighted sum of the values.}
  \label{fig:attention}
\end{figure}

Figure \ref{fig:attention} shows a diagram of a neural network using attention mechanism.
There are three important components of this paradigm which are \textit{query}, \textit{keys} and \textit{values}.
\textit{Query} is the signal for the model to know which option in the list of \textit{(keys, values)} it should pay attention to.
Let $\mathbf{q}$ be the query vector, $\mathbf{(k_i, v_i)}$ be the $i^{th}$ key, value pair in the candidate lists, function $\operatorname{f}$ to calculate the attentive output follows the Equation \ref{eq:attention}.

\begin{equation}
    \label{eq:attention}
    f\left(\mathbf{q},\left(\mathbf{k}_{1}, \mathbf{v}_{1}\right), \ldots,\left(\mathbf{k}_{n}, \mathbf{v}_{n}\right)\right)=\sum_{i=1}^{n} \alpha\left(\mathbf{q}, \mathbf{k}_{i}\right) \mathbf{v}_{i}
\end{equation}

\noindent In which $\alpha\left(\mathbf{q}, \mathbf{k}_{i}\right)$ is calculated by applying a softmax function on the values returned by function $a$ calculating the alignment score between a query vector and a key vector as in Equation \ref{eq:attention_softmax}.

\begin{equation}
    \label{eq:attention_softmax}
    \alpha\left(\mathbf{q}, \mathbf{k}_{i}\right)=\operatorname{softmax}\left(a\left(\mathbf{q}, \mathbf{k}_{i}\right)\right)=\frac{\exp \left(a\left(\mathbf{q}, \mathbf{k}_{i}\right)\right)}{\sum_{j=1}^{m} \exp \left(a\left(\mathbf{q}, \mathbf{k}_{j}\right)\right)}
\end{equation}

\subsection{Self-attention}

Self-attention is the featured mechanism of the Transformer-based architecture.
This computation allows the model to integrate the signal from different positions in a sequence to obtain better representations without a recurrent design.
This architecture is the base for a series of novel state-of-the-art pretrained models in many \gls{nlp} tasks.

\begin{figure}[ht]
  \centering
\includegraphics[width=.95\linewidth]{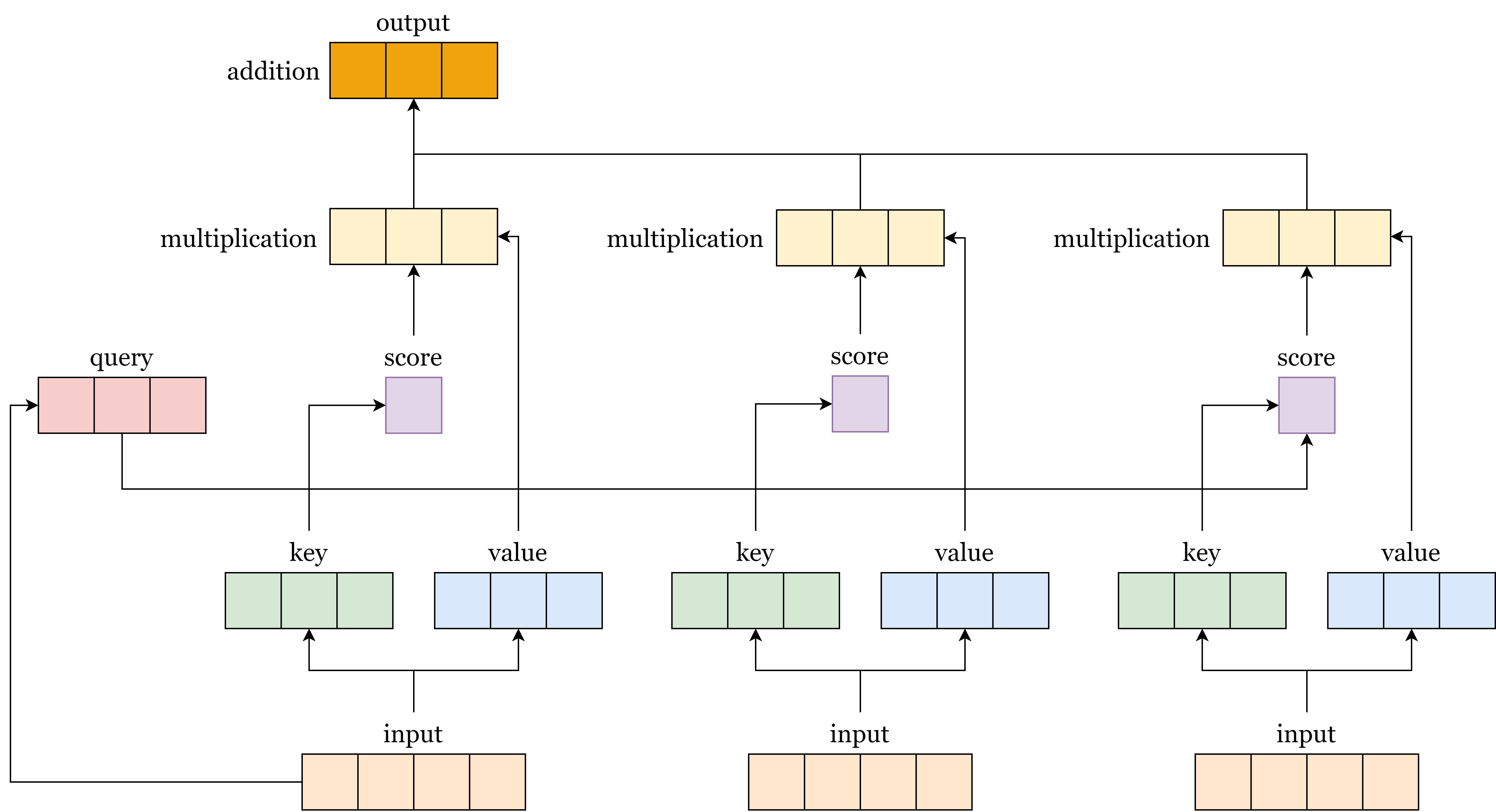}
  \caption{An example of self-attention computation.}
  \label{fig:self_attention}
\end{figure}

Figure \ref{fig:self_attention} demonstrates an example of self-attention computation.
In this example, we are calculating the new representation for the first input vector.
First, the query, key and value vectors are extracted from the inputs vector by corresponding weight matrices.
After that, we apply the computation as described in Section \ref{sec:attention_mechanism}  on the query vector of the first input and the (key, value) pairs of all the inputs to get the result.
The process is repeated for the whole sequence and the new representations contain signals from all of the different positions in the original sequence with different weights.

\subsection{Multi-Head Attention}

\begin{figure}[ht]
  \centering
\includegraphics[width=.95\linewidth]{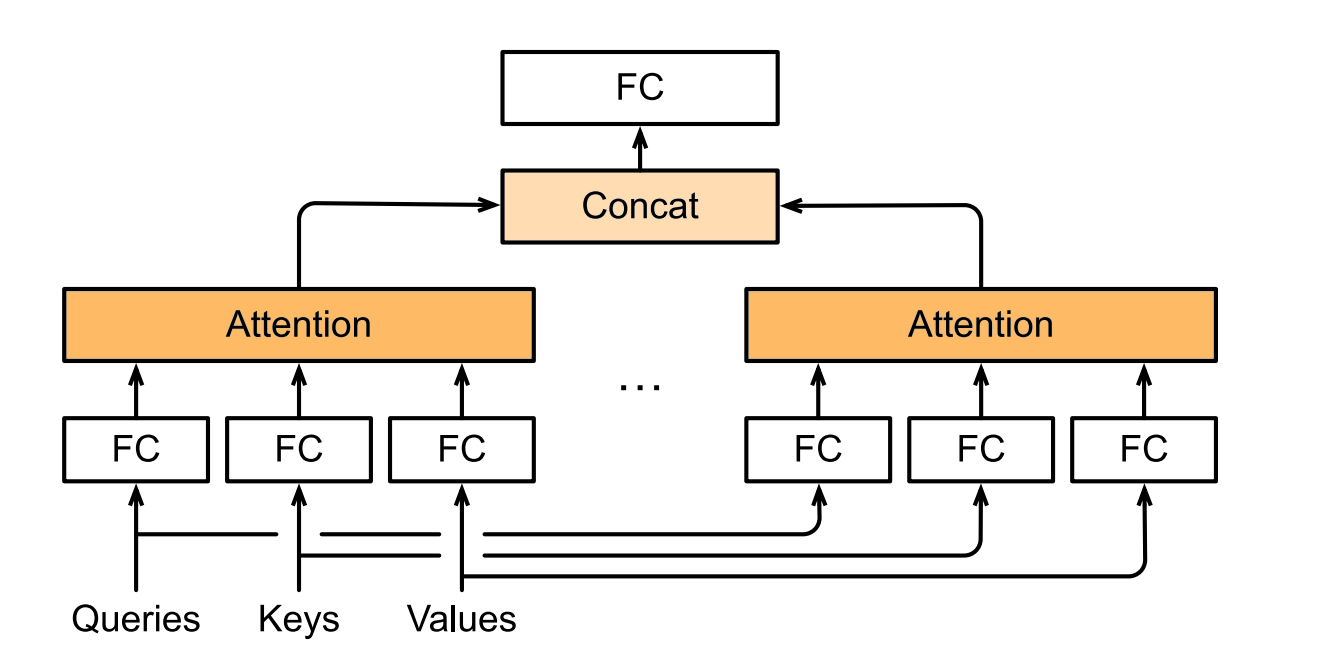}
  \caption{The representations of all heads are concatenated and transformed to get the final representation.}
  \label{fig:multihead_attention}
\end{figure}

Multi-head attention is another important idea in Transformer-based architectures.
This paradigm allows the model to have multiple viewpoints in the way the alignments are constructed in an input sequence.
As demonstrated in Figure \ref{fig:multihead_attention}, instead of using only one single attention module, this architecture calculates the attention representation in multiple subspaces (\ie multiple heads) then concatenate the signal from all heads to a vector.
After that, this vector is transformed by a fully-connected (\gls{FC}) layer to have the appropriate dimension.

\section{Characteristics of Legal Language}
\subsection{Legal Vocabulary}
Language is the main communication method of human beings.
For effective communication, the parties need to use the common language.
However, in reality, every different group of people uses a different sublanguage. 
Young people may use language that older people do not understand and vice versa.
To the extreme, everyone has unique thoughts so the vocabulary used by any two people can hardly be the same.
Not being an exception, legal language is essentially a sublanguage dedicated to describing concepts in law with its own vocabulary and grammatical~rules.

\begin{figure}[ht]
  \centering
\includegraphics[width=.9\linewidth]{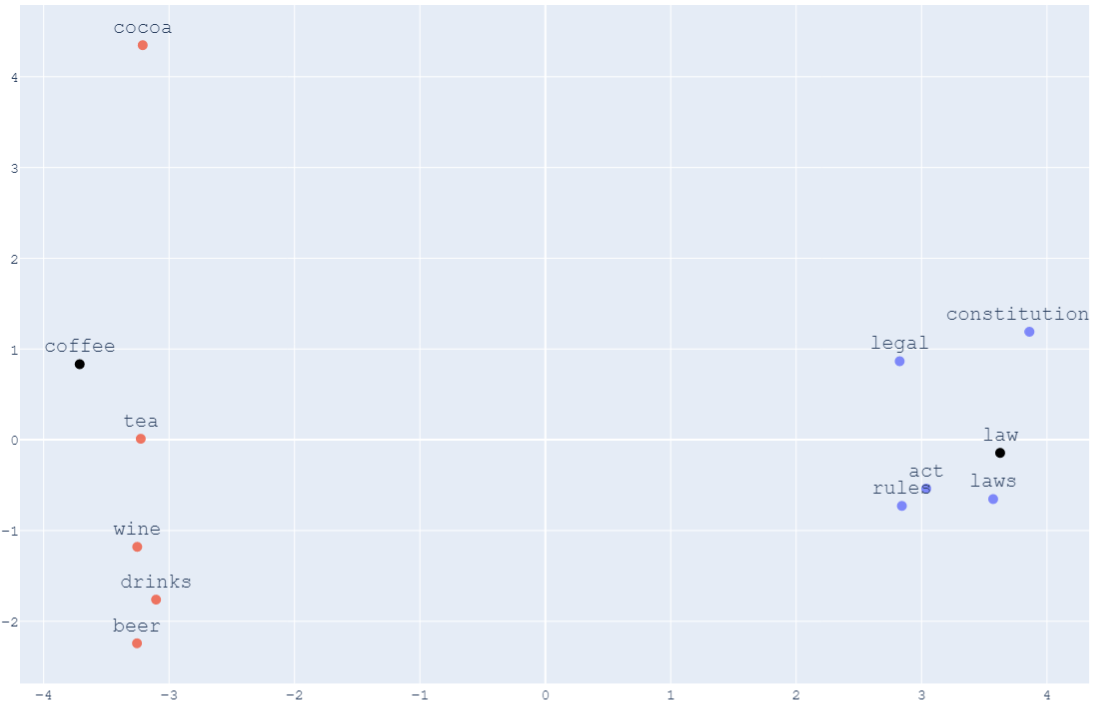}
  \caption{Distribution in the Word2Vec's vector space of the daily words and legal words.}
  \label{fig:w2v_difference}
\end{figure}

Figure \ref{fig:w2v_difference} visualizes the distribution in Word2Vec's vector space of some daily words and some legal words.
The position distribution is very different between the two groups.
This phenomenon may create a challenge for the state-of-the-art models in the general domain to perform well in the legal domain.
Indeed, legal language is not easy for lay readers to use and understand.
For example, not every English speaker understands the word \textit{mutatis mutandis}, which is a common word in the legal domain.
Hence, there need to be appropriate approaches to obtain good results in this domain.

\subsection{Challenges In Legal Processing}
Legal processing refers to automated extracting meaningful information from legal text on the aspect of interest.
Because of its characteristics, there are challenges that need to be solved to obtain a good performance in this domain.
These challenges come from the legal language, the complexity of the legal sentence and the scarcity of annotated data for some specific tasks in legal processing.

The following is an example of a legal article in the Japanese Civil Code:

\begin{quotation}
\textbf{Article 395}

(1) A person that uses or profits from a building subject to a mortgage by virtue of a lease that cannot be duly asserted against the mortgagee, and that is set forth as follows (in the following paragraph referred to as ``mortgaged building user'') is not required to deliver that building to the purchaser thereof until six months have passed from the time when the purchaser purchased that building at auction: 

(i) a person that has been using or profiting from the building since prior to the commencement of auction procedures; or 

(ii) a person that is using or profiting from the building by virtue of a lease given after the commencement of auction procedures by the administrator of compulsory administration or execution against earnings from immovable collateral. 

(2) the provisions of the preceding paragraph do not apply if the purchaser, specifying a reasonable period of time, issues a notice to the mortgaged building user demanding payment of consideration for a period of one month or more with respect to the use of the building referred to in that paragraph that has been made after the time of purchase by the purchaser, and no payment is made within that reasonable period of time.
\end{quotation}

\noindent It can be seen that the above legal article contains long sentences, its format is very different from paragraphs in other types of documents like news, reports or novels.
At this length, legal documents can make it difficult for robust models that have made many breakthroughs in the general domain.

Adding to the challenges of the inputs in legal processing, the outputs of legal automated systems need to meet higher quality conditions compared to ordinary systems.
The accuracy of the system, as well as the quality of generated content, is crucial to bring these systems to real life.
\section{Literature Review}
Automated legal text processing is not a new research direction. 
This is a direction of research that dates back to the early years of computers. 
Approaches to various problems of automatic legal text processing evolve with the growth of the computing power of computers as well as the solidification of scientific and technological foundations.
This is a difficult field, so each study can only solve part of the problems that exist in it.
Even so, they are important foundations for driving the advancement of automated legal document processing.

\begin{figure}[ht]
  \centering
\includegraphics[width=.9\linewidth]{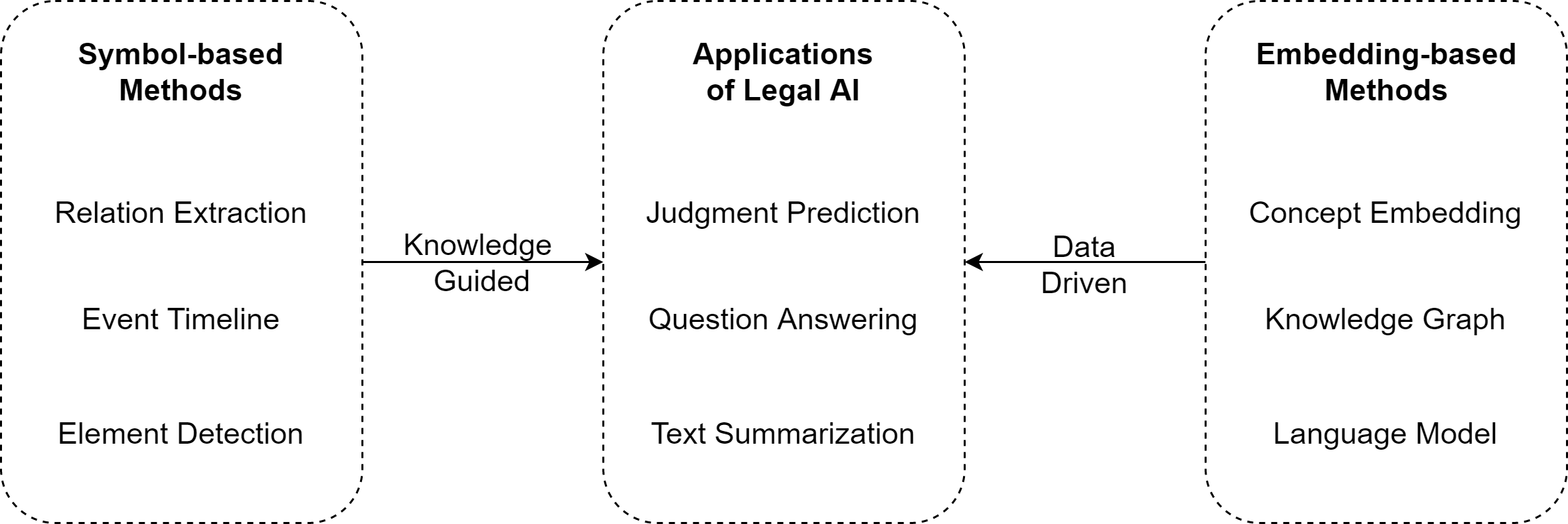}
  \caption{An overview of methods and applications in legal AI \cite{zhong2020does}.}
  \label{fig:legalAI}
\end{figure}

Zhong \etal \cite{zhong2020does} demonstrate an overview of methods and applications in legal AI as in Figure \ref{fig:legalAI}. 
They divide the approaches of automated legal processing into two groups, symbol-based methods and embedding-based methods. The first group uses known knowledge to guide the system, the second group uses patterns that the model learns from the data to make decisions.
The legal AI applications they list in their work are representative and do not cover the full range of real-life applications of the field.

Laying the first bricks in this field are rule-based systems \cite{ulmer1963quantitative,buchanan1970some,susskind1986expert,rissland1988artificial}.
These systems take the form of expert systems or lexical matching systems. 
They make finding and retrieving information in the legal field easier. 
In addition, they can perform logical inferences in the law, as long as the data is described and represented appropriately for computers to understand.
However, the disadvantage of these systems is the requirement of rules designed by humans. 
With simple sets of rules, these systems become rigid. 
More sophisticated rule sets require a large human effort.

With the development of the internet and the explosion of digital data, methods of using machine learning \cite{basgalupp2009legal,pflueger2015predicting}, especially deep learning \cite{chalkidis2019deep,tran2020encoded,kien2020answering}, are becoming more and more popular in NLP in general and automated legal processing in particular.
Catering to this trend, a variety of datasets \cite{xiao2018cail2018,duan2019cjrc,chalkidis2019large,zhong2020jec,leitner2020dataset} and tasks \cite{aletras2016predicting,cardellino2017legal,ye2018interpretable,bhattacharya2019comparative} are introduced. 
Besides, competitions \cite{kano2018coliee,rabelo2019summary,rabelo2021coliee,nguyen2021summary} are also organized to aggregate and search for optimal solutions to tasks within a given resource.

COLIEE \footnote{https://sites.ualberta.ca/~rabelo/COLIEE2021/} is an annual competition for automated legal document processing with high-quality data provided in both case law and statute law, which is the valuable resource for this dissertation.
Case law and statute law are the two largest sources of law in the world, based on these two sources of law, social relations are adjusted in accordance with international practices and national laws. 
For case law, the cases that are heard first will be used as the basis for handling the following cases.
In statute law, legal documents are the main basis for court decisions.
The competition on building automated legal text processing systems is a challenging and inspiring one.
Until COLIEE 2020, there are in total of 4 tasks:
\begin{itemize}
    \item Task 1 and Task 2 are case law retrieval and entailment problems.
    \item Task 3 and Task 4 are statute law retrieval and entailment problems.
\end{itemize}
COLIEE 2021 introduces one more task for legal question answering (Task 5).

\chapter{Factor Analysis for Deep Legal Models}
\label{chap:investigate}
\section{Overview}

This chapter introduces our three works in verifying what factors that affect the performance of deep learning models in general and deep legal models in particular.
The finding in this chapter is the base for the problem formulation as well as the proposed approaches in the following chapters.
The contribution of this chapter is as follows.
First, we propose a novel way to assess different methods of data representation so that we can understand the correlation between data representation and performance in specific tasks.
Second, we conduct experiments and study how data amount affects the learning ability of the models.
Third, we verify the importance of understanding the data characteristic in designing appropriate architectures for a given task.
\section{Impact of Data Representation}

\subsection{Introduction}
Data representation is crucial to the construction of any effective system in computer science.
Significantly contributing to the breakthroughs of the deep learning approach are the embedding techniques.
Nowadays, there are two common families of embedding, which are word vectors (\eg Word2Vec~\cite{mikolov-efficient-estimation} and GloVe~\cite{pennington-glove}) and contextual embeddings (\eg BERT \cite{devlin2018bert},  GPT-2~\cite{radford2019language}, GPT-3~\cite{brown2020language}, BART \cite{lewis2019bart}).
We need the embeddings in deep learning models because the text needs to be converted into a numerical form to be processed by computers.
Embedding emerges from the need to represent words in natural language in a numerical form that computers can process.
These techniques evolved from the very simple approaches to the state-of-the-art methods we see today.
In this section, we examine different embeddings and quantitatively compare their representational capacities for the legal domain.

Mentioning the classical methods, we have \gls{BOW}, \gls{TF-IDF}, which is merely based on the word occurrence to make the numeric representation.
These approaches are good for lexical matching problems in the earlier days of NLP.
However, in science and technology, we never stop expecting the systems to become more and more powerful and these methods become ineffective for problems requiring semantic or global context understanding.
Bag-of-words vectors are obtained by spotting the word position in the dictionary while TF-IDF distributional vectors are based on the popularity of the words in the whole corpus.

\begin{figure}[ht]
  \centering
\includegraphics[width=.8\linewidth]{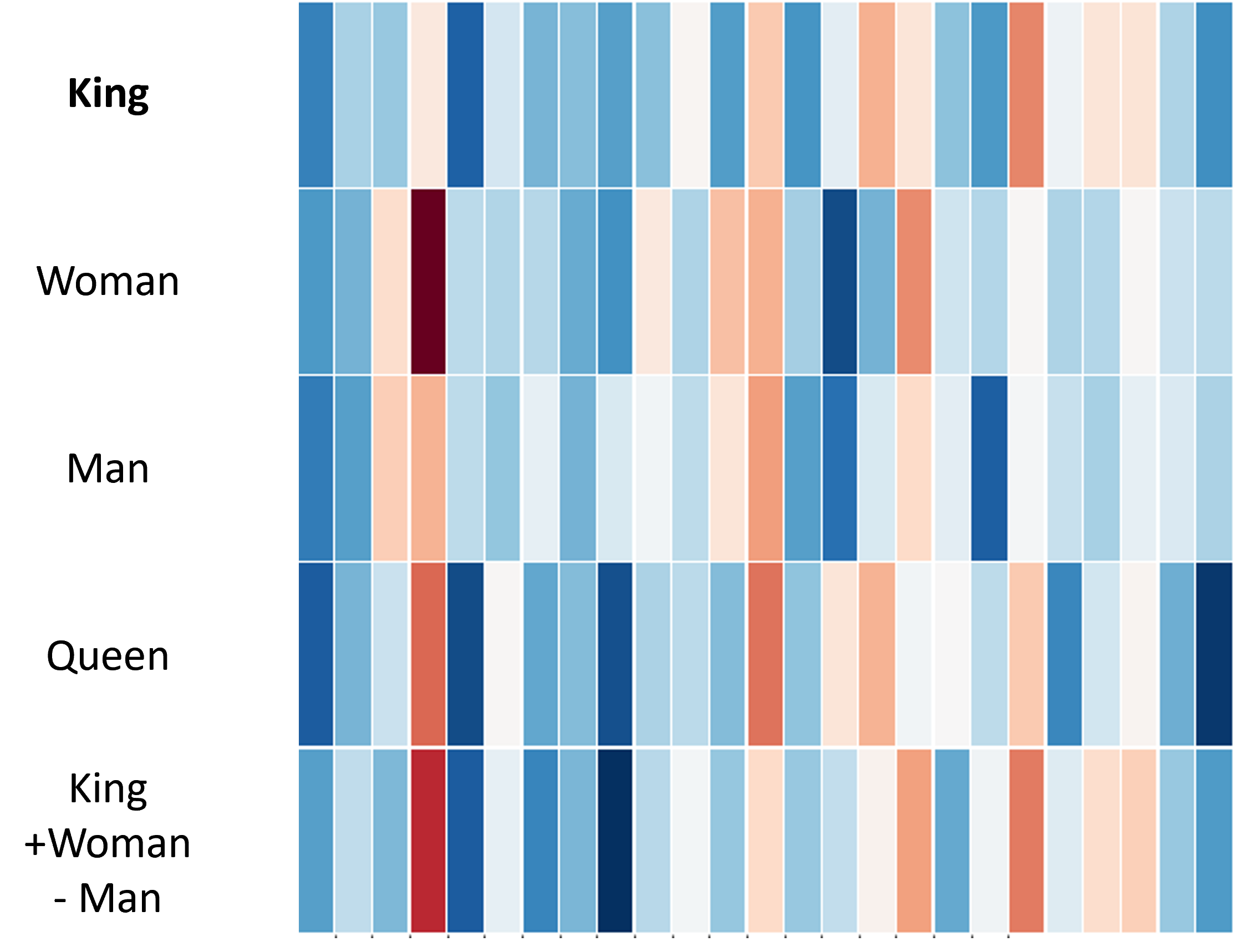}
  \caption{Visualizing in GloVe embedding space, we can create a vector that closely resembles the vector of \textit{Queen} by adding and subtracting the vectors of \textit{King}, \textit{Woman} and \textit{Man}.}
  \label{fig:word2vec_example}
\end{figure}

The word embedding approaches are an improvement over counting-based data representation methods.
Word2Vec is the algorithm based on prediction, the most well-known word embedding technique. 
The idea of this approach is that when trying to predict a word using its context, the model obtains the correlation information in the corpus.
Continous Bag-of-words (\gls{CBOW}) and Skip-gram are two methods in the Word2Vec approach.
As a simple approach, Word2Vec's algorithms only consider the surrounding context and the target word, but not the full-text content.
Dealing with that limit, GloVe uses neural embedding to parse co-occurrence matrices into more significant and weighted vectors.
Trained on a large corpus, these embeddings provide interesting information from word vectors. 
For example, with well-trained vectors, we can find synonyms, antonyms, or even add and subtract meanings of words to get a new word in  as the example in Figure \ref{fig:word2vec_example}.

\begin{figure}[ht]
  \centering
\includegraphics[width=.8\linewidth]{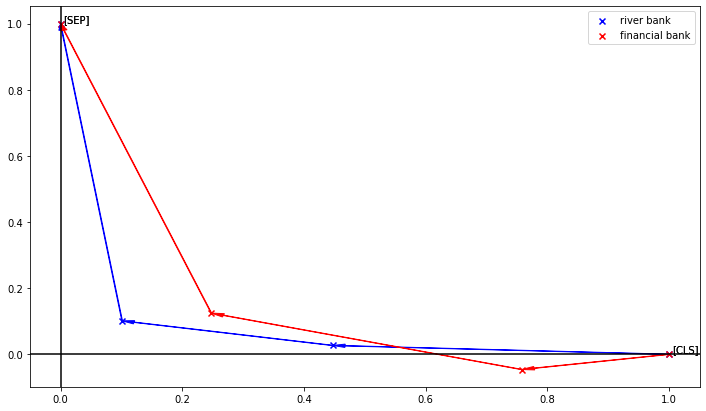}
  \caption{BERT represents the same word \textit{bank} in two different phrases as different representation vectors.\label{fig:bert_example}}
\end{figure}

Instead of using word vectors represented in the vector space, the Transformer \cite{vaswani2017attention} models use multi-head attention mechanism to identify the relationship between words in specific context usage, in different aspects.
This method is therefore called contextual embedding.
This method proves to be advantageous in cases where a word has multiple meanings. 
For example, the word \textit{bank}, which could be a river's bank or financial bank. In word embedding, it can only be represented by a single vector, whereas, with contextual embedding, it can be interpreted in many aspects, depending on the surrounding words.
In addition, some variants of this architecture that use subword tokenizer instead of word tokenizer have been shown to provide better performance in embedding representation.
Figure \ref{fig:bert_example} is the representation of word vectors of BERT in two-dimensional space for the two phrases \textit{river bank} and \textit{financial bank}. We can see that the word \textit{bank} in two different phrases has different representation vectors.
This allows the model to understand the semantics of words more flexibly, solving technical problems in the case of synonyms.

\vfill

The common mission of word embedding and contextual embedding is to convert words in the discrete representation into vectors in the continuous space.
These representations are pretrained and can be used in downstream tasks.
The information in these continuous representations directly affects the performance of the model.
In this section, we study different pretrained approaches to represent the raw data into the embeddings' vector space.

\subsection{Research Method}
The quality of data representation has a direct influence on the performance of the models.
If the input to the neural network contains all words that are outside the vocabulary, the model can not form a meaningful mapping from the input to the output in the training and prediction samples.
In addition, the performance is also affected if the embedding layer is not well trained, or in other words, the words in the vocabulary are not represented by an exact relative position in the vector space.
Hence, in this section, we propose quantitative metrics (Equations \ref{eq:lvc}, \ref{eq:leca}) and introduce a visualization approach for embedding on the legal domain.

These metrics are based on an assumption that we have a set of standard legal terms $\mathcal{L}$ and a verification legal corpus $\mathcal{D}$ as a set of sentence $s_i$.
The quality of embeddings is evaluated by the ability to map words in $\mathcal{D}$'s samples to the vector space and their relative positions.

Let $\mathcal{V}_E$ be the vocabulary of the embedding $E$, the first proposed metric $LVC_E$ (Legal Vocabulary Coverage) is calculated as follow:
\begin{align}
    \mathcal{C}_E&=\mathcal{V}_E \cap \mathcal{L}\\
    LVC_E&=\frac{|\mathcal{C}_E|}{|\mathcal{L}|}\label{eq:lvc}
\end{align}

Although an important metric, $LVC$ is not good enough to evaluate $E$ because it does not take into account the position of the legal terms represented.
Therefore, we propose $LECA_E$ (Legal Embedding Centroid-based Assessment) as a metric based on the position of the word vectors of the legal terms appearing in $\mathcal{V}$.
Let $O_E$ be the centroid of points represented by $\mathcal{C}_E$'s vectors, $P_i^j$ be vector of the $j^{th}$ word in $s_i$, the $i^{th}$ sample in the corpus $\mathcal{D}$ and $d$ be a vectorial distance metric function,  $LECA$ is calculated as follow:

\begin{align}
    LECA_E&= \frac{1}{|\mathcal{D}|} \sum_{i=0}^{|\mathcal{D}|}\frac{1}{|s_i|} \sum_{j=0}^{|s_i|} d(P_i^j, O_E)\label{eq:leca}
\end{align}

When applying $LVC$ and $LECA$ in practice, we need to use these two scales together to avoid misjudgments in extreme cases.
In particular, with the $LVC$ being too small, the $LECA$ may not accurately reflect the effectiveness of the embedding.
In contrast, an embedding with large $LVC$ but its results on LECA are insignificant, the embedding fails to represent legal terms in its space.

In addition to quantitative measurement, visualization is an important aspect of the explanation.
It helps us to focus on the important aspects of the phenomenon in order to understand it. 
Embedding visualization is essentially the representation of dimensional data that humans can perceive.
For the problem posed in this section, the relative position of the legal term in the entire vocabulary is the most important information for understanding the nature of embedding in the legal domain.
The core information that makes up the relative position of the vector representing terms in the corpus is the similarity between them.
Hence, we propose to use the t-distributed stochastic neighbor embedding \cite{van2008visualizing} technique to transform the vector space and visualize them in a smaller dimensional space.

\subsection{Experiments}
\label{sec:ch3_data_rep_exp}
\subsubsection{Experimental Settings}
We conduct experiments to measure existing embedding systems on the metrics proposed in the previous section.
The representatives of word embedding technique we choose are FastText~\cite{mikolov2018advances}, GloVe~\cite{pennington-glove} and Law2Vec~\cite{chalkidis2019deep}. For contextual embedding, we use BERT~\cite{devlin2018bert}, LEGAL-BERT~\cite{chalkidis2020legal} and BERTLaw~\cite{nguyen2020jnlp}. 
Their configurations are all base, uncased.
For the legal term set $\mathcal{L}$, we use 1,000 top terms provided by LexPredict\footnote{https://www.lexpredict.com/}.
For the legal corpus $\mathcal{D}$, we use 10,000 arbitrary legal sentences sampled from SigmaLaw dataset\footnote{https://osf.io/qvg8s/}.
We use \textit{cosine distance} as the distance metric function $d$.

\begin{table}
\centering
\caption{Experimental settings for word and contextual embeddings.\label{tab:exp_setting}}
\begin{tabular}{|l|c|c|}
\hline
\textbf{Systems} & \textbf{Vocabulary Size} & \textbf{Dimension} \\ \hline
FastText~\cite{mikolov2018advances}               & 30,000                   & 300                          \\ \hline
GloVe~\cite{pennington-glove}            & 30,000                    & 300                          \\ \hline
Law2Vec~\cite{chalkidis2019deep}           & 30,000                    & 200                          \\ \hline 
BERT~\cite{devlin2018bert}               & 30,522                    & 768                          \\ \hline
LEGAL-BERT~\cite{chalkidis2020legal}              & 30,522                    & 768                          \\ \hline
BERTLaw~\cite{nguyen2020jnlp}              & 32,000                    & 768                          \\ \hline
\end{tabular}
\end{table}

As shown in Table~\ref{tab:exp_setting} the vocabulary of contextual embeddings contains about 30K tokens. Towards a fair comparison, we also only consider the first 30K words in word embeddings.
To extract word vectors from contextual embeddings, we take the sum of the values of the last 4 hidden layers in the architecture. 
The dimensional size of contextual embeddings is the size of the hidden vector which is 768. In our experiment, we select the largest size of word embeddings provided by their authors (300D for GloVe, FastText and 200D for Law2Vec).

For word embedding, the vectors are taken directly from the pretrained data.
For contextual embedding, vectors of legal terms in $\mathcal{L}$ are computed without context, whereas word vectors of samples in $\mathcal{D}$ are extracted from the computation on their contextual sentences. 
Since we could not find a reputable source for a set of legal subwords, we tokenize the input for contextual embedding on word level. 
This workaround may slightly degrade their actual performance. 
Therefore, in our experiment, we do not directly compare the results between word embedding and contextual embedding, but only compare embedding of the same type with each other.

\subsubsection{Experimental Result}

\begin{table}
\centering
\caption{\textit{LVC} and \textit{LECA} score of word embeddings.\label{tab:wv_result}}
\begin{tabular}{|l|c|c|}
\hline
\textbf{Systems} & \textbf{LVC} & \textbf{LECA} \\ \hline
GloVe~\cite{pennington-glove}            & 0.719                   &0.478                          \\ \hline
FastText~\cite{mikolov2018advances}               &  0.725                    & 0.467                          \\ \hline
Law2Vec~\cite{chalkidis2019deep}           & \textbf{0.770}                   & \textbf{0.434}                      \\ \hline 
\end{tabular}
\end{table}

Table \ref{tab:wv_result} shows the results of word embeddings on the \textit{LVC} and \textit{LECA} metrics. 
Law2Vec achieves state-of-the-art results on both of these metrics.
Comparatively, GloVe has a better \textit{LVC} score and a worse \textit{LECA} score than FastText.
Models with lower \textit{LVC} scores are more prone to out-of-vocab problems, and models with higher \textit{LECA} scores are more likely to fall into local extremes or require more training iterations to converge.
Therefore, for a problem in the legal domain, using Law2Vec with the word vector embedding method may lead to better results.

\begin{table}
\centering
\caption{\textit{LVC} and \textit{LECA} score of contextual embeddings\label{tab:ce_result}}
\begin{tabular}{|l|c|c|}
\hline
\textbf{Systems} & \textbf{LVC} & \textbf{LECA} \\ \hline
BERT~\cite{devlin2018bert}               & 0.680                    & 0.758                          \\ \hline
LEGAL-BERT~\cite{chalkidis2020legal}              & \textbf{0.737}                    & 0.618                          \\ \hline
BERTLaw~\cite{nguyen2020jnlp}              & 0.689              & \textbf{0.612}                          \\ \hline
\end{tabular}
\end{table}

Table \ref{tab:ce_result} shows the results of contextual embeddings.
LEGAL-BERT scores state-of-the-art on \textit{LVC} and BERTLaw scores state-of-the-art on \textit{LECA}.
From the experimental results, it can be seen that embeddings pretrained on the legal domain give better performance on the \textit{LVC} and \textit{LECA} metrics.

It can also be seen that contextual embeddings have lower \textit{LVC} and higher \textit{LECA} than word embeddings.
This can be explained by the fact that these contextual embeddings are forced to calculate the vectors on the word level instead of the subword level as in their pretraining phase. Therefore, the cross-comparison between word embeddings and contextual embeddings may lead to a bias conclusion.

\subsubsection{Visualization}
\label{sec:viz}
We visualize how the aforementioned embeddings depict legal and non-legal vectors in their space.
We first filtered out the 2000 most common words in the vocabulary in each embedding for visualization purposes.
We obtain the vector values for these terms and color the points respectively based on the \textit{legal} and \textit{nonlegal} labels.
Then, we use the t-distributed stochastic neighbor embedding algorithm to find the appropriate 3-dimensional representation and display it.

\begin{figure*}
     \centering
     \begin{subfigure}[b]{0.3\textwidth}
         \centering
         \includegraphics[height=0.8\textwidth]{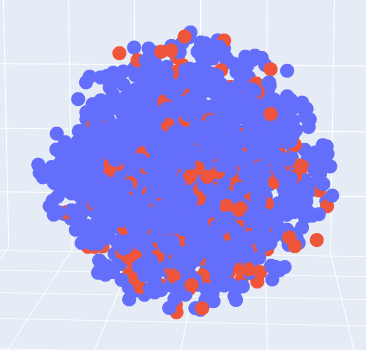}
         \caption{GloVe}
     \end{subfigure}
     \hfill
     \begin{subfigure}[b]{0.3\textwidth}
         \centering
         \includegraphics[height=0.8\textwidth]{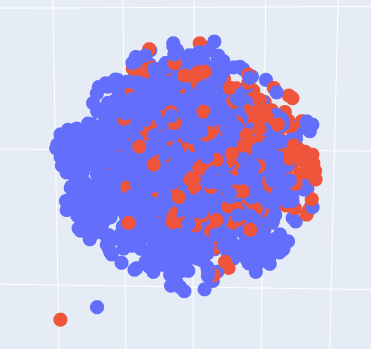}
         \caption{FastText}
     \end{subfigure}
     \hfill
     \begin{subfigure}[b]{0.3\textwidth}
         \centering
         \includegraphics[height=0.8\textwidth]{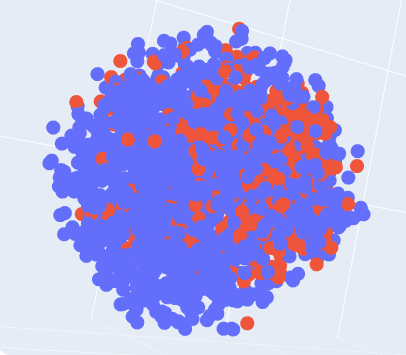}
         \caption{Law2Vec}
     \end{subfigure}
        \caption{Visualization of the word embeddings GloVe, FastText, and Law2Vec, respectively. The red points correspond to the legal terms in the set $\mathcal{L}$, the remaining points are blue.}\label{fig:ch3_we_viz}
\end{figure*}

Figure \ref{fig:ch3_we_viz} shows the visualization of the word embedding space of GloVe, FastText, and Law2Vec. 
For all 3 embeddings, the positions of the points form a sphere in space.
We can see that the legal points, although interspersed with other points, are concentrated in a particular region in space. 
Based on what is observed with the visualization, GloVe's embedding distinguishes legal points worse than FastText and Law2Vec.

\begin{figure*}
     \centering
     \begin{subfigure}[b]{0.35\textwidth}
         \centering
         \includegraphics[width=0.9\textwidth]{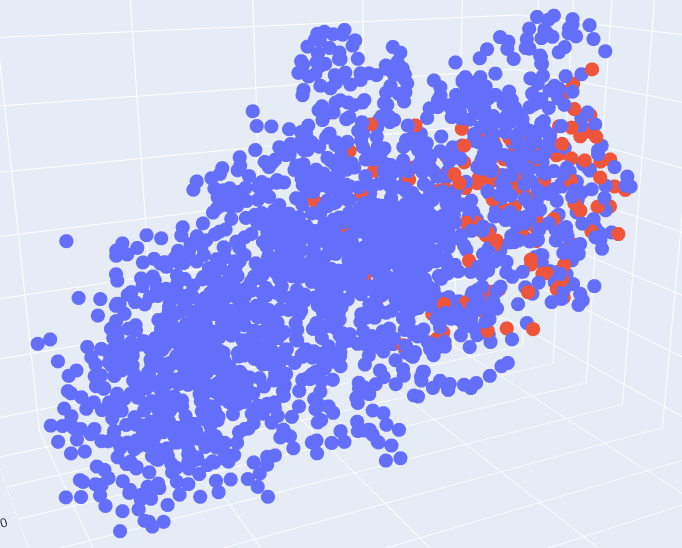}
         \caption{BERT}
     \end{subfigure}
     \hfill
     \begin{subfigure}[b]{0.278\textwidth}
         \centering
         \includegraphics[width=0.9\textwidth]{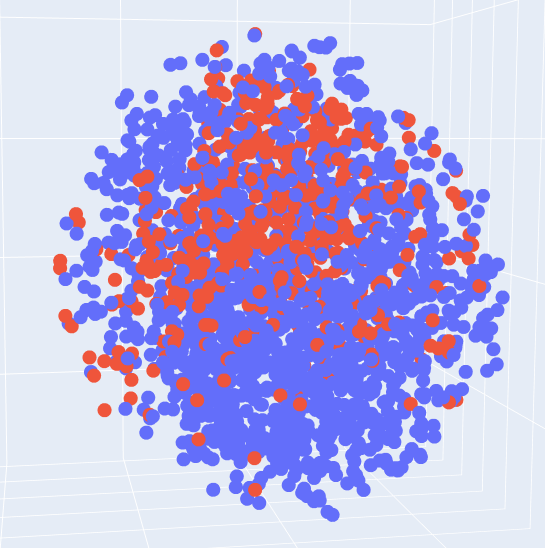}
         \caption{LEGAL-BERT}
     \end{subfigure}
     \hfill
     \begin{subfigure}[b]{0.298\textwidth}
         \centering
         \includegraphics[width=0.9\textwidth]{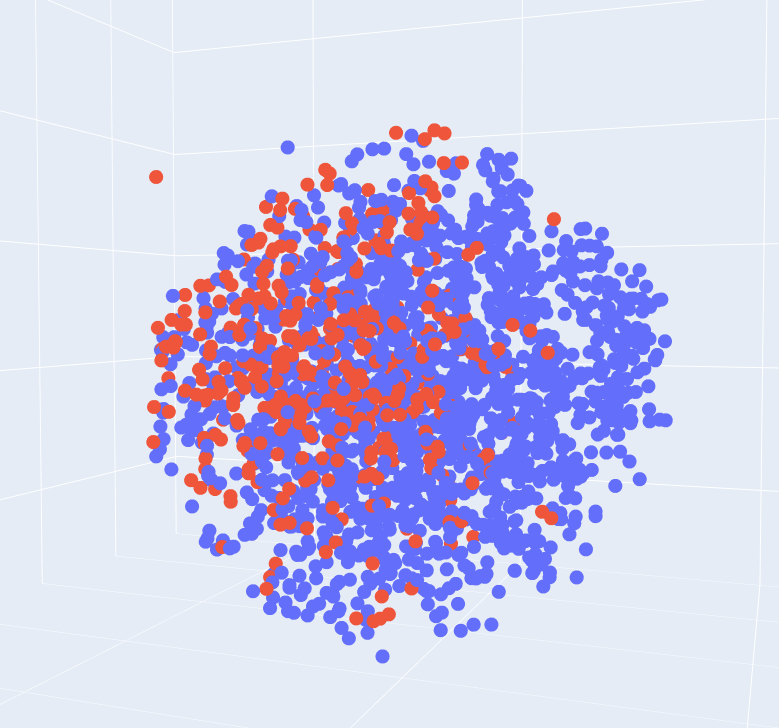}
         \caption{BERTLaw}
     \end{subfigure}
        \caption{Visualization of the contextual embeddings BERT, LEGAL-BERT, and BERTLaw, respectively. The red points correspond to the legal terms in the set $\mathcal{L}$, the remaining points are blue.}\label{fig:ch3_ce_viz}
\end{figure*}

Figure \ref{fig:ch3_ce_viz} demonstrates the visualization of the embedding space of BERT, LEGAL-BERT, and BERTLaw.
These visualizations all show a pretty good distinction between legal points and the rest.
The positions of the vectors in LEGAL-BERT and BERTLaw form a sphere in space, whereas the shape of the embedding distribution of BERT tends to be more distorted. This can be explained by the fact that this model is trained on a general domain with more diverse data domains than LEGAL-BERT or BERTLaw. Therefore, the survey in interdisciplinary data of legal and other domains is an interesting research direction in the future.

\subsection{Discussions}

This section analyzes the characteristics of different legal embedding techniques and proposes quantitative metrics and visualization for an explanation purpose.
Through the experimental results and visualization, we have several conclusions.
Firstly, the \textit{LVC} and \textit{LECA} metrics proposed in the article are suitable for the properties of the measured object and with the results reported in related articles. 
Secondly, embeddings that are pretrained with data in the legal domain tend to achieve higher results on these two metrics.
Thirdly, a combination of quantitative and visualization scales can increase the explainability of embeddings in the legal domain.

In addition, we have some discussions about using the result in the section and the future directions.
First, the values of the two proposed metrics would change for a different legal term set $\mathcal{L}$ and a legal corpus $\mathcal{D}$. Therefore, for a fair conclusion, they should be used for comparisons under the same conditions.
Second, although limited by resources as described in Section \ref{sec:ch3_data_rep_exp}, the formulas of these metrics are general. As a result, it is possible to compare word embeddings with contextual embeddings directly when the resource of the legal subword set is available.
Third, visualization of the original BERT shows a difference in its word representation compared to other variants. Although legal variants are reported to be better on legal tasks, surveys of interdisciplinary data are also an interesting research direction.

\newpage

\section{Impact of Data Amount}
\label{sec:ch3_data_amount}
\subsection{Introduction}
Data is the most important resource in machine learning, especially in deep learning.
In this section, we qualitatively test this hypothesis by proposing a way to increase the number of training samples in a legal problem and compare the performance.
The conclusions we draw from this section are important to prove the requirement of data amount to improve the performance in legal processing.
Facing the sparse data problem, the same architecture may dramatically decrease the performance.

The idea of this research comes up when trying to solve the problem of legal question answering based on textual entailment in legal text in COLIEE 2019.
Given a statement, we need to verify its lawfulness by finding its supportive article in the law as the following example:
\begin{itemize}
        \item \textbf{Legal Statement:} An unborn child may not be given a gift on the donor's death.
        \item \textbf{Relevant Article:} \\ \underline{Article 3 in Japanese Civil Code} \\(1) The enjoyment of private rights shall commence at birth.
        \\(2) Unless otherwise provided by applicable laws, regulations or treaties, foreign nationals shall enjoy private rights.
        \item \textbf{Gold Answer:} True
\end{itemize}
Naturally, this problem is considered as a textual entailment problem with given related articles in the Japanese Civil Code.
Textual entailment (\gls{TE}) or Natural language inference (\gls{NLI}) is a task in NLP. 
Given a termed text $t$ and hypothesis $h$, the models need to predict if $t$ entails $h$ ($t => h$) or not. The answers should be where $yes$ or $no$. 

Currently, there are several machine learning datasets for this problem. Samuel R. Bowman et al. \cite{bowman2015large} have introduced the Stanford Natural Language Inference (SNLI) Corpus that contains 570k pairs of written English sentences with three labels $entailment$, $contradiction$, and $neutral$. The Multi-Genre Natural Language Inference (MultiNLI) corpus \cite{williams2017broad} contains 433k written and spoken sentence pairs obtained by crowd-sourcing.

% \begin{table}[]
% \centering
% \caption{Data amount in COLIEE, SNLI and MultiNLI datasets\label{tab:data_amount}}
% \begin{tabular}{|l|c|}
% \hline
% \textbf{Dataset} & \textbf{Number } \\ \hline
% SNLI                        & 570,000                          \\ \hline
% MultiNLI                    & 433,000                          \\ \hline
% COLIEE                      & 716                          \\ \hline
% \end{tabular}
% \end{table}

% \begin{figure}[ht]
%   \centering
% \includegraphics[width=.8\linewidth]{figures/chap3/comparison.png}
%   \caption{Data Amount}
%   \label{fig:data_amount}
% \end{figure}

Compared to SNLI and MultiNLI, the total number of questions is extremely small. 
SNLI contains about 570,000 samples, MultiNLI contains 433,000 samples.
This number is 716 in COLIEE 2019's dataset.
In order to obtain good performance with this dataset, we propose a method to increase the number of training data for this task.

\subsection{Research Method}
As previously introduced, the goal of this task is to find the textual entailment between the statement and the articles to decide whether the statement is lawful or not.
In this section, we propose three techniques to increase the number of training data which are \textit{Problem Derivation}, \textit{Sample Splitting} and \textit{Data Augmentation}.

The given textual entailment problem can be stated as given a bar question $Q$ and all relevant articles $A$, the system needs to answer if $A$ entails $Q$ or not.
Solving this problem, the model first needs to have an efficient way to encode each pair of the statement and relevant article, after that learn to abstract the patterns for entailed and non-entailed relationships between them.
The biggest challenge in this problem is that the number of samples is limited in the samples given by the organizer.

\begin{figure}[ht]
  \centering
\includegraphics[width=.8\linewidth]{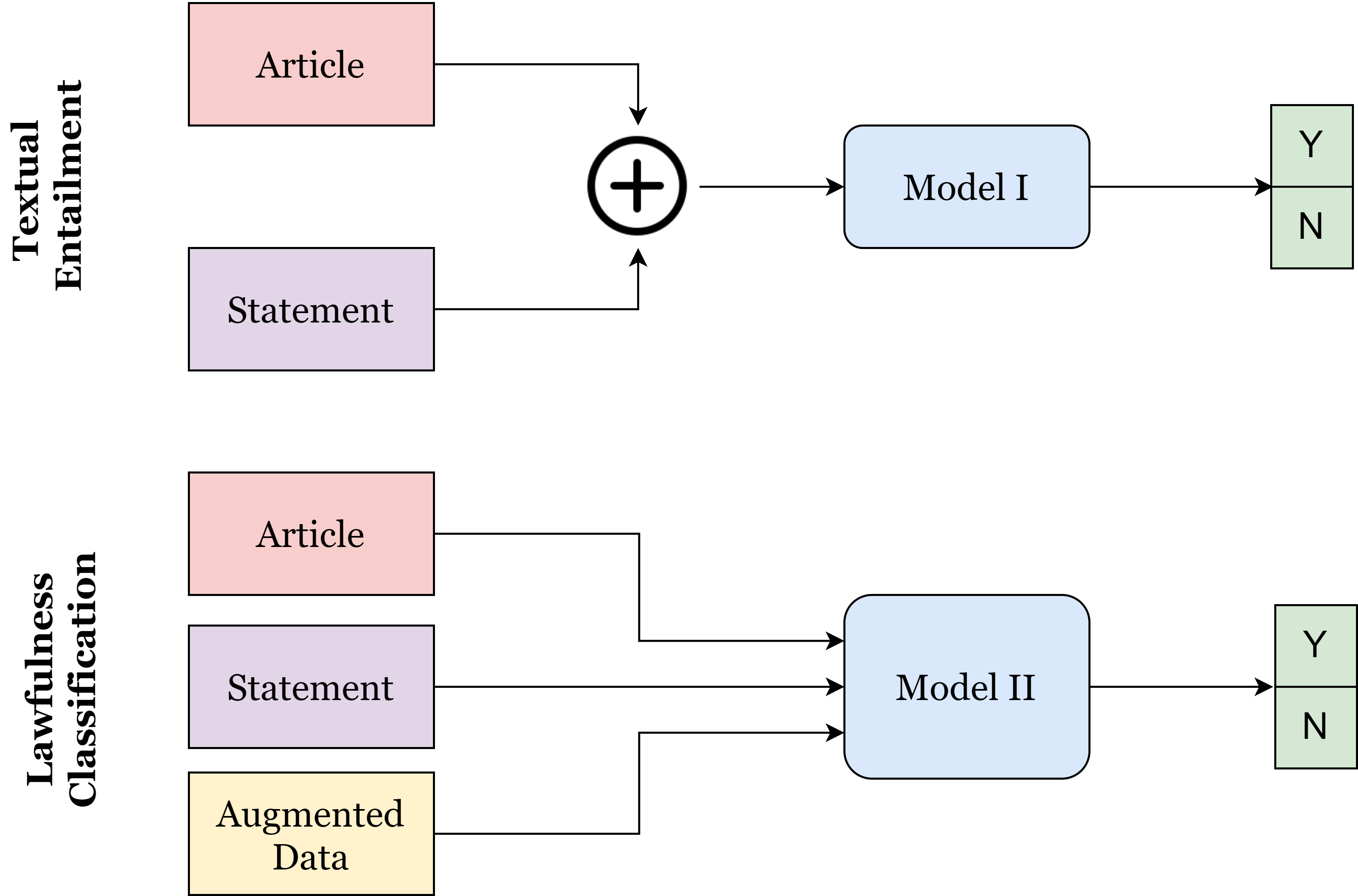}
  \caption{Problem derivation from textual entailment to lawfulness classification.\label{fig:problem_derivation}}
\end{figure}

Figure \ref{fig:problem_derivation} demonstate the original textual entailment problem and the new lawfulness classification problem.
In the original problem, the model needs to process the pairs of articles and statements as given by the competition organizer.
In the derived problem, the model treats each document as an independent input. 
This approach also allows us to create more augmented data for the learning process.

There are two parts in the data set, one contains articles in the Japanese Civil Code and the other contains statements in previous year competitions. For the lawfulness classification problem, the original data can be formatted as in the following rules:
\begin{itemize}
    \item All articles in Japan Civil Code are lawful
    \item All statements in previous year datasets that are entailed by articles in Japan Civil Code are lawful
    \item All statements in previous year datasets that are not entailed by articles in Japan Civil Code are not lawful.
\end{itemize}

The average length of the articles is 43.72 words, with a standard deviation of 59.58 while these measurements for the questions are 19.61 and 40.71. 
The sentences in the Japanese Civil Code are averagely longer and more variant than the sentences in the questions in previous year datasets.
From that observation, we split the articles to obtain more, shorter sentences with stabler length.

For each article, instead of adding the whole content into the data set, it is chunked into single statements. 
For example, article 329 (Order of Priority of General Statutory Liens) containing two statements could contribute 2 samples  to the data set as \textit{``(1) In cases where there is conflict among general statutory liens, [...] follow the order listed in each item of Article 306.''} and \textit{``(2) In cases where there is conflict between a general statutory lien [...] who received the benefit of the same.''}.
We assume that sub-articles in a lawful article are lawful.
After applying the article chunking method, we have the average value and standard deviation in length of text in the civil code data set is 23.34 and 39.65. 

Next, we use negation as the main data augmentation data method.
Negation of examples in the data set is obtained by a set of rules listed in Table \ref{tab:ch3_en_rules}. 
After this phase, we obtained a data set, of which the total number of examples is 4,748. 
This data set is then fed into the deep neural network.
\begin{table}[t]
  \caption{Rules applied for negation statement generation.}
  \label{tab:ch3_en_rules}
  \begin{center}

  \begin{tabular}{|c|l|}
    \hline
    \textbf{Original Statement}& \textbf{Negation Statement Generation}\\
    \hline
    contains $not$&Remove $not$ from original statement\\
    contains $shall$&Replace $shall$ with $shall\ not$\\
    contains $should$&Replace $should$ with $should\ not$\\
    contains $may$&Replace $may$ with $may\ not$\\
    contains $must$&Replace $must$ with $must\ not$\\\hline
    contains $is$&Replace $is$ with $is\ not$\\
    contains $are$&Replace $are$ with $are\ not$\\
    contains $will\ be$&Replace $will\ be$ with $will\ not\ be$\\\hline
    contains $can$&Replace $can$ with $cannot$\\
    contains $cannot$&Replace $cannot$ with $can$\\
    contains $with$&Replace $with$ with $without$\\
    contains $without$&Replace $without$ with $with$\\\hline
    contains $A$&Replace $A$ with $No$\\
    contains $An$&Replace $An$ with $No$\\
  \hline
\end{tabular}
\end{center}
\end{table}

\subsection{Experiments}

To understand the impact of the amount of data on the model's performance, we use the same model (\ie Bi-LSTM) as proposed by Borges et al.  \cite{borges2019combining} to run on the COLIEE dataset.
In the textual entailment approach, the representation of the statement and the article are concatenated right before the final MLP layer.
In the lawfulness classification approach, each input is fed individually to the network.

\begin{table}
\centering
\caption{Performance of BiLSTM in different datasets and approaches.\label{tab:bilstm}}
\begin{tabular}{|l|c|}
\hline
\textbf{Dataset} & \textbf{Accuracy} \\ \hline
BiLSTM on SNLI \cite{borges2019combining}                          & 83.3\%	                          \\ \hline
BiLSTM on MultiNLI \cite{borges2019combining}                      & 67.5\%                          \\ \hline
BiLSTM on COLIEE Textual Entailment                                & 49.0\%                          \\ \hline
BiLSTM on COLIEE Lawfulness Classification                         & 57.1\%                          \\ \hline
\end{tabular}
\end{table}

Table \ref{tab:bilstm} shows the experimental results in accuracy.
Bi-LSTM performs well on SNLI and MultiNLI with hundreds of thousand training samples.
This model obtains 83.3\% accuracy on SNLI and 67.5\% on MultiNLI. 
In the same problem on COLIEE 2019's dataset, this model can only achieve 49.0\%, which is even lower than a random guess (\ie 50\%).
With the problem of lawfulness classification, this model significantly improves its own performance by 8.1\% (from 49.0\% to 57.1\%).

From this result, we can see that the amount of data is an important feature to consider when applying deep learning techniques to a problem in general and a legal problem in particular.
Data processing does not essentially produce more information. However, with the approach in this section, we can feed more information into the deep learning model, thereby making the technique applicable to the given problem.
In addition, the trade-off of sentence number and sentence length in this case increases the advantage of deep learning models.
This approach we proposed at COLIEE 2019, continues to be utilized and reaps good results at COLIEE 2020 and COLIEE 2021 with more robust deep learning models, which will be introduced in Chapter 4.

\subsection{Discussions}
Similar to mathematics, an automated legal processing problem can have many approaches.
For each approach, we need to accept the trade-offs that come with it.
For the deep learning approach, with the ability to self-synthesize patterns through examples, these models need a certain amount of data.
Therefore, in this section, we propose a solution suitable for this characteristic.
By deriving the problem from textual entailment to lawfulness classification, we achieve data superiority for deep learning models based on 3 aspects: data amount, sentence length and augmentation.
Our hypothesis is that trading-off fewer long examples for more short examples might give the model advantage in this case.

From our experiments, we can see that with the same architecture, a lack of data may lead to a serious reduction of performance.
Problem derivation, data reformation and data augmentation are good options in the domain with limited data.
With the problem proposed in this section, we find that it is necessary to combine all three methods as a common approach to be effective. 
The reason could be that when we apply these methods to the law problem, we need to take into account the logical aspect of legal statements. 
For example, if chunking is not applied, we are negating only part of a multiple-item legal sentence, inverting the labels may produce incorrect results.
This solution was proposed by us at COLIEE 2019, then applied and succeeded at COLIEE 2020 and COLIEE 2021.
In addition, the observation in this section suggests we study the transfer learning techniques and pretrained models in legal text processing.

\newpage

\section{Impact of Model Architecture}
\label{sec:ch3_architecture}
\subsection{Introduction}
In addition to effective data representation and adequate data amount, model architecture is also a crucial factor that affects the performance in a deep legal problem.
Although pretrained models have proven effective on a wide range of NLP tasks, with some specific data domains such as law, using the default settings of these models does not provide maximum performance.
In this section, we find out that, in the specific domain of legal, a simple architecture designed with the understanding of the characteristic of data may surpass a bulky and powerful pretrained model.

This observation is obtained when we solve the problem of legal information retrieval (\ie given a query, the system needs to return the most appropriate legal articles related to the query).
First, we construct a Vietnamese real-life dataset for legal question-answering. 
The dataset contains the real legal questions from legal consultant websites and a Vietnamese law corpus.
Second, we experiment with different architectures and candidates on the dataset and observe the phenomenons.
Finally, we propose a simple architecture solely empowered by a convolutional neural network and attention mechanism which obtain state-of-the-art performance.

The control model we use in this work is XLM-RoBERTa \cite{conneau2019unsupervised}.
This model attests to the dominant of pretrained models with large scale in architecture as well as many pretraining tasks and datasets.
Being trained with more than 2.5 terabytes of multilingual data, XLM-RoBERTa gains a significant performance in both high-resource and low-resource languages.
However, in this work, the powerful model can not obtain a good result because its architecture is not designed to work with long text, an important characteristic of legal documents.

\subsection{Reseach Method}
The research in this section shows that, no matter how powerful, a model with incomplete information will make false predictions and is easily defeated by a simple model with enough information.
The weakness of Transformer based models in this domain is that they treat inputs as continuos sequences and truncate samples which surpass theirs max length limitation.
To design an effective architecture, we base on the key observation is that legal articles are mostly written in the form of a set of sentences.
Hence, we design a simple attentive convolution neural network to capture the signal from every sentence and combine them together by a global attention mechanism.

\vfill

\begin{figure}[ht]
  \centering
\includegraphics[width=.9\linewidth]{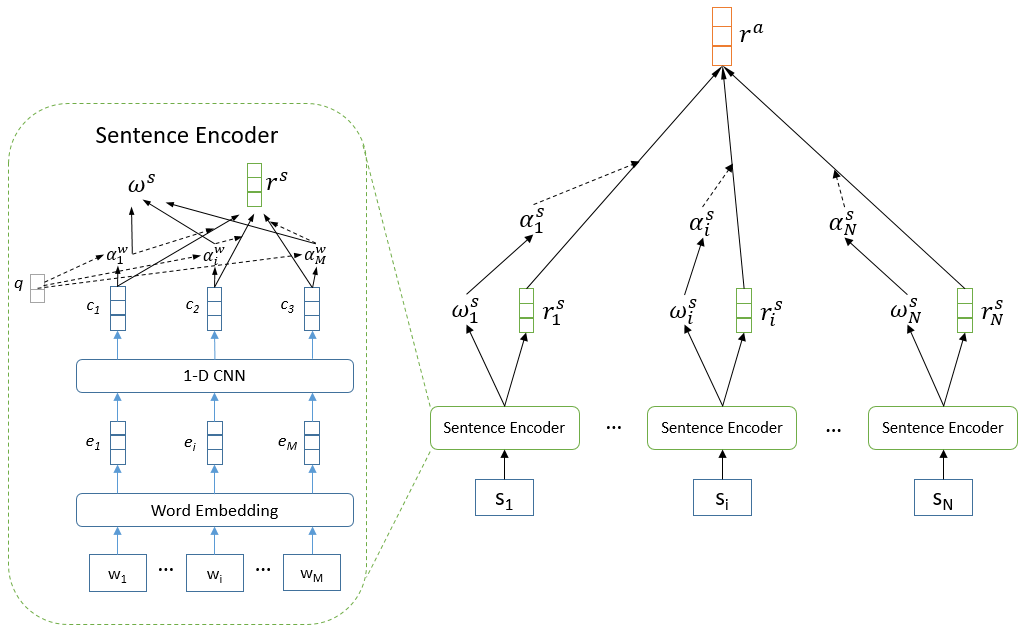}
  \caption{Attentive CNN architecture for long legal text processing.\label{fig:ch3_attentive_cnn}}
\end{figure}

Figure \ref{fig:ch3_attentive_cnn} demonstrates the architecture of the attentive convolution network.
This architecture is constructed in a hierarchical paradigm.
At the sentence level, each embedded word is processed by a convolutional layer.
After that, a query-based attention mechanism is applied to the outputs to get the representation of the whole sentence.
At the paragraph level, a sparse-max attention layer integrates the signal from the sentence representation to get the paragraph vector.

We train the proposed models using a negative sampling paradigm.
First, the query is encoded by the sentence encoder and the article is encoded with the paragraph encoder.
After that, we calculate the dot product between the two vectors as the similarity measurement.
The dot product cares about both angle and magnitude, its value range is from negative to positive infinity. 
We normalize the probability article $i$ is related to a given query following Formula \ref{eq:nomalize}.
We use the cross-entropy loss as the loss function for this approach.

\begin{equation}
\label{eq:nomalize}
p_{i}=\frac{\exp \left(\hat{y}_{i}^{+}\right)}{\exp \left(\hat{y}_{i}^{+}\right)+\sum_{j=1}^{K} \exp \left(\hat{y}_{i, j}^{-}\right)}
\end{equation}
where $\hat{y}_{i}^{+}$ and $\hat{y}_{i, j}^{-}$ are the probabilities that article $i$ and article $j$, which belongs to the negative set of article $i$, is related to the query respectively.

\subsection{Experiments}
The dataset contains two parts, which are a corpus of Vietnamese legal documents and a Vietnamese legal question-answering dataset containing queries coming along with their relevant articles.
There are in total 8,586 documents with 117,545 articles in the legal document corpus, 5,922 queries coming along with their relevant articles in the question-answering dataset. 
We use 10\% of the queries for testing and the remaining for the training and validation set.
To rank the candidates, we combine the lexical score from BM25 and the deep learning model's score.
Parameters of Attentive CNN are presented in Table \ref{tab:parameters_attentive_cnn}.

\begin{table}
\caption{\label{tab:parameters_attentive_cnn}
Value of parameters in Attentive CNN}
\begin{center}
\begin{tabular}{|l|c|}
\hline  \textbf{Parameter} & \textbf{Value} \\
\hline
Size of Word Embedding layer & 512 \\
Number of CNN filter & 512 \\
Size of attention query vector & 200 \\
Dropout rate & 0.2 \\
\hline
\end{tabular}
\end{center}
\end{table}

\begin{table}
\centering
\caption{Experimental Results on Vietnamese Legal Dataset}\label{tab:ch3_vi_res}
\begin{tabular}{|l|c|c|c|}
\hline
\textbf{Systems} & \textbf{Precision} & \textbf{Recall} & \textbf{F2} \\ \hline
BM25             & 0.2395                                  & 0.1966                               & 0.2006                           \\ \hline
XLM-RoBERTa      & 0.2395                                  & 0.1966                               & 0.2006                           \\ \hline
Attentive CNN    & 0.5919                                  & 0.4660                               & 0.4774                           \\ \hline
\end{tabular}
\end{table}

Table \ref{tab:ch3_vi_res} shows the experimental result on the Vietnamese legal dataset.
XLM-RoBERTa contributes no improvement compared to the naive lexical approach using BM25.
In contrast, Attentive CNN makes a big gap between the two control systems, overperforms them by a 0.2768 F2-score.
This is a very interesting result because pretrained language models are often expected to yield better results than simple, unpretrained models.
This result made us curious as to what makes the powerful model XLM-RoBERTa perform worse than Attentive CNN.
Looking for the cause of this phenomenon, we discovered that the length of the article in our dataset is up to 253K characters, far beyond the encoding capacity of XLM-RoBERTa (514 tokens).
This shows that lengthy content is an existential problem for legal documents and a suitable architecture for deep learning models is needed to solve this problem.

\subsection{Discussions}
In this section, by proposing and comparing a simple attentive convolutional neural network for a legal retrieval information problem, we obtain some important observations.
First, besides data amount and data representation, model architecture is crucial to obtain a good performance in a deep legal problem.
Second, using an attention mechanism to integrate the signals from elements of a long legal article can help the model to keep the full information without truncation. 
This observation is the basis for Paraformer, which is introduced in Chapter \ref{chap:lm}.
Third, a simple architecture does not always perform worse than a complicated and bulky one especially when it can make better use of information from the data.

\newpage
\section{Summary of Chapter}

In this chapter, we answer the question of what factors can impact an end-to-end model in deep legal processing.
We examine in detail the factors that can affect deep learning models such as data representation, data amount, and model architecture. 
Better data in both quality and quantity is the requirement for a good performance in the legal domain.
Besides, an appropriate architecture is also important for a model to access full information from the training data and perform better on a specific task.
The observations from this chapter are the basis in exploring the knowledge about deep legal processing. 

In the data representation aspect, we provide an insight into how embeddings perform across the legal domain.
We propose two quantitative scales, LVC and LECA, for this purpose.
In addition, our visualization method represents the position of word vectors in three-dimensional space.
From there we can intuitively understand the existence of these vectors and their significance for the performance of the whole system.
Not only based on speculation but also from our experimental results, we draw that pretrained embeddings in the law domain tend to represent better legal concepts than the embeddings pretrained with the general documents.

In terms of data amount, we experimentally prove that lack of data is a problem that directly affects the performance of deep learning models.
With the same architecture, a deep learning model can give excellent or bad results depending on the amount of data it is trained on.
Although this problem has been pointed out in many previous works, our conclusion once again confirms its existence in the legal domain.
Within our scope, we offer specific solutions to increase data volume through problem derivation, data reformation and data augmentation.

In analyzing the architecture of the model, we prove that there is no free lunch for every problem.
Through experimentation, we show that a simple architecture that properly models the nature of data can have better results than a bulky, pretrained architecture trained with a massive amount of data.
This finding is the premise for us to design improvements to pretrained language models so that they can achieve the best performance on legal domains.

\chapter{Pretrained Language Models for Deep Legal Processing}
\label{chap:lm}
\section{Overview}

This chapter dedicates to introducing our works related to pretrained language models in deep legal processing.
There are three models mentioned in this chapter.
The first model is BERTLaw, a language model trained on the legal corpus from scratch and finetuned for the question-answering task.
This model achieves impressive state-of-the-art results in Task 4, COLIEE 2020.
Second, we introduce Paralaw Nets, a family of language models pretrained by legal multilingual resources, in which the NMSP model becomes the best system in Task 5, COLIEE 2021.
The third model is Paraformer, an attentive architecture leveraging the power of language models and succeeding in legal retrieval problems.
This model is the extension of the model introduced in Section \ref{sec:ch3_architecture}.
\section{Legal Contextual Embedding}

\subsection{Introduction}
As introduced in Chapter \ref{chap:background} law documents are written in a special sublanguage.
This sublanguage is very different from the language we use in daily life.
Legal English is used in drafting contracts, terms of service, regulations and other legal documents.
As a result, a language model pretrained with data in the general domain may face trouble in tasks in the legal domain.

In this section, we introduce BERTLaw, a language model pretrained from scratch with a large legal corpus.
Because of the superior legal vocabulary understanding, this model achieves impressive state-of-the-art results in Task 4, COLIEE 2020.
This idea also attests to the observation in Chapter \ref{chap:investigate}, with the same architecture, a better dataset could lead to better performance.

BERT \cite{devlin2018bert} is one of the first pretrained language models based on Transformer achieving state-of-the-art results in a wide range of NLP tasks.
The authors of this model propose two pretraining tasks which are masked language model (\gls{MLM}) and next sentence prediction (\gls{NSP}). 
In the MLM task, the model needs to make predictions to guess the masked words in a sentence and in the NSP task, the model needs to predict whether the two sentences are consecutive in a paragraph.

\subsection{Research Method}

We construct BERTLaw, a model pretrained with a large amount of legal text.
A legal word stands alone can not convey any meaning. In a legal statement, there are both common words and legal words.
As a result, we choose legal cases as our training data.
In legal cases, everyday vocabulary and legal vocabulary appear together.
This is important to get a good pretrained language model from scratch.
We use 8.2 million sentences of American legal case data to pretrain our model.

Constructing the vocabulary of BERTLaw, we apply the method of SentencePieces \cite{kudo2018sentencepiece}.
A word can be split into multiple subwords for the most efficient representation.
With subword representation, the problem of \gls{OOV} could be mitigated.
For example, suppose that \textit{reconvention} does not appear in the vocabulary, it can be represented by 4 subwords as \textit{rec}, \textit{on}, \textit{vent} and \textit{ion}.
Note that, in embedding, mapping a word into an index without appropriate weights, the mapping has little meaning.
That's why we need to pretrain the model to learn the relationship between subwords in the vocabulary.

\begin{figure}[ht]
  \centering
\includegraphics[width=.6\linewidth]{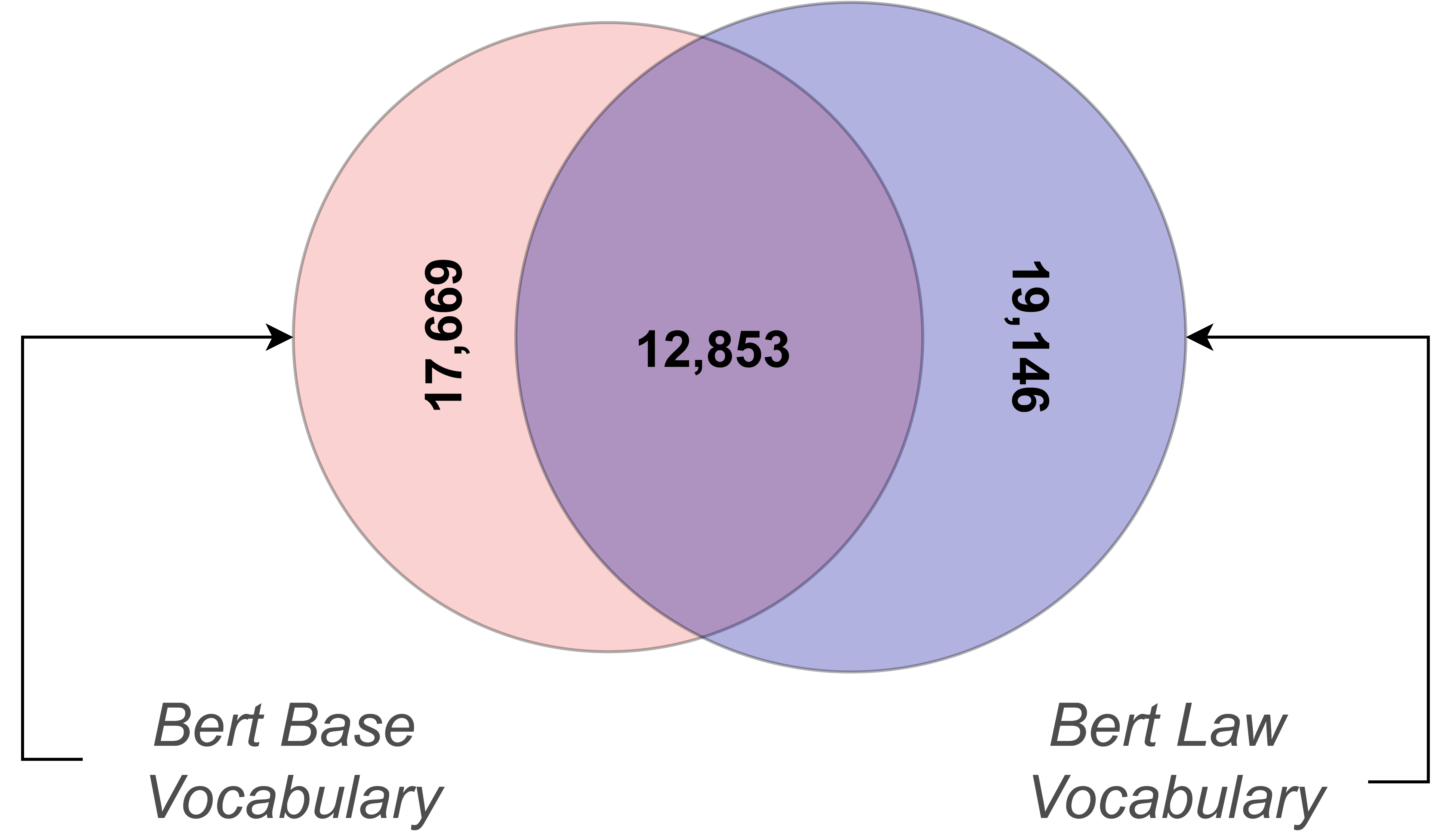}
  \caption{BERTLaw vocabulary compared to Google's BERT Base vocabulary.}
  \label{fig:bertlaw_vocab}
\end{figure}

Figure \ref{fig:bertlaw_vocab} demonstrates the comparison between the vocabulary of BERTLaw and Google's BERT Base vocabulary.
The overlapped words are more than half of the vocabulary of each model.
The vocabulary of BERTLaw is slightly bigger than which of BERT Base.
We pretrain our BERTLaw using Google's \gls{TPU} on the corpus until the loss value on the MLM and NSP tasks stop reducing.

\subsection{Experiments}
We verify the effectiveness of our pretrained language model on the COLIEE 2020's task 4 dataset.
This task is designed to verify the ability of the paralegals in answering the legal questions.
Given a statement, the answer should be whether this statement is true or false.
Our approach for this problem is the same as introduced in Section \ref{sec:ch3_data_amount}.
The model is finetuned on the augmented data and needs to classify the lawfulness of an input.

After data augmentation, we have in total 5,000 samples, 90\% for training and 10\% for validation. 
The test set provided by the organizer contains 112 testing samples.
Since we approach the task as a lawfulness classification problem, we do not need to deal with long input, the max length is set to 128.

We obtain the best configuration after 5 epochs of finetuning.
Table \ref{tab:ch4_legalemb_result} demonstrates the experimental results on the validation set and the official test set.
Our BERTLaw outperforms Google's BERT Base by 4\% on the validation set and 16\% on the test set.
This achievement brings us to the first position in the leaderboard of the competition in 2020\footnote{https://sites.ualberta.ca/~rabelo/COLIEE2020/task4\_res.html}.

\begin{table}
  \caption{BERTLaw and BERT Base Performance.}
  \label{tab:ch4_legalemb_result}
  \centering
  \begin{tabular}{|l|c|c|}
    \hline
    Model&Validation Accuracy&Test Accuracy\\
    \hline
    BERT Base          &0.7784&0.5625\\
    BERT Law         &0.8168&0.7232\\
  \hline
\end{tabular}
\end{table}

Looking deeper into the vocabulary of the two models, we find interesting differences.
Table \ref{tab:bertlaw_vocab_examples} shows some examples that appear in BERT Law vocabulary, not in BERT Base vocabulary, which can impact the difference in performance in legal-related tasks.
The subword  \textit{legal} does not appear in BERT Base vocabulary, which can make this model's vocabulary comprehension flawed in the legal domain.
In addition, with the LVC and LECA measurements presented in Chapter 3 to evaluate embeddings in the legal domain, BERT Law is superior to BERT Base.

To better understand model behavior, we observe how they handle input queries.
Here's how two models tokenize the same query ``A contract of sales concluded by a minor may not be rescinded if it relates to daily life, even in cases the consent of the parental authority is not obtained":
\begin{itemize}
    \item \textbf{BERT Base:} a contract of sales concluded by a minor may not be \textbf{res \#\#cin \#\#ded} if it relates to daily life, even in cases the consent of the parental authority is not obtained
    \item \textbf{BERT Law:} a contract of sales \textbf{conclude \#\#d} by a minor may not be \mbox{\textbf{rescind \#\#ed}} if it \textbf{relate \#\#s} to daily life [UNK] even in cases the consent of the parental authority is not \textbf{obtain \#\#ed}
\end{itemize}
We bold the words that the models must use subwords to represent. In this example, BERT Base must use fewer subwords than BERT Law. Words like ``concluded'', ``obtained'', ``related'', BERT Law must separate the original word and the ending, BERT Base keeps the same words. However, with the word ``rescinded'', BERT Base must use 3 subwords to represent, while BERT Law separates it only into ``rescind'' and the ``ed'' suffix. ``Rescind'' is a legal term and the representation of BERT Law proves its legal embedding capacity.

\begin{table*}
  \caption{Examples that appear in BERT Law Vocabulary, not in BERT Base Vocabulary. Subwords start with $\sharp\sharp$.}
  \label{tab:bertlaw_vocab_examples}
  \centering
  \footnotesize
  \begin{tabular}{|p{2.5cm}|p{10cm}|}
    \hline
    \textbf{Token}& \textbf{Explanation}\\
    \hline
    $\sharp\sharp$legal&Wordpiece in words containing ``legal" (e.g. illegal, legally, legality, legalization)\\\hline
    
    contravention&An act which violates the law, a treaty or an agreement that the party has made. \cite{bouvier1870law}\\\hline
    
    construe&To determine the meaning of the words of a written document, statute, or legal decision, based upon rules of legal interpretation as well as normal meanings. \cite{bouvier1870law}\\\hline

    demurrer&An assertion by the defendant that although the facts alleged by the plaintiff in the complaint may be true, they do not entitle the plaintiff to prevail in the lawsuit. \cite{bouvier1870law}\\\hline
    
    depose&To make a deposition; to give evidence in the shape of a deposition; to make statements that are written down and sworn to; to give testimony that is reduced to writing by a duly qualified officer and sworn to by the deponent. \cite{bouvier1870law}\\\hline

    guardianship& The power or protective authority given by law, and imposed on an individual who is free and in the enjoyment of his rights, over one whose weakness on account of his age, renders him unable to protect himself. \cite{bouvier1870law}\\\hline
    
    infringe&To transgress or exceed the limits of; violate: infringe a contract; infringe a patent. \cite{morris1969american}\\\hline
    
    malfeasance&The commission of an act that is unequivocally illegal or completely wrongful. \cite{bouvier1870law}\\\hline
    
    misdemeanor&Offenses lower than felonies and generally those punishable by fine, penalty, forfeiture, or imprisonment other than in a penitentiary. \cite{morris1969american}\\\hline
    
    reimburse&To repay (money spent); refund. \cite{morris1969american}\\\hline
    
    renounce& To give up a right; for example, an executor may renounce the right of administering the estate of the testator; a widow the right to administer to her intestate husband's estate. \cite{morris1969american}\\\hline
    
    rescind& To declare a contract void—of no legal force or binding effect—from its inception and thereby restore the parties to the positions they would have occupied had no contract ever been made. \cite{morris1969american}\\\hline
    
    rescission&The termination of a contract by mutual agreement or as a result of fraud or some legal defect. \cite{morris1969american}\\\hline
    
    revoke&To invalidate or cause to no longer be in effect, as by voiding or canceling. \cite{morris1969american}\\\hline

    tort& a civil wrong. Tortious liability arises from the breach of a duty fixed by law; this duty is towards persons generally and its breach is redressable by an action for unliquidated damages. \cite{morris1969american}\\\hline
    tortious&Wrongful; conduct of such character as to subject the actor to civil liability under Tort Law. \cite{morris1969american}\\

  \hline
\end{tabular}
\end{table*}

\subsection{Discussions}
This section aims to verify that the issue of sublanguage in law happens with humans and also language models.
We propose to construct the vocabulary and pretrain a language model from scratch using a large legal corpus \ie BERTLaw.
Our contribution is not in suggesting a new method but in adapting an existing method in the general domain to the legal domain and analyzing the reasons for its success.
The experimental result proves that the proposed approach is reasonable.
Although BERTLaw is a single model pretrained in the legal domain, the sublanguage problem may occur in other fields. Therefore, building such pretrained language models is a suggestion to obtain the optimal performance for this kind of model.

\newpage
\section{Legal Multilingual Capacity}

\subsection{Introduction}
In this section, we propose to use the multilingual information in the aligned translation pair as the pretraining resource for pretrained language models.
The idea of this work starts from the observation that a translation of a sentence into another language could be used as information to reduce the ambiguity in that sentence.
Over more, this information may help the language model to obtain better positions representing the words in the vector space.

\begin{figure}[ht]
  \centering
\includegraphics[width=.8\linewidth]{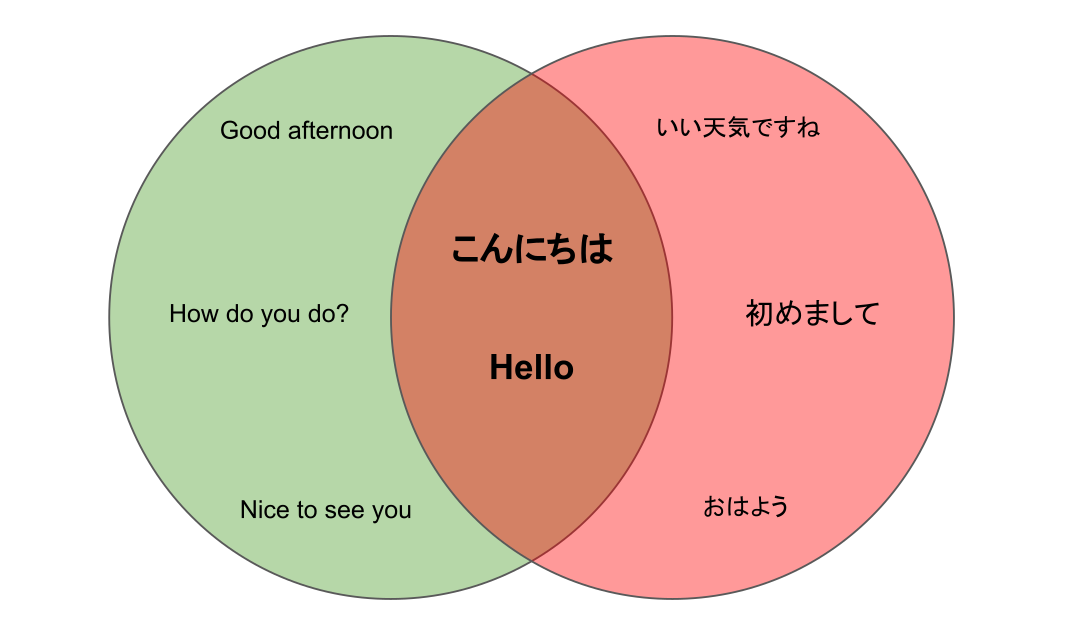}
  \caption{An example of a translation where a word can be translated by multiple candidates, depending on different contexts.}
  \label{fig:ch4_translation}
\end{figure}

As can be seen Figure \ref{fig:ch4_translation}, there are multiple ways to translate \begin{CJK*}{UTF8}{min}こんにちは\end{CJK*} to English and multiple ways to translate ``Hello'' to Japanese.
When these two words are aligned together, we know exactly that the speaker wants to use a word for the greeting purpose.
This example presents to us that a good translation can be used to make the meaning clearer.
Fortunately, there are good translations available in the legal domain.

\subsection{Research Method}

From the observation about the meaning of translation, we design two pretraining tasks to train multilingual pretrained language models.
The first task is Next Foreign Sentence Prediction (\gls{NFSP}), given a pair of sentences in different languages, the model needs to predict whether the two sentences are consecutive or not.
As a simple example, we have a pair of sentences as ``Hello! How are you?'' and their translations \begin{CJK*}{UTF8}{min}``こんにちは。お元気ですか？''\end{CJK*}, examples can be constructed as follow:

\begin{CJK*}{UTF8}{min}
\begin{itemize}
    \item Hello. お元気ですか？

    \item こんにちは。How are you?
\end{itemize}
\end{CJK*}

The second task is Neighbor Multilingual Sentence Prediction (\gls{NMSP}).
In this task, the model needs to predict not only whether the two multilingual sentences are consecutive but also which sentence is before, which sentence is after.
The training data for this task accepts pairs of sentences in the same language.
We apply a random negative sampling to construct the samples in which two sentences are not semantic consecutive.

To obtain these novel pretrained language models, we use the base architecture of BERT multilingual \cite{devlin2018bert}.
We reuse the vocabulary of the case-sensitive configurations. 
After that, we construct the dataset as described above with English-Japanese bilingual data crawled from the Japanese Law Translation website\footnote{https://www.japaneselawtranslation.go.jp}, we obtain 239,000 samples for NFSP and 718,000 samples for NMSP. 
We use 10\% of the dataset as the validation set.
We keep training the models until the performance stops to increase in the validation set. Table~\ref{tab:ch4_pretraining_paramters} shows the hyperparameters and the performances in pretraining the models.

\begin{table*}
\caption{Hyperparameters and performances in pretraining the models}
\centering
\begin{tabular}{|l|c|c|}
\hline
\textbf{Hyper Parameter / Performance} & \textbf{NFSP} & \textbf{NMSP} \\ \hline
Max Length                             & 512           & 512           \\ \hline
Batch Size                             & 16            & 16            \\ \hline
Number of Batches                      & 24,000        & 320,000       \\ \hline
Validation Accuracy                    & 94.4\%        & 88.0\%          \\ \hline
\end{tabular}

\label{tab:ch4_pretraining_paramters}
\end{table*}

\subsection{Experiments}

We finetune our multilingual language models for the question answering task in COLIEE 2021.
The general approach is similar as introduced in Section \ref{sec:ch3_data_amount}.
However, with the multilingual models, we can double our augmented data by using both English and Japanese datasets provided by the organizer.
Besides the English negation rules in Table \ref{tab:ch3_en_rules}, we introduce the Japanese negation rules in Table \ref{tab:ch4_jp_rules}.
For contextual embeddings, the input text representation of the model depends on its entire weights and not just on the embedding layer as for word embeddings.
For that reason, we can augment more  training data in both languages even though the legal question answering task is not multilingual.

\begin{table*}[t]
  \caption{Rules applied for negation statement generation in Japanese}
  \label{tab:ch4_jp_rules}
  \begin{center}
      
  \small
  \begin{tabular}{|l|l|}
    \hline
    \textbf{Original Statement}& \textbf{Negation Statement Generation}\\
    \hline
    \begin{CJK*}{UTF8}{min}ません\end{CJK*}&\begin{CJK*}{UTF8}{min}ません\end{CJK*} $\rightarrow$ \begin{CJK*}{UTF8}{min}ます\end{CJK*}\\
    \begin{CJK*}{UTF8}{min}できる\end{CJK*}&\begin{CJK*}{UTF8}{min}できる\end{CJK*} $\rightarrow$ \begin{CJK*}{UTF8}{min}できない\end{CJK*}\\
    \begin{CJK*}{UTF8}{min}できない\end{CJK*}&\begin{CJK*}{UTF8}{min}できない\end{CJK*} $\rightarrow$ \begin{CJK*}{UTF8}{min}できる\end{CJK*}\\
    \begin{CJK*}{UTF8}{min}した\end{CJK*}&\begin{CJK*}{UTF8}{min}した\end{CJK*} $\rightarrow$ \begin{CJK*}{UTF8}{min}しなかった\end{CJK*}\\
    \begin{CJK*}{UTF8}{min}でない\end{CJK*}&\begin{CJK*}{UTF8}{min}でない\end{CJK*} $\rightarrow$ \begin{CJK*}{UTF8}{min}である\end{CJK*}\\
    \begin{CJK*}{UTF8}{min}できた\end{CJK*}&\begin{CJK*}{UTF8}{min}できた\end{CJK*} $\rightarrow$ \begin{CJK*}{UTF8}{min}できなかった\end{CJK*}\\
    \begin{CJK*}{UTF8}{min}させる\end{CJK*}&\begin{CJK*}{UTF8}{min}させる\end{CJK*} $\rightarrow$ \begin{CJK*}{UTF8}{min}させない\end{CJK*}\\
    \begin{CJK*}{UTF8}{min}ている\end{CJK*}&\begin{CJK*}{UTF8}{min}ている\end{CJK*} $\rightarrow$ \begin{CJK*}{UTF8}{min}ていない\end{CJK*}\\
    \begin{CJK*}{UTF8}{min}がない\end{CJK*}&\begin{CJK*}{UTF8}{min}がない\end{CJK*} $\rightarrow$ \begin{CJK*}{UTF8}{min}がある\end{CJK*}\\
    \begin{CJK*}{UTF8}{min}ではない\end{CJK*}&\begin{CJK*}{UTF8}{min}ではない\end{CJK*} $\rightarrow$ \begin{CJK*}{UTF8}{min}ではある\end{CJK*}\\
    \begin{CJK*}{UTF8}{min}ことがある\end{CJK*}&\begin{CJK*}{UTF8}{min}ことがある\end{CJK*} $\rightarrow$ \begin{CJK*}{UTF8}{min}ことがない\end{CJK*}\\
    \begin{CJK*}{UTF8}{min}しなければならない\end{CJK*}& \begin{CJK*}{UTF8}{min}しなければならない\end{CJK*} $\rightarrow$ \begin{CJK*}{UTF8}{min}してはいけません\end{CJK*}\\
    \begin{CJK*}{UTF8}{min}ならない\end{CJK*}&\begin{CJK*}{UTF8}{min}ならない\end{CJK*} $\rightarrow$ \begin{CJK*}{UTF8}{min}なる\end{CJK*}\\
    
  \hline
\end{tabular}
\end{center}
\end{table*}

After augmenting the data, we obtain 7,000 samples.
We use 700 samples for validation, the rest is for training.
Because of the complexity in Japanese data, we first train 3 epochs with data augmented with the first three Japanese negation rules before training on the data augmented with the whole rules.
On the validation test, NFSP achieves 71.0\% accuracy while NMSP achieves 79.5\%.
We also train the vanilla BERT Multilingual in this task, the model only achieves 64.1\% accuracy.
This again confirms the usefulness of multilingual data for language models.

\begin{table*}
\caption{Result of final runs on the test set}
\centering
\begin{tabular}{|l|l|l|r|}
\hline
\textbf{Team} & \textbf{Run ID}                     & \textbf{Correct} & \textbf{Accuracy}     \\ \hline
              & BaseLine                            & No 43/All 81     & 0.5309                \\ \hline
JNLP    & NFSP                                & 49                & 0.6049 \\ \hline
UA            & UA\_parser                          & 46               & 0.5679                \\ \hline
JNLP    & NMSP                          & 45               & 0.5556          \\ \hline
UA            & UA\_dl                              & 45               & 0.5556                \\ \hline
TR            & TRDistillRoberta                    & 44               & 0.5432                \\ \hline
KIS           & KIS\_2                              & 41               & 0.5062                \\ \hline
KIS           & KIS\_3                              & 41               & 0.5062                \\ \hline
UA            & UA\_elmo                            & 40               & 0.4938                \\ \hline
JNLP        & BERT Multilingual                   & 38               & 0.4691          \\ \hline
KIS           & KIS\_1                              & 35               & 0.4321                \\ \hline
TR            & TRGPT3Ada                           & 35               & 0.4321                \\ \hline
TR            & TRGPT3Davinci                       & 35               & 0.4321                \\ \hline
\end{tabular}
\label{tab:ch4_multilingual_final_runs}

\end{table*}

Table \ref{tab:ch4_multilingual_final_runs} shows the performance of different systems on the formal test set in COLIEE 2021.
We are surprised that the NFSP overperforms NMSP on the test set and achieves state-of-the-art performance, nearly 4\% higher than the second-best system.
On the test set, BERT Multilingual performs not very well, its performance 0.4691 even lower than a random guess (\ie 0.5 accuracy).

\subsection{Discussions}

This section proposes and discusses using sentence-level cross-lingual information to pretrained language model.
We introduce two pretraining tasks corresponding with two proposed models namely NFSP and NMSP.
The models achieve impressive results in our validation set as well as the formal COLIEE 2021's test set.
These results confirm the effectiveness of using translation as a training resource for language model pretraining.
The finding in this section can also be applied to other domains in which good translations are available.

\newpage
\section{Legal Structural Representation}

\subsection{Introduction}
In Section \ref{sec:ch3_architecture}, we point out the limitation of the vanilla architecture of current Transformer based language models when working with legal data.
Even so, language models have undeniable power when training data becomes scarce.
In this section, we propose a novel architecture that combines the strength of both pretrained language models and the advantage of legal structural representation by using the global attention mechanism.

The length of legal documents is always a challenge for automated processing systems.
A longer document means more signals need to be processed.
More seriously, meaningful information may be in the later parts, which has been truncated due to exceeding the maximum length of the architecture.
In a field that prioritizes correctness like law, solving this problem is a prerequisite for practical applications in the future.

The following example demonstrates how a model views an article exceeding its maximum length. 
The grey parts demonstrate the truncated content that the model does not process. 
The content could be trimmed at any position of the text depending on the maximum length configuration, which makes the information incomplete and incomprehensible.

\vfill

\begin{quotation}
\textbf{Article 330}

(1) In cases where there is conflict among special statutory liens with respect to the same movables, the order of priority shall follow the order listed below. In such cases, if there are two or more preservers with respect to the statutory liens for preservation of movables listed in item (ii), a new preserver shall prevail over previous preservers.

(i) Statutory liens for leases of immovable properties, lodging at hotels and transportation;

(ii) Statutory liens for the preservation of movables; and

(iii) Statutory liens for the sale of \color{gray!40} movables, the supply of seed or fertilizer, agricultural labor and industrial labor.

(2) In the cases provided for in the preceding paragraph, if a holder of a statutory lien ranked first knew at the time he/she acquired that claim of the existence of a holder of a statutory lien of the second or third rank, he/she cannot exercise his/her rights against those persons. The same shall likewise apply against persons who preserved Things on behalf of the holder of a statutory lien of the first rank.

(3) Regarding fruits, the first rank shall belong to persons who engage in agricultural labor, the second rand shall belong to persons who supply seed or fertilizer, and the third rank shall belong to lessors of land.

\end{quotation}

In a specific legal retrieval case, if the query is related to the order of priority of statutory liens over fruit, the model may not consider this article as a good candidate.
The reason is simply that it does not have the information of fruit in the first part of the content.

\subsection{Research Method}
To overcome the problem of long legal documents and still take advantage of the pretrained language model, we propose an architecture namely Paraformer, which uses a transformer network to encode each sentence and integrate the signals to obtain the representation for the paragraph level.
With this architecture, we can input the whole long article into the model as long as the physical memory condition is adequate.

\begin{figure}
    
    \centering
    \includegraphics[width=.7\textwidth]{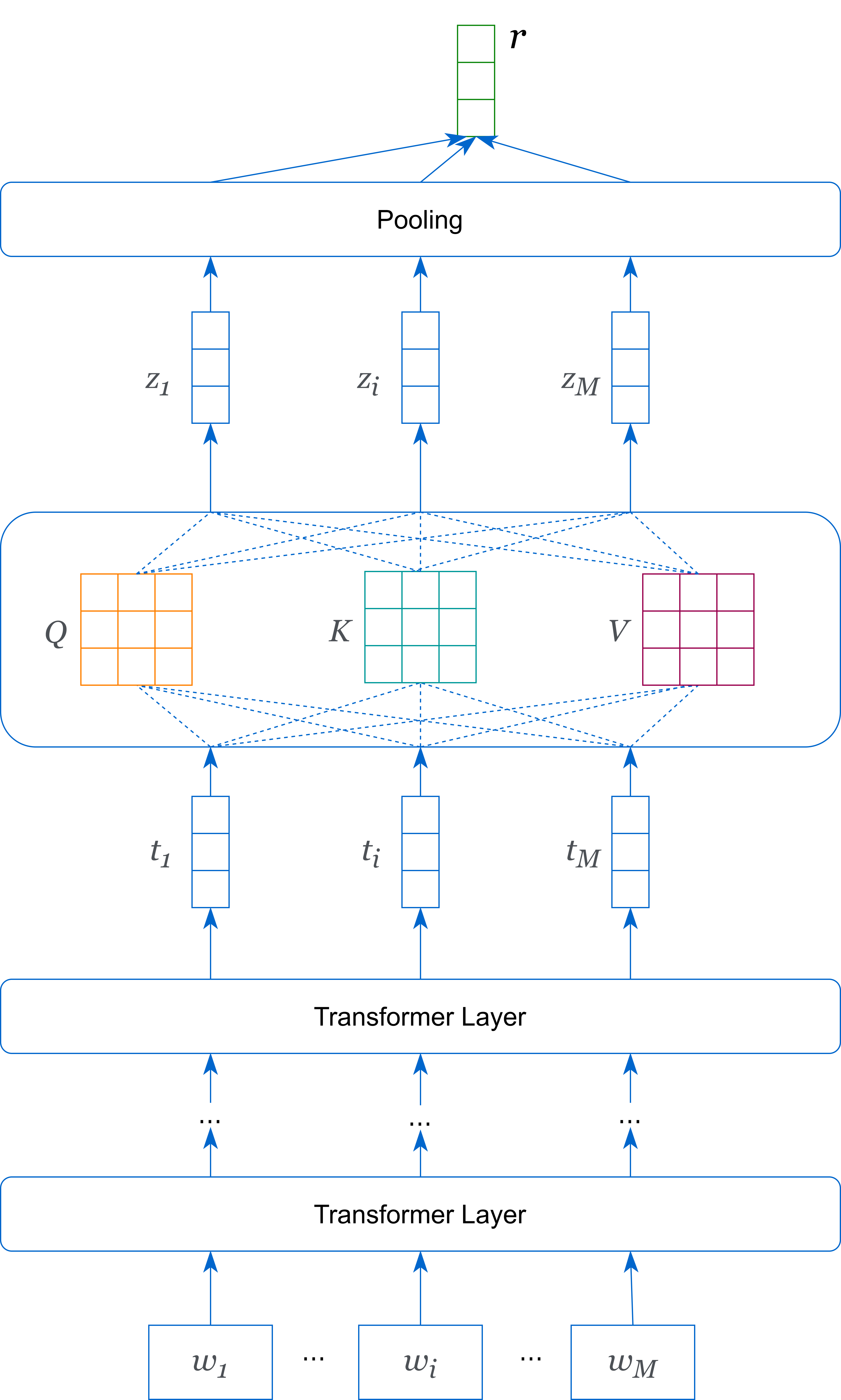}
    \caption{Paraformer's architecture of sentence encoder.}
    \label{ch4_sentenc}
\end{figure}

Figure \ref{ch4_sentenc} demonstrates the architecture of the sentence encoder, which encodes an input sentence into a vector representation.
Suppose that the language model contains $L$ layers, the input has $M$ tokens, after the layer $L-1^{th}$, we have the output $T=(t_1, t_2, . .., t_M)$.
At the last layer, $T$ is multiplied by the attention matrices $Q$, $K$, $V$ to get the corresponding attention values.
These attention values are combined with the softmax function (Equation~\ref{eq:softmax}) to get the corresponding output at the last transformer layer $Z=(z_1, z_2, ..., z_M)$.

\begin{equation}
    \label{eq:softmax}
    Z=softmax(\frac{Q \times K^{\top}}{\sqrt{d}}V)
\end{equation}

An average pooling is applied on the $Z$ to get the final representation of the sentence:

\begin{equation}
    r=\frac{1}{M}\sum_{i=1}^{M}z_i
\end{equation}

The advantage of using the pretrained sentence encoder is that this module has been trained on a large corpus so it requires less data for the fine-tuning phase.

\begin{figure}
    
    \centering
    \includegraphics[width=.7\textwidth]{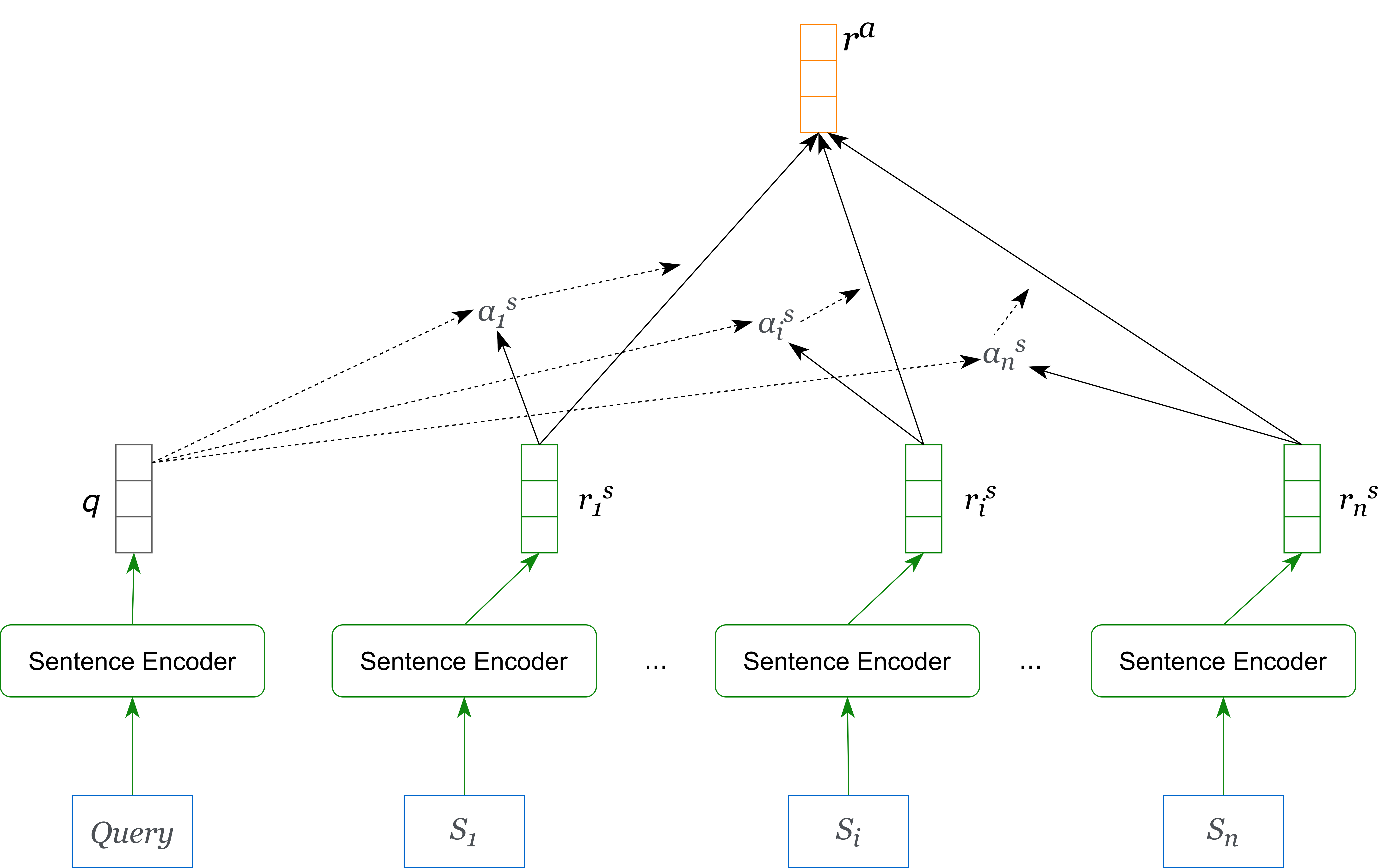}
    \caption{Paraformer's architecture of paragraph encoder.}
    \label{ch4_paraenc}
\end{figure}

Figure \ref{ch4_paraenc} shows the architecture of the paragraph encoder, which combines the signal from the sentence encoders into the final representation. First, both the query and the article's sentences are encoded using the sentence encoder to get $q$ and $r_i^s$ vectors respectively.
Then, with general attention, the representation of an article with a query is calculated by the Formula \ref{eq:a_i_s}, \ref{eq:alpha_i_s}, and \ref{eq:r_a}.

\begin{equation}
\label{eq:a_i_s}
    a_i^s=q^{T} \tanh \left(A \times r_i^s+b\right)
\end{equation}

\begin{equation}
\label{eq:alpha_i_s}
    \alpha_{i}^{s}=\operatorname{sparsemax}\left(a_{i}^{s}\right)^{*}
\end{equation}

\begin{equation}
\label{eq:r_a}
    r^a=\sum_{i=1}^{M} \alpha_i^s r_i^s
\end{equation}

\noindent In which, $\alpha_{i}^{s}$ be the attention weights and $r^a$ be the final representation.
\subsection{Experiments}
The experimental data we use in this section is COLIEE 2021. 
This dataset is drawn from Japanese Legal Bar exams. The scope of the questions is limited to the Japanese Civil Code.
Total samples in the training set and validation set are 806, the number of samples for testing is 81.
The data is very valuable for us to experiment with our proposed approaches.
Our validation set and test set are the formal evaluation data in COLIEE 2020 (65 queries) and COLIEE 2021 (81 queries), respectively, the rest is for training.

The retrieval process is done in two steps.
The first step is choosing the top $N$ articles based on lexical matching to filter out clearly unrelated articles.
This number can be different in the training phase and the prediction phase, we also have experiments to find the optimal value.
The second step is reranking the article using the deep learning models.

We construct a relevance scoring function as in Equation \ref{eq:scoring_function}.
\begin{equation}
\label{eq:scoring_function}
S_{f}=\alpha \cdot S_{l}+(1-\alpha) \cdot S_{s}
\end{equation}
where $ S_{f}$ is the final score calculated from the lexical score $S_{l}$, and the semantic score $ S_{s}$ is given by the deep learning model. $\alpha \in [0, 1]$ is the hyperparameter determining the weights of the two scores. 
We apply grid-search to obtain the optimal value of $\alpha$. 

We use XLM-RoBERTa as the backbone of our Paraformer.
Table \ref{tab:ch4_coliee_res} is the performance comparison between Attentive CNN proposed in Section \ref{sec:ch3_architecture}, vanilla XLM-RoBERTa and Paraformer without the ensembling with the lexical score. 
With Macro-F2@1 as the main evaluation metric, in general, Paraformer surpasses XLM-RoBERTa and Attentive CNN in both datasets.
Its F2 scores in the English dataset and Japanese dataset are $0.3498$, and $0.3182$ respectively.
Interestingly, from this result, pretrained models tend to achieve higher performance than the non-pretrained model (\ie Attentive CNN).

Table \ref{tab:ch4_performance_coliee} presents the final performance on the test set after ensembling with the lexical score by the optimal value of $\alpha$. 
Paraformer outperforms other models and achieves state-of-the-art results in  Precision ($0.7901$) and Macro-F2 ($0.7407$) and surpasses the current state-of-the-art system by Wehnert \etal \cite{wehnert2021legal}.
The best recall belongs to the systems of Nguyen \etal \cite{nguyen2021jnlp} and Wehnert \etal \cite{wehnert2021legal}.

\begin{table}
\centering
\caption{Results on COLIEE Datasets without Lexical Score}\label{tab:ch4_coliee_res}
\begin{tabular}{|l|c|c|c|}
\hline
\textbf{Systems} & \textbf{Precision} & \textbf{Recall} & \textbf{F2}     \\ \hline
\multicolumn{4}{|c|}{English Dataset}                                     \\ \hline
XLM-RoBERTa      & 0.2099             & 0.1975          & 0.1989          \\ \hline
Attentive CNN    & 0.0864             & 0.0864          & 0.0864          \\ \hline
Paraformer       & 0.3827             & 0.3450          & \textbf{0.3498} \\ \hline
\multicolumn{4}{|c|}{Japanese Dataset}                                    \\ \hline
XLM-RoBERTa      & 0.2940             & 0.3086          & 0.3086          \\ \hline
Attentive CNN    & 0.2593             & 0.2222          & 0.2263          \\ \hline
Paraformer       & 0.3457             & 0.3148          & \textbf{0.3182} \\ \hline
\end{tabular}
\end{table}

\begin{table}
\caption{Performance comparison on the COLIEE 2021's formal test set}
\label{tab:ch4_performance_coliee}
\center
\begin{tabular}{|l|c|c|c|}
\hline
\textbf{Run ID}                        & \textbf{Precision} & \textbf{Recall} & \textbf{F2} \\ \hline
\textbf{Paraformer*}                   & \textbf{0.7901}             & 0.7346          & \textbf{0.7407}       \\ \hline
OvGU\_run1                    & 0.6749             & 0.7778          & 0.7302      \\ \hline
JNLP.CrossLMultiLThreshold    & 0.6000             & \textbf{0.8025}          & 0.7227      \\ \hline
BM25.UA                       & 0.7531             & 0.7037          & 0.7092      \\ \hline
JNLP.CrossLBertJP             & 0.6241             & 0.7716          & 0.7090      \\ \hline
R3.LLNTU                      & 0.6656             & 0.7438          & 0.7047      \\ \hline
R2.LLNTU                      & 0.6770             & 0.7315          & 0.7039      \\ \hline
R1.LLNTU                      & 0.6368             & 0.7315          & 0.6875      \\ \hline
JNLP.CrossLBertJPC15030C15050 & 0.5535             & 0.7778          & 0.6838      \\ \hline
OvGU\_run2                    & 0.4857             & \textbf{0.8025}          & 0.6717      \\ \hline
TFIDF.UA                      & 0.6790             & 0.6543          & 0.6571      \\ \hline
LM.UA                         & 0.5679             & 0.5432          & 0.5460      \\ \hline
TR\_HB                        & 0.3333             & 0.6173          & 0.5226      \\ \hline
HUKB-3                        & 0.2901             & 0.6975          & 0.5224      \\ \hline
HUKB-1                        & 0.2397             & 0.6543          & 0.4732      \\ \hline
TR\_AV1                       & 0.2622             & 0.5123          & 0.3599      \\ \hline
TR\_AV2                       & 0.1490             & 0.5556          & 0.3369      \\ \hline
HUKB-2                        & 0.3272             & 0.3272          & 0.3258      \\ \hline
OvGU\_run3                    & 0.1570             & 0.7006          & 0.3016      \\ \hline
\end{tabular}

\end{table}

To better explain the result, we visualize the attention weights of the model.
Table \ref{tab:paraformer_weight_visualization} demonstrates the attention weights of Paraformer when answering queries related to Article 87 in Japanese Civil Code.
As we can see in the table, the model focuses differently on the content of Article 87 depending on the given query. 
For the first two queries, the model pays attention to the first path of the article, for the last query, the weight of item (2) is superior.

\begin{table}
\footnotesize
\center
\begin{tabular}{|p{4.5cm}|p{4.4cm}|p{2.8cm}|}
\hline
& \multicolumn{1}{p{4.4cm}|}{\textbf{Article 87 (1):}  If the owner of a first thing attaches a second thing that the owner owns to the first thing to serve the ordinary use of the first thing, the thing that the owner attaches is an appurtenance.} & \multicolumn{1}{p{2.8cm}|}{\textbf{Article 87 (2):} An appurtenance is disposed of together with the principal thing if the principal thing is disposed of.} \\ \hline
\textbf{Query 1:} Extended parts of a house shall be regarded as appurtenance.                  & \cellcolor[HTML]{9900FF}                                                                                                                                                                                                           & \cellcolor[HTML]{FFFFFF}                                                                                                                  \\ \hline
\textbf{Query 2:} Extended parts of a house shall be disposed when the house is no longer used. & \cellcolor[HTML]{B13BFF}                                                                                                                                                                                                           & \cellcolor[HTML]{E8C5FF}                                                                                                                  \\ \hline
\textbf{Query 3:} When an appurtenance is disposed of together with the principal thing?        & \cellcolor[HTML]{D79AFF}                                                                                                                                                                                                           & \cellcolor[HTML]{C266FF}                                                                                                                  \\ \hline
\end{tabular}
\caption{Weight visualization of Paraformer when answering querries related to Article 87 in Japanese Civil Code. }
  \label{tab:paraformer_weight_visualization}
\end{table}

Through visualization, we see that, besides improving performance, this approach also allows us to debug the model more easily.
Instead of accepting the model's prediction as an output of a black box, we can look at the attention weight to debug the model.
This is important for improving models in the legal field.
In addition, it also opens up potential on the application aspect as users can learn how to focus on important information from the data.

\subsection{Discussions}
For a better understanding of the model behavior on different content lengths, we divide the English dataset into five chunks.
The first chunk contains articles that are fewer than 100 characters in length.
The second, the third, and the fourth chunks contain articles between 100-200, 200-300, and 300-400 characters in length, respectively.
The fifth chunk contains articles that are longer than 400 characters.
We use the trained XLM-RoBERTa and Paraformer to predict and record the performance of each model in each chunk without the lexical score.

\begin{figure}[ht]
  \centering
\includegraphics[width=.8\linewidth]{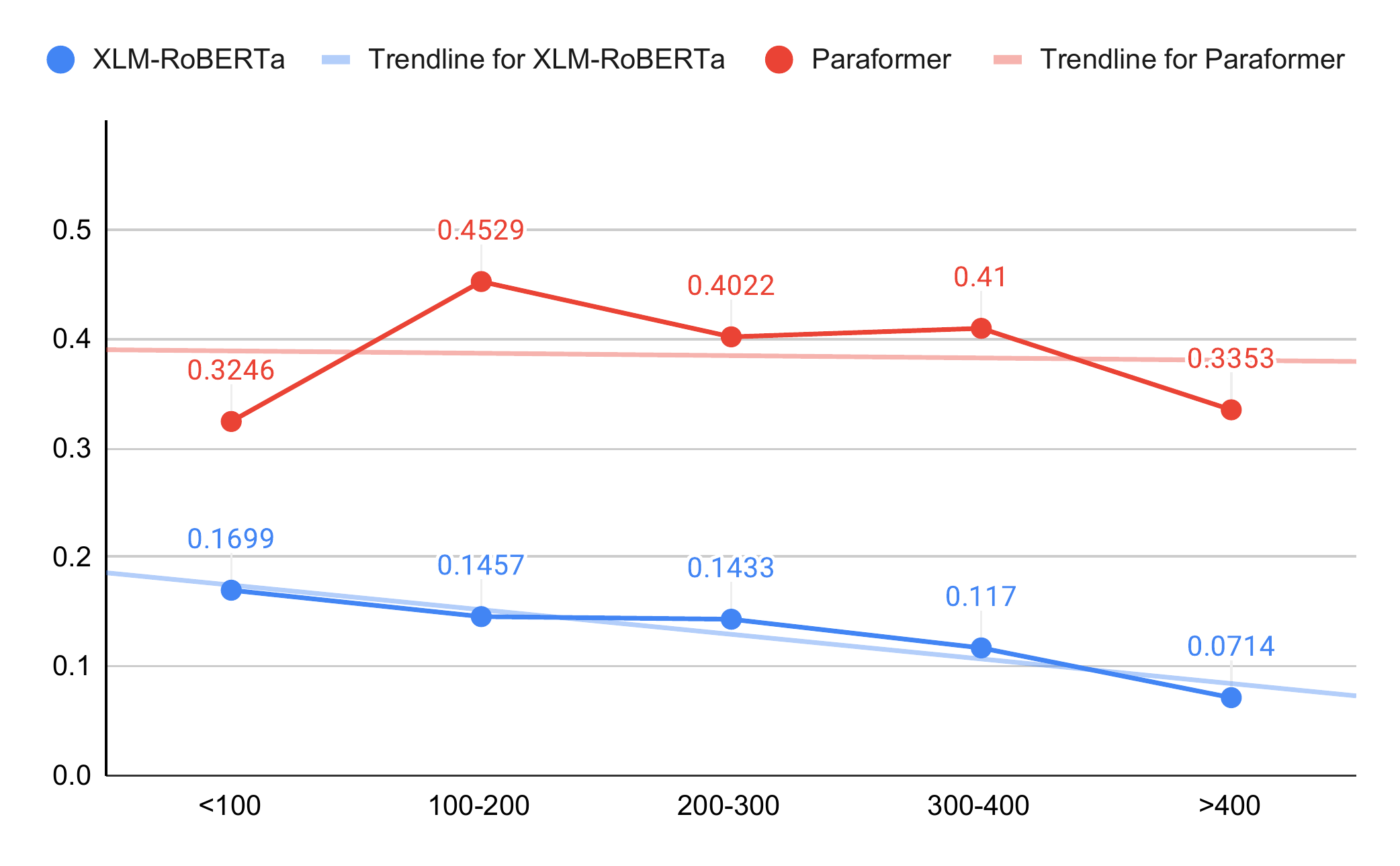}
  \caption{Performance of XLM-RoBERTa and Paraformer when working on different lengths of inputs.}
  \label{fig:ch4_different_length}
\end{figure}

Figure \ref{fig:ch4_different_length} shows the performance of the two models and their trendlines.
Although Paraformer overperforms XLM-RoBERTa in all chunks, the performance of both models reduces when the length of content increases.
However, looking at the trendlines, we can see that XLM-RoBERTa's performance tends to reduce faster than Paraformer in long content.
Note that in this chart, we only display the query length but in fact, these models need to handle both queries and articles as their inputs.
From this observation, we can see that our proposed architecture is appropriate to solve the problem of content length and take advantage of pretrained language models.

\newpage

\section{Summary of Chapter}

This chapter is about pretrained language models in deep legal processing.
We propose BERTLaw, ParaLaw and Paraformer as three candidates to solve different limitations of current pretrained language models.
From the results in Chapter 3, we see that the three factors that make a deep learning model work well in the legal domain are: good data representation, an adequate amount of data, and appropriate model architecture.
The models proposed in this chapter are all based on those observations.

The proposal of BERTLaw proves that there is the problem of sublanguage in law and it can be solved by pretraining the model with the dataset in such sublanguage.
The method of generating BERT Law is not a new method in deep learning.
Even so, we apply this method to legal domains and prove it's effectiveness.
BERT Law is created from scratch in both vocabulary and network weights.
Note that BERT Law was introduced at the same time as LEGAL-BERT \cite{chalkidis-etal-2020-legal} and both confirm the superiority of this approach.

ParaLaw Nets show us the ability to use aligned translations available in legal documents to improve the performance of language models.
With the need for legal internationalization of countries, high-quality translation resources of legal documents are available.
This is a great strength that we can further pretrain multilingual embeddings to help them better model legal concepts.
The two pretraining tasks we proposed in ParaLaw Nets force the models to predict the continuation of two multilingual legal sentences, thereby indirectly correcting the representation of concepts in vector space.
In addition, this approach allows us to obtain more training data in the downstream tasks (English and Japanese).
Since BERTLaw does not support Japanese, adding more Japanese data to train BERTLaw reduces the accuracy of this model, so it is difficult to have a fair configuration to compare the two winners of the two COLIEE seasons.

Paraformer contributes a novel architecture that leverages the power of pretrained language models with long content.
A robust model can still fail without complete information.
With technical limitations, pretrained language models are usually limited by maximum length setting.
This is their weakness in legal domains with lengthy legal sentences.
Paraformer's divide-and-conquer technique with the general attention mechanism helped us solve lengthy content problems while retaining the power of pretrained language models.

\chapter{Knowledge Injection for Deep Legal Models}
\section{Overview}

In this chapter, we propose different techniques for injecting expert knowledge into the language model to improve the performance in deep legal processing.
First, we investigate the linguistic knowledge and design a framework to inject this knowledge into the bulky language model without pretraining models from scratch.
This is an efficient approach to improve the performance without increasing the computational resource.
Second, we propose a novel technique to inject the knowledge about the structure of legal sentences into the Transformer models.
We also conduct detailed experiments to find the best configuration for this approach.
Third, we construct a novel conditional generative system that uses the knowledge of fairness to guide the generative models for high-quality terms of service content.
Our recommendations in this chapter come from the insights gained from working on the previous chapters.

\section{Linguistic Knowledge Injection}

\subsection{Introduction}
In the digital age, the amount of data generated per second is enough to train a model capable of abstracting many different patterns.
With a sufficiently large amount of data, these models are shown to be able to abstract the linguistic features on their own \cite{liu2019linguistic}.
Even so, training the model unsupervised using data on the internet can generate noisy, biased models \cite{bender2021dangers} and make deep learning black boxes to humans.
This section introduces a novel method to pretrain a small part of the whole transformer architecture of the pretrained language model.
Instead of pretrain a whole model, we append one more transformer layer, pretrain it to represent the linguistic knowledge.
By doing so, the knowledge in this layer is utilized for the model to have better performance in the downstream tasks.

This approach has three advantages compared to other methods.
First, this is a knowledge injection method that can take the edge of many different types of annotated resources in NLP. 
Linguistic is an important feature to understand language, especially in the legal domain.
Second, the injected information is not rigid but only a reference source for the model to reach its final conclusion.
Third, training and storing the appended component are extremely efficient, helping to solve environmental and financial problems.

\subsection{Research Method}

\begin{figure*}
\centering
\includegraphics[width=0.8\textwidth]{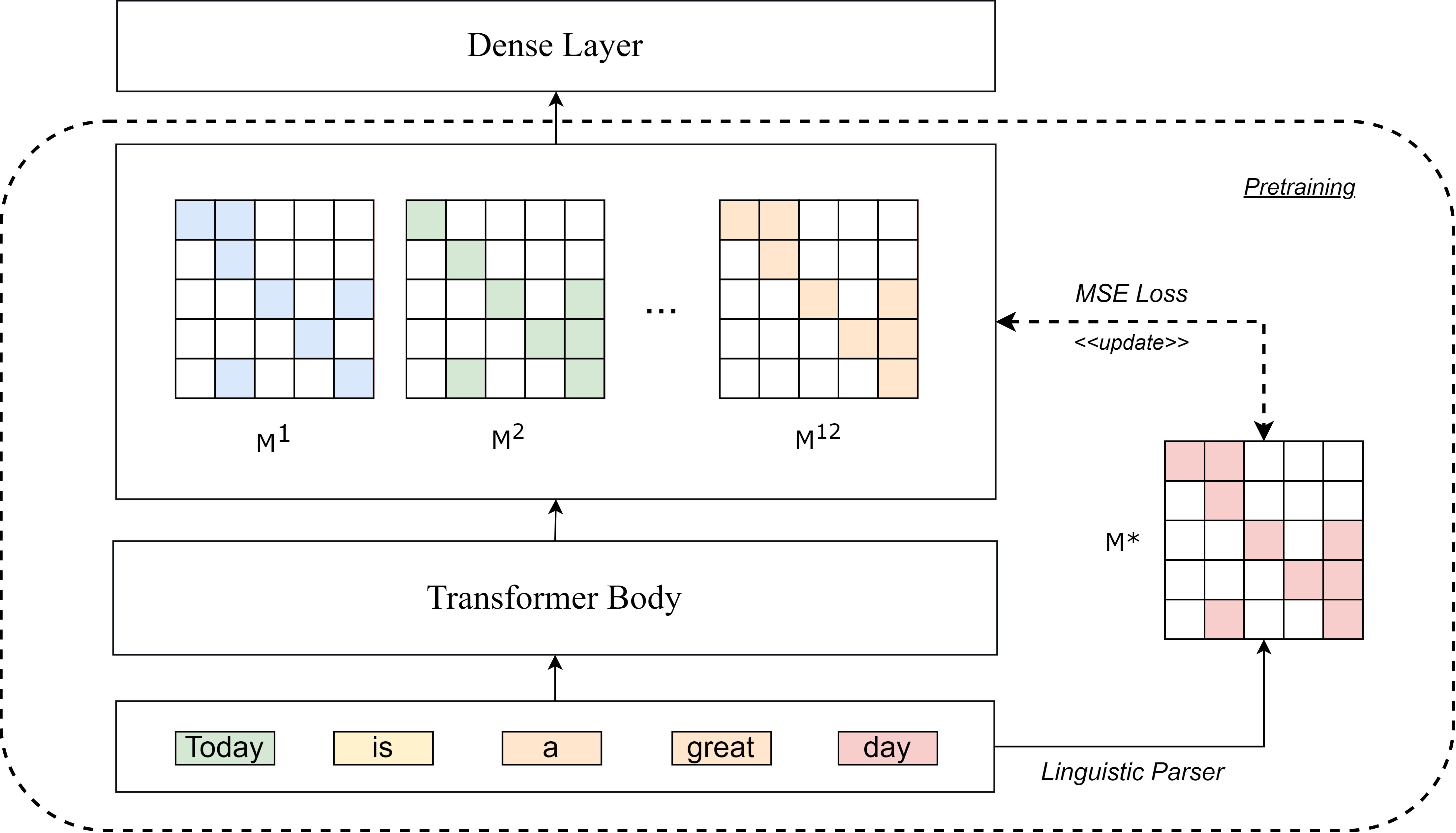}
  \caption{The general architecture of the framework. \textit{Transformer Body} is the pre-trained model (e.g. BERT). SDOI matrix $(M^*)$ is the relation matrix parsed from an external Linguistic Parser tool (e.g. Stanford NLP tool or spaCy).  \label{fig:ch5_hydra_framework}}
\end{figure*}

The overall architecture of the proposed framework is shown in Figure \ref{fig:ch5_hydra_framework}. Based on the transformer's architecture, we add one more transformer layer at the end of the architecture containing heads, which are pretrained with linguistic knowledge.
The model can choose to use the information of these heads or ignore it with the residual connections. 
Therefore, we can use linguistic knowledge as a reference source for the model without forcing it to accept such knowledge rigidly.
Since the idea is derived with a dependency structure, we give this architecture an inspiring name, \textit{HYDRA}, which stands for \textit{Hyper Dependency Representation Attentions}.

\subsubsection{Pretraining HYDRA Heads}

Let the input be $X = \{x_1, x_2, ..., x_n\}$ with $n$ be the sequence length, we obtain an $n \times n$ matrix $M^*$ containing the linguistic relationship of the words as introduced by Zhang et al.~\cite{zhang2020sgnet}.
After passing the input through the transformer layers, we get hidden state $H_{l}$ at the last layer with $l$ being the number of layers of the original transformer body.
We initiate and append  $l+1^{th}$ layer containing HYDRA heads and pass $H_{l}$ to this layer.
With $W_q$, $W_k$ are learnable parameters in the layer ${l+1}^{th}$, we calculate query and key vectors as in Equations \ref{eq:qvec} and \ref{eq:kvec}:
\begin{align}
q_{l+1}^h &= W_q \cdot H_{l}     \label{eq:qvec} \\
k_{l+1}^h &= W_k \cdot H_{l}    \label{eq:kvec}
\end{align}

\noindent With $d_k$ be the dimension of the key matrix $k_{l+1}^h$, we calculate the attention matrix for each head $M^h$ following Equation \ref{eq:attn_mtx}:

\begin{equation}
\label{eq:attn_mtx}
M^h = \frac{q_{l+1}^h \cdot {k_{l+1}^h}^\intercal}{\sqrt{d_k}}
\end{equation}

\noindent We then minimize the element-wise MSE loss ($\mathcal{L}$) between $M^*$ and $M^h$ for every HYDRA head:
\begin{align}
\mathcal{L} &= \frac{1}{n^2} \sum_{i=0}^{n^2} (M^*_i - M^h_i)^2 
\end{align}
\noindent where $i$  $(0 \leq i < n^2 )$  is the index of element in flatten attention matrix SDOI ($M^*_i$). The weights of the transformer body are frozen and the weights in $Q_{l+1}^h$ and $K_{l+1}^h$ in every HYDRA heads are updated via the backpropagation process.

After pretraining, we get the HYDRA heads corresponding to the transformer body coupled to it. These HYDRA heads are saved in the storage and ready to be delivered as a form of deep learning resource. Compared to storing pretrained transformer models, storing HYDRA heads saves much more storage space. A transformer model checkpoint can be measured in gigabytes, while a HYDRA head checkpoint in our experiment is less than 10 megabytes.

\subsubsection{Fine-tuning with HYDRA Heads}
Different from approaches that use linguistic knowledge that is rigidly injected into the model, we attach pretrained HYDRA heads to the transformer bodies. These models can refer to but are not limited by the linguistic knowledge learned in the HYDRA heads.

At this phase, the weights of the transformer body and HYDRA head are both updated to match the training data. 
The final model now contains $l+1$ layers, the first $l$ layers are from the original transformer body and the last layer contains HYDRA heads.
We simply pass the $H_{l+1}$ to a fully connected layer to get the logits for each downstream task. 
The loss function of each downstream task is defined similarly to the work \cite{devlin2018bert}. 
Intuition in this phase is that by acquiring linguistic knowledge, the system can properly model the problem rather than try to fit into the local minima.

\subsection{Experiments}

In the pretrain phase, we collect data from Wikimedia Downloads\footnote{https://dumps.wikimedia.org/} using WikiExtractor package~\cite{Wikiextractor2015}. 
We use the variant of dependency parser described by Honnibal and Johnson~\cite{honnibal2015improved}, provided by spaCy\footnote{https://spacy.io/}.
After data processing, we obtain 330,000 samples for training and 50,000 samples for validation.
We use BERT~\cite{devlin2018bert} as the base model to pretrain the HYDRA heads. 
We set the maximum sequence length to 512 and only use sentences whose length does not exceed this number to pretrain the HYDRA heads. 
The goal of this setting is to help the model learn the linguistic structure of complete sentences of the longest possible length.
With 1.8 million parameters, in our experiment, it takes only one or two epochs for the HYDRA heads to reach optimal loss on both the training set and the validation set.

To understand the model's behavior and the method's effectiveness, we run experiments and compare the performance of the HYDRA variants with the corresponding non-HYDRA baselines.
The experiments are run on datasets and standard settings of several famous benchmarks in natural language processing which include:

\begin{itemize}
    \item QNLI~\cite{rajpurkar2016squad}, MNLI~\cite{N18-1101,bowman2015large}, RTE~\cite{dagan2005pascal,bar2006second,giampiccolo2007third,bentivogli2009fifth} : Natural Language Inference and Texture Entailment
        \begin{itemize}
          \item Metric for QNLI and RTE: Accuracy.
          \item Metric for MNLI: Accuracy for both matched (m) and mismatched (mm) versions.
        \end{itemize}
    \item QQP~\cite{iyer2017qqp} and STS-B~\cite{cer2017semeval}: Semantically Equivalence Judgment
        \begin{itemize}
          \item Metric for QQP: Accuracy.
          \item Metric for STS-B: Pearson Spearman Correlation.
        \end{itemize}
    \item SQuAD~\cite{rajpurkar2016squad}, SQuAD 2.0~\cite{rajpurkar2018know}, COLIEE 2021 (Task 5) ~\cite{rabelo2021coliee}: Question Answering
        \begin{itemize}
          \item Metric for SQuAD and SQuAD 2.0: Exact Match and Macro-averaged F1 Score.
          \item Metric for COLIEE 2021 (Task 5): Accuracy.
        \end{itemize}

\end{itemize}

Table \ref{tab:result} shows the results of the models on the benchmarks and metrics described above.
BERT is a very strong baseline, it achieved high performance on all benchmarks.
Even so, our model still can slightly improve the results of BERT with the appended HYDRA heads.
This result supports the hypothesis that linguistic knowledge pretrained in HYDRA heads can boost the performance of the vanilla model.

% \usepackage{multirow}
\begin{table}
\centering
\caption{Experimental results on dev set of benchmark datasets\label{tab:result}}
\begin{tabular}{|l|c|c|}
\hline
\textbf{Benchmark} & \textbf{BERT} & \textbf{BERT HYDRA} \\ \hline
QNLI               & 0.9065                    & 0.9090                          \\ \hline
MNLI\_m            & 0.8373                    & 0.8401                          \\ \hline
MNLI\_mm           & 0.8419                    & 0.8458                          \\ \hline
RTE                & 0.6300                    & 0.6336                          \\ \hline
QQP                & 0.9062                    & 0.9067                          \\ \hline
STS-B              & 0.8800                    & 0.8812                          \\ \hline
SQuAD EM           & 0.8105                    & 0.8124                          \\ \hline
SQuAD F1           & 0.8838                    & 0.8851                          \\ \hline
SQuAD 2.0 EM       & 0.7137                    & 0.7161                          \\ \hline
SQuAD 2.0 F1       & 0.7458                    & 0.7480                          \\ \hline
COLIEE Task 5      & 0.5432                    & 0.5679                          \\ \hline
\end{tabular}
\end{table}

In addition to the performance improvement, this is a novel approach to inject linguistic knowledge into language models.
One notable feature of our improvement is that this new component is lightweight, requires low pretraining computation cost and storage.
With this paradigm, we can improve the bulky transformers models without pretraining the whole network again or forcing them to follow rigid linguistic rules.

\vfill 

\subsection{Discussions}
This section proposes an architecture-friendly and extensible method to improve the effectiveness of the transformer-based language models by pretraining and appending new knowledge-guided heads to their architecture.
We conduct the experiment with BERT as the base model and the dependency information as the external knowledge.
Our experiment shows that our lightweight component can help to boost the performance of transformer models and provide a flexible paradigm to partially inject the knowledge into bulky models without pretraining them again.
Extending this work, we can analyze the possibility of pretraining this component with different knowledge forms for problems in narrower domains or explore the potential of this approach for other data such as photos or videos.

\newpage
\section{Legal Knowledge Injection}

\subsection{Introduction}
In this section, we introduce TRE framework which is a knowledge injection framework for Transformer based models with requisite and effectuation data. Applying this framework, we investigate and pretrain variants of  TREBERT, pretrained models on BERT.
Logic structures are integral parts of legal sentences.
Identifying criminals, breaches of contracts, and a host of other important legal decisions are all based on logic.
A novel author can use better language than a lawyer but standing in court they cannot justify a person from the death penalty with their words without logic.
Similarly, a language model trained on a giant corpus without knowledge of logic is intrinsically useless in law.
This can make it difficult to answer the inference questions of the law, which are critical in being able to bring the results to reality.

Analyzing the previous pretraining methods in the legal domain, we find that these methods have the same thing in common, that they are pretrained unsupervised on a large corpus.
By doing so, we can create language models that accurately describe the relationships of concepts, terms, and syntax used in legal documents.
These models can also find latent rules expressed in words, use extrapolation to make decisions.
However, it is impossible for the model to find all the latent rules just by identifying co-occurring terms.
This is a process that requires much time, a lot of computational power, a huge amount of data.
Similar to the issues in math problems, it is difficult for the model to find logical rules through unsupervised training.
Our approach is a further pretraining method based on a supervised paradigm.

\subsection{Research Method}
\label{sec:ch5_legal_knowledge_method}

With a different purpose than daily life sentences, legal sentences often require rigor and logic.
As the product of thousands of years of human civilization, the logic of the existing laws reaches a very high level.
From a syntactic point of view, two important components of a law sentence to form an equivalent logical proposition are requisite and effectuation.
Requisite and effectuation can be formed from smaller logical parts such as antecedence, consequence, and topic.

With a classic example in the logic ``If it rains, the road is wet.", we can easily see that this sentence has a requisite segment and an effectuation segment. 
The requisite and effectuation segments are often complex in practice, they can be nested and even interleaved.
In legal sentences, besides \textit{requisite} and \textit{effectuation}, another common logical structure is \textit{unless}, which indicates exceptions where the main requisite and effectuation do not apply.
Let's consider the following example: ``Gifts not in writing may be revoked by either party; provided, however, that this shall not apply to any portion of the gift for which performance has been completed." 
With such a complex sentence, it is easy to see that there is more than one requisite and effectuation pair in this sentence.
Therefore, it is difficult for a language model with averaging and interpolation capabilities to infer logical structures on its own through unsupervised training.
To correctly annotate law sentences with many interlocking logical structures, we need to use multilayer annotation \cite{nguyen2018recurrent}. 
Table \ref{tab:ch5_annotation_sample} is the annotation of the above example.

\begin{table*}

\centering
\caption{Multilayer annotation in the BIOE schema of  \textit{requisite}, \textit{effectuation} and \textit{unless} segments for the sample ``Gifts not in writing may be revoked by either party; provided, however, that this shall not apply to any portion of the gift for which performance has been completed."}
\label{tab:ch5_annotation_sample}
\begin{tabular}{|l|c|c|c|l|l|c|c|c|}
\cline{1-4} \cline{6-9}
\textbf{Token} & \textbf{L1} & \textbf{L2} & \textbf{L3} & \textbf{} & \textbf{Token} & \textbf{L1} & \textbf{L2} & \textbf{L3} \\ \cline{1-4} \cline{6-9} 
Gifts          & B-R              & B-E             & -                &           & shall          & -                & I-E             & I-U              \\ \cline{1-4} \cline{6-9} 
not            & I-R              & -               & -                &           & not            & -                & I-E             & I-U              \\ \cline{1-4} \cline{6-9} 
in             & I-R              & -               & -                &           & apply          & -                & I-E             & I-U              \\ \cline{1-4} \cline{6-9} 
writing        & E-R              & -               & -                &           & to             & -                & I-E             & I-U              \\ \cline{1-4} \cline{6-9} 
may            & -                & I-E             & -                &           & any            & -                & I-E             & I-U              \\ \cline{1-4} \cline{6-9} 
be             & -                & I-E             & -                &           & portion        & B-R              & I-E             & I-U              \\ \cline{1-4} \cline{6-9} 
revoked        & -                & I-E             & -                &           & of             & I-R              & I-E             & I-U              \\ \cline{1-4} \cline{6-9} 
by             & -                & I-E             & -                &           & the            & I-R              & I-E             & I-U              \\ \cline{1-4} \cline{6-9} 
either         & -                & I-E             & -                &           & gift           & I-R              & E-E             & I-U              \\ \cline{1-4} \cline{6-9} 
party          & -                & E-E             & -                &           & for            & I-R              & -               & I-U              \\ \cline{1-4} \cline{6-9} 
;              & -                & -               & -                &           & which          & I-R              & -               & I-U              \\ \cline{1-4} \cline{6-9} 
provided       & -                & -               & B-U              &           & performance    & I-R              & -               & I-U              \\ \cline{1-4} \cline{6-9} 
,              & -                & -               & I-U              &           & has            & I-R              & -               & I-U              \\ \cline{1-4} \cline{6-9} 
however        & -                & -               & I-U              &           & been           & I-R              & -               & I-U              \\ \cline{1-4} \cline{6-9} 
,              & -                & -               & I-U              &           & completed      & E-R              & -               & E-U              \\ \cline{1-4} \cline{6-9} 
that           & -                & -               & I-U              &           & .              & -                & -               & -                \\ \cline{1-4} \cline{6-9} 
this           & -                & B-E             & I-U              &           &                &                  &                 &                  \\ \cline{1-4} \cline{6-9} 
\end{tabular}

\end{table*}

\begin{figure*}
    
    \centering
    \includegraphics[width=.9\textwidth]{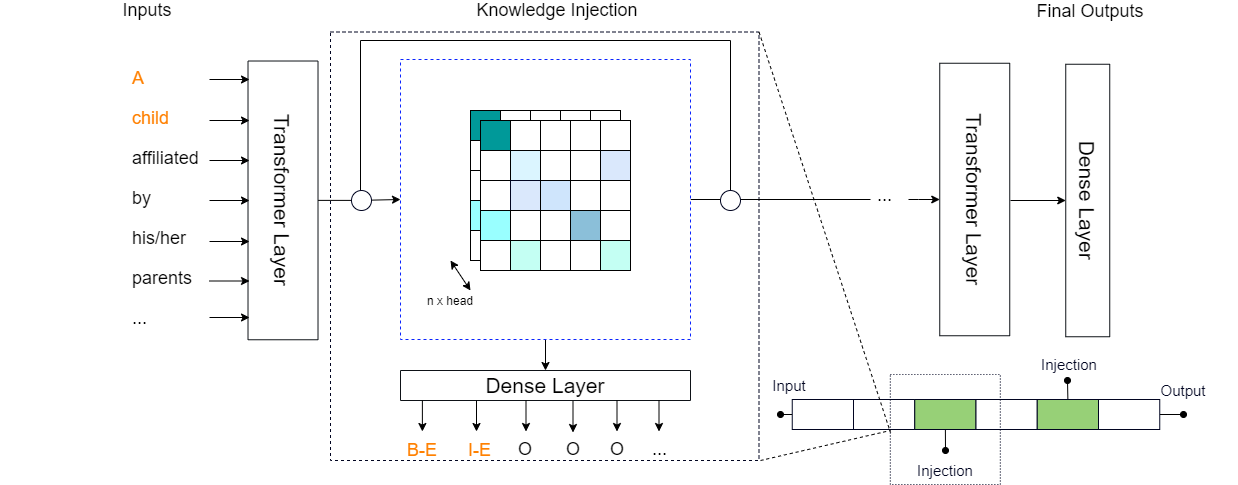}
    \caption{General flow of TRE Framework for Transformer models.}
    \label{fig:ch5_general_flow}
\end{figure*}

With the goal of building a Transformer model that can learn to recognize the segments of logical structures, we propose the \gls{TRE} (Transferred-Requisite-Effective) Framework.
This framework makes it possible to inject the logical structure information into the self-attention layers of the Transformer so that the model can form the corresponding abstractions.
Unlike conventional pretraining approaches, labels are usually provided at the last Transformer layer, within this framework, information about logical structures can be injected into the hidden layers (Figure \ref{fig:ch5_general_flow}).

With this framework, knowledge injection is done through a gradient descent process.
Instead of rigidly specifying information about logical structures through constants, the model's parameters need to be updated so that corresponding abstractions can be formed.
The novelty of this framework lies in the Transferred-Requisite-Effective self-attention layers (TRE layers).
These layers stores the knowledge about recognizing logical structures in the legal sentence, which is learned from the provided labels in pretraining data.

Suppose after the layer $i-1^{th}$, we have a signal sequence of length $M$ which can be presented as $E^{i-1}=(e^{i-1}_1, e^{i-1}_2, . .., e^{i-1}_M)$.
The $i^{th}$ layer is a TRE layer, basically, the information flow is the same as in a regular Transformer layer.
At the $i^{th}$ layer, vector $E^{i-1}$ is multiplied by the attention matrices $Q^i$, $K^i$, $V^i$ to get the corresponding attention vectors.
These attention vectors are combined according to Equation~\ref{eq:ch5_softmax} to get the corresponding output at the $i^{th}$ transfomer layer $Z^i=(z^i_1, z^i_2, ..., z^i_M)$.

\begin{equation}
    \label{eq:ch5_softmax}
    Z^i=softmax(\frac{Q^i \times K^{i\top}}{\sqrt{d}}V^i)
\end{equation}

In the forward direction of Transformer architecture, we normalize $Z^i$ with a layer normalization \cite{ba2016layer} and a dense layer, the signal can also be transmitted directly through the residual connections.
To pass the labels of the logical structures to the network, at the branching direction as an injection needle, $Z^i$ after going through a dense layer and the softmax function as in Equation~\ref{eq:ch5_injected_label}, predicted labels and gold labels are compared and the loss is backpropagated for updating parameters.

\begin{equation}
    \label{eq:ch5_injected_label}
    L=softmax(Z^i \times W + b)
\end{equation}

\subsection{Experiments}

We pretrain TREBERT as described in Section \ref{sec:ch5_legal_knowledge_method}.
This pretraining process has the following characteristics.
Firstly, it is further pretraining on a Transformer model, \ie the original model needs to be pretrained on basic linguistic tasks to form contexture embedding on a specific vocabulary before being pretrained with logical structure data.
Secondly, it is implemented as supervised learning, \ie trained with labeled data.
Thirdly, the TRE layers where the label is injected will be trained in parallel.

From the characteristics of this process, we see that, instead of assigning just one configuration, we can experiment with different configurations of the framework.
This experiment not only helps to find the best configuration but also helps us better understand the behavior of the model during this phase.
Since pretraining is done in the form of supervised training, we can track the performance of the model on a validation set.
Our assumption is that a good configuration can help the model to make a good abstraction, thereby making accurate predictions on the validation set.

\begin{table}
\centering
\caption{Tag distribution following BIOE schema of the pretraining data.}
\label{tab:tag_distribution}
\begin{tabular}{|l|l|r|}
\hline
\textbf{Tag} & \textbf{Meaning}& \textbf{Occurence} \\ \hline
B-E          & Begin Effectuation Part& 1,982               \\ \hline
B-R          & Begin Requisite Part& 2,408               \\ \hline
B-U          & Begin Unless Part& 260                \\ \hline
E-E          & End Effectuation Part& 2,088               \\ \hline
E-R          & End Requisite Part& 2,395               \\ \hline
E-U          & End Unless Part& 259                \\ \hline
I-E          & Inside Effectuation Part& 26,762              \\ \hline
I-R          & Inside Requisite Part& 31,474              \\ \hline
I-U          & Inside Unless Part& 6,270               \\ \hline
O            & Others& 148,540             \\ \hline
\end{tabular}
\end{table}

With statistics from Table \ref{tab:tag_distribution}, we can see that this is a classification problem with unbalanced labels.
Therefore, we use the precision, recall, and F1 metrics on logical structure parts to evaluate and compare 

We pretrain the variants of TREBERT considering two aspects: the position of the TRE Layers and the loss portion between them.
Just considering the position of the TRE layer on the 12 Transformer layers of BERT, we have $^{12}P_3 = 1320$ cases.
Testing all configurations is resource-intensive, so we use the boundary value analysis technique to generate representative configurations for the positions and random search for the loss portion.
Table \ref{tab:positional_configs} represents positional configurations and their performances on pretraining data.
We also included in the table 30 configurations where the TRE layers are randomly decided.
When conducting experiments with TRE layer position, we fixed the portion loss of all three logical structure layers equally.

\begin{table*}
\centering
\caption{Representative positional configuration's performances on validation set. TREBERT\_X\_Y\_Z stands for the best configuration that the knowledge is injected in Transformer layers $X^{th}$, $Y^{th}$ and $Z^{th}$.}
\label{tab:positional_configs}
\begin{tabular}{|l|c|c|c|}
\hline
\textbf{Configuration}   & \multicolumn{1}{l|}{\textbf{Precision}} & \multicolumn{1}{l|}{\textbf{Recall}} & \multicolumn{1}{l|}{\textbf{F1 Score}} \\ \hline
\multicolumn{4}{|l|}{\textit{Uniform loss portion}}                                                                                            \\ \hline
TREBERT\_8\_10\_12   & 0.5890                                  & 0.7000                               & 0.6373                                 \\ \hline
TREBERT\_7\_8\_9     & 0.5441                                  & 0.7102                               & 0.6151                                 \\ \hline
TREBERT\_7\_9\_11    & 0.5420                                  & 0.7305                               & 0.6140                                 \\ \hline
TREBERT\_10\_11\_12  & 0.5544                                  & 0.6661                               & 0.6064                                 \\ \hline
TREBERT\_6\_9\_12    & 0.5262                                  & 0.7424                               & 0.5959                                 \\ \hline
TREBERT\_4\_8\_12    & 0.4500                                  & 0.6966                               & 0.5102                                 \\ \hline
TREBERT\_2\_7\_12    & 0.3994                                  & 0.7136                               & 0.4549                                 \\ \hline
TREBERT\_4\_5\_6     & 0.3550                                  & 0.5814                               & 0.4235                                 \\ \hline
TREBERT\_1\_6\_11    & 0.3490                                  & 0.7085                               & 0.4027                                 \\ \hline
TREBERT\_2\_4\_6     & 0.3266                                  & 0.6678                               & 0.3894                                 \\ \hline
TREBERT\_1\_5\_9     & 0.3260                                  & 0.7254                               & 0.3704                                 \\ \hline
TREBERT\_1\_4\_7     & 0.2745                                  & 0.6695                               & 0.3254                                 \\ \hline
TREBERT\_1\_3\_5     & 0.2199                                  & 0.5695                               & 0.2726                                 \\ \hline
TREBERT\_1\_2\_3     & 0.1655                                  & 0.4932                               & 0.2124                                 \\ \hline
Avg Random (30 runs) & 0.3654                                  & 0.5826                               & 0.4088                                 \\ \hline
\multicolumn{4}{|l|}{\textit{Optimized loss portion}}                                                                                          \\ \hline
TREBERT\_8\_10\_12   & 0.5807                                  & 0.7373                               & 0.6555                                 \\ \hline
TREBERT\_7\_8\_9     & 0.5725                                  & 0.6814                               & 0.6313                                 \\ \hline
\textbf{TREBERT\_7\_9\_11}    & \textbf{0.6300}                                  & \textbf{0.7723}                                      & \textbf{0.6939}                                       \\ \hline
\end{tabular}
\end{table*}

From the experiment, it can be seen that the good positions to inject the knowledge of the logical structure are in the deeper layers.
This can be explained that the knowledge injection needs to match the level of abstraction of the neural network.
For deep learning models, in the early layers, the abstractions are low-level.
At the deeper layers, the level of abstraction increases.
The knowledge injected into the early layers will deviate from the abstraction level compared to the unsupervised features formed during the previous pretraining.
The table also shows that the distance between the injection needle positions should be 1 to 2 layers.
We select the 3 best configurations to continue conducting random searches to find the best portion of the loss functions for each.
Different positional configurations have different optimal loss portions.

We use question answering data of COLIEE 2021 competition to verify the effectiveness of TREBERT.
For each statement, the model needs to predict whether the statement is lawful or not.
This is data that can evaluate the strength of pretrained models because of the small number of samples.
The systems need to perceive the semantics in the sentence to give the correct answer.

In the data provided by COLIEE, the official test set includes 81 yes/no questions.
To generate the training data, we use the method proposed by \cite{nguyen2019deep}, data from previous years and from the Japanese civil code are augmented with simple negation rules.
After the augmentation process, we have 4,000 samples. We spend 10\% for the development set and the rest is for finetuning TREBERT and baseline models.

Our experimental baselines include original BERT, LegalBERT\_SC (Legal BERT from scratch), and LegalBERT\_FP (Legal BERT further pretrained).
These baselines have the common feature of being pretrained unsupervised on lexical tasks and not pretrained with logical structures.
The measurement used in this task is accuracy.
 TREBERT\_8\_10\_12, TREBERT\_7\_8\_9 and TREBERT\_7\_9\_11, the variants of TREBERT that performed best on the pretraining task, are included in the experiment.
In addition, we also used a configuration with poor pretraining results, TREBERT\_1\_2\_3, to further understand the behavior of this model family.

To ensure fairness in the number of parameters, the baselines and variants of TREBERT all use the architecture and configuration of BERT Base Uncased.
Note that TREBERT's injection needles are removed after pretraining so they don't affect the parameter count. The experiments are conducted with GPU Tesla P100-PCIE-16GB.

\begin{table}
\caption{Performance on the test set of COLIEE 2021.}
\centering
\begin{tabular}{|l|c|c|}
\hline
\textbf{Model / Configuration} & \textbf{Correct} & \textbf{Accuracy} \\ \hline
% BaseLine                   & No 43/All 81     & 0.5309            \\ \hline
TREBERT\_8\_10\_12         & 52              & 0.6420            \\ \hline
TREBERT\_7\_9\_11          & 52              & 0.6420           \\ \hline
TREBERT\_7\_8\_9           & 49               & 0.6049            \\ \hline
LEGAL\_BERT\_SC            & 47               & 0.5802            \\ \hline
LEGAL\_BERT\_FC            & 46               & 0.5679            \\ \hline
Original BERT                       & 44               & 0.5432            \\ \hline
TREBERT\_1\_2\_3           & 44               & 0.5432            \\ \hline
\end{tabular}
\label{tab:ch5_result}

\end{table}

Experimental results in Table \ref{tab:ch5_result} show that TREBERT\_8\_10\_12 and TREBERT\_7\_9\_11 lead the rankings with 52 correct answers out of a total of 81 questions in the test set, followed by TREBERT\_7\_8\_9, LEGAL\_BERT\_SC and LEGAL\_BERT\_FC with 49, 47 and 46 correct answers. 
Original BERT and TREBERT\_1\_2\_3 answered 44 questions correctly.
This result shows us that the hypothesis made at the beginning of the section is reasonable.
Pretraining the models using a logical structure helps them to make better predictions in tasks that require understanding in the legal domain.
Compared with the results announced by COLIEE-2021 \footnote{https://sites.ualberta.ca/~rabelo/COLIEE2021/results\\/task5\_res.html}, top TREBERT variants (TREBERT\_8\_10\_12, TREBERT\_7\_9\_11) achieved state-of-the-art performance.

To better understand the behavior of the model, we visualize the self-attention weight of the last layer where the injection needle is attached.
If the model generates the correct abstraction, the attention matrix must reflect that.
Table \ref{tab:ch5_trebert_attention} is about the self-attention visualization of the $11^{th}$ layer in TREBERT\_7\_9\_11 with the input as  ``Gifts not in writing may be revoked by either party; provided, however, that this shall not apply to any portion of the gift for which performance has been completed.".

% \begin{figure}
    
%     \centering
%     \includegraphics[width=.55\textwidth]{figures/chap5/BERT_attention.png}
%     \caption{Self-attention visualization of the 11th layer of TREBERT\_7\_9\_11 with the input as  ``Gifts not in writing may be revoked by either party; provided, however, that this shall not apply to any portion of the gift for which performance has been completed.".}
%     \label{fig:ch5_trebert_attention}
% \end{figure}

\begin{table}
\center
\footnotesize
\begin{tabular}{|lllllllll|}
\hline
\cellcolor[HTML]{D3D3D3}Gift  & not                           & in                               & writing                         & may                               & be                         & revoked                         & by                           & either                         \\
party                         & ;                             & provided                         & ,                               & however                           & ,                          & that                            & this                         & shall                          \\
not                           & apply                         & to                               & any                             & portion                           & of                         & the                             & gift                         & for                            \\
which                         & performance                   & has                              & been                            & completed                         & .                          &                                 &                              &                                \\ \hline
Gift                          & \cellcolor[HTML]{4991C2}not   & \cellcolor[HTML]{BDD7E9}in       & \cellcolor[HTML]{DBE9F3}writing & \cellcolor[HTML]{E2EDF5}may       & \cellcolor[HTML]{C0D9EA}be & \cellcolor[HTML]{E0ECF4}revoked & \cellcolor[HTML]{EBF3F8}by   & \cellcolor[HTML]{FAFCFD}either \\
\cellcolor[HTML]{FAFCFD}party & \cellcolor[HTML]{F8FBFD};     & \cellcolor[HTML]{FCFDFE}provided & ,                               & \cellcolor[HTML]{FBFCFE}however   & \cellcolor[HTML]{E8F1F7},  & \cellcolor[HTML]{F4F8FB}that    & \cellcolor[HTML]{FDFEFE}this & \cellcolor[HTML]{FDFEFE}shall  \\
\cellcolor[HTML]{FBFCFE}not   & \cellcolor[HTML]{FEFEFF}apply & \cellcolor[HTML]{FCFDFE}to       & \cellcolor[HTML]{FBFCFE}any     & \cellcolor[HTML]{F4F8FB}portion   & \cellcolor[HTML]{FEFEFF}of & \cellcolor[HTML]{FDFEFE}the     & \cellcolor[HTML]{FBFCFE}gift & \cellcolor[HTML]{FEFEFF}for    \\
\cellcolor[HTML]{EDF4F9}which & performance                   & \cellcolor[HTML]{FEFEFF}has      & \cellcolor[HTML]{FDFEFE}been    & \cellcolor[HTML]{FDFEFE}completed & \cellcolor[HTML]{FDFEFE}.  &                                 &                              &                                \\ \hline
\end{tabular}
\caption{Self-attention visualization of the $11^{th}$ layer in TREBERT\_7\_9\_11.}
    \label{tab:ch5_trebert_attention}
\end{table}

Looking at Table \ref{tab:ch5_trebert_attention}, it can be seen that TREBERT\_7\_9\_11 has adjusted its attention to the logical structures.
With the token ``gift", from Table~\ref{tab:ch5_annotation_sample}, we can see that it belongs to both logical parts requisite and effectuation.
TREBERT\_7\_9\_11's attention weights correctly tie ``gift" to the tokens to the requisite (gift not in writing) and to the effectuation (gift may be revoked by either party) parts.
This could explain the outperforming on the test set of the model.

TREBERT\_1\_2\_3 has bad performance on the pretraining task, resulting in no performance improvement compared to the original BERT.
As analyzed above, knowledge injection fails due to the incompatibility of abstraction between data and model architecture.
This result is consistent with the assumption that a model capable of analyzing the logical structures can answer the legal questions better.
Working with different configurations in the pretraining phase helps us find the right place to inject each type of knowledge.

\subsection{Discussions}
In this study, we propose and investigate TRE framework, a logic-structure knowledge injection approach for pretrained Transformer models.
We then apply the TRE framework to pretrain the variants of TREBERT from the original BERT model.
Our detailed experiments and surveys show the effectiveness and explainability of the method.
A model having good skill in recognizing logical structure performs better on legal question answering.

Although in this section, the TRE framework is proposed with logical structures, the idea is general and extensible.
Variations of knowledge can be discovered and utilized in different tasks.
In future studies, we want to inject NLP annotated resources into Transformer models with TRE framework to get better and more explanatory pretrained models.
In addition, the limitation of this approach is that finding good configurations consumes a lot of computing power for hyperparameter optimization, overcoming this limitation is also an interesting research direction.

\newpage
\section{Pretrained Self-regulated Generation}

\subsection{Introduction}

In this section, we introduce a novel pretrained self-regulated generative framework with the aim of introducing the idea that knowledge learned from one model can be used to improve the results of another generative model in the framework. 
Generative models have long been introduced as a way of demonstrating machine intelligence to language.
They developed from rule-based models \cite{weizenbaum1966eliza} to statistical-based models \cite{angeli2010simple,holtzman2018learning,wolf2019transfertransfo} and the latest to attentive deep learning-based models like GPT-2~\cite{radford2019language}, GPT-3~\cite{brown2020language} and BART \cite{lewis2019bart}, which recently received great attention the society.

Not only that, these models also raise public concerns with destructive applications such as fake news.
This shows that it is almost impossible to distinguish human-written text from the language-generated text.
The question is, is it possible to use a form of human-accepted knowledge to guide these models to produce high-quality texts.
Looking for an answer to this question, we experiment with the terms-of-service generation problem.
This problem contains two main difficulties. First, it requires a balance between automation and the will of the editor. Second, the meaning of the content that the system generates should be of high quality and fairness.

In this section, we introduce BART2S, a framework that contains two components, generator and discriminator.
The generator is responsible for generating terms of use, taking into account the balance between editor will and automation.
To this end, we design this component as a sequence-to-sequence model on the title-based generation problem.
The discriminator is trained with the knowledge of fairness, then plays the role of a regulation component to ensure the output quality of the generative framework with few-shot tuning.
With this process, our framework introduces a novel way to use encoded knowledge for the high-quality content generation problem.

\subsection{Research Method}
The challenge of the title-based generation problem is signal recovery.
We need to recover the information of an entire clause based on their brief title.
Therefore, we propose 3 training tasks for generators: next sentence generation, title-based generation and paraphrasing.
Each task contributes a skill to the model that can generate content from the title.

For the discriminator, we let this component learn the knowledge of fairness through the fairness classification task.
In essence, this is a binary learning task.
For each text input, the model needs to learn to evaluate whether the text is a legally fair text.
This is the component that sets our framework apart from other systems and helps us to create high-quality content from the generative model.

\begin{figure}[ht]
  \centering
\includegraphics[width=.8\linewidth]{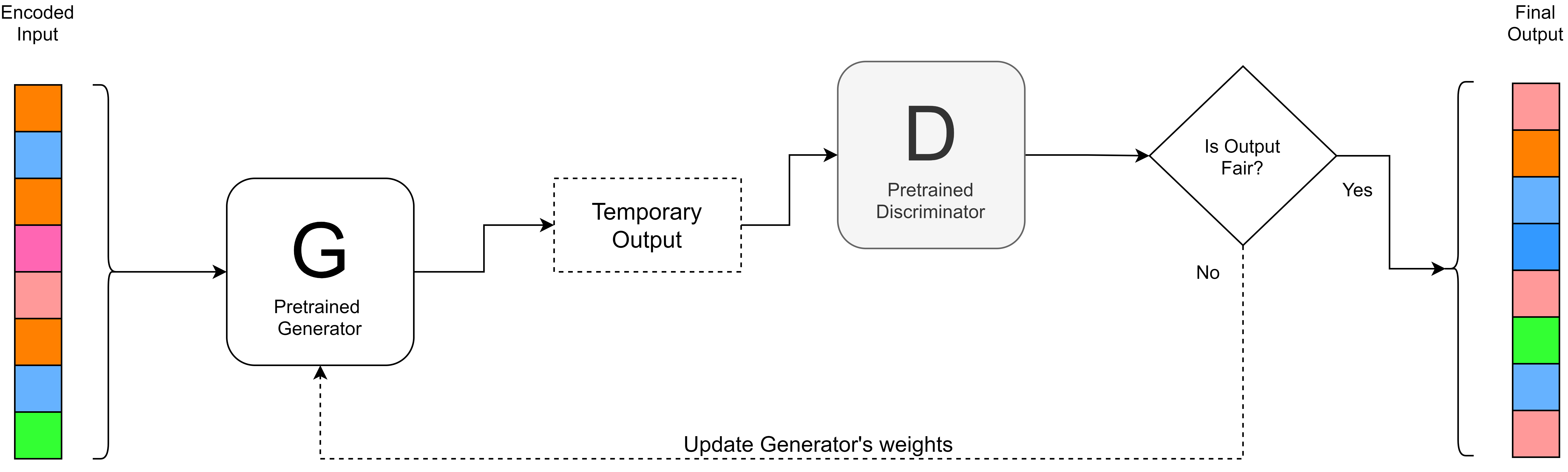}
  \caption{Fewshot tuning for generation in BART2S framework.}
  \label{fig:ch5_fewshot_tuning}
\end{figure}

The few-shot tuning process can be described as in Figure \ref{fig:ch5_fewshot_tuning}.
We consider the two models as differentiable functions $G(x, \theta_g)$ and $D(x, \theta_d)$ with $x$, $\theta_g$, and $\theta_d$ are the input, generator's parameters and discriminator's parameters, respectively. 
To generate high quality content, we minimize $log(1-D \circ G(x, \theta_g))$ using gradient descent process.
An important modification in our framework is the \textit{SoftArgMax} function at the last layer of the generator. 
To pass the loss across the two models, we need to replace the standard \textit{ArgMax} with the function as follows:

\begin{equation}
\operatorname{SoftArgMax}(x)=\sum_{i} \frac{e^{\beta x_{i}}}{\sum_{j} e^{\beta x_{j}}} i    
\label{eq:softargmax}
\end{equation}

\noindent where $x = [x_1, x_2, x_3, ..., x_n]$ and $ \beta \geq 1 $.

In the few-shot tuning process, we do not update the weights of the discriminator by freezing their gradient.
This is important so that this component becomes an independent observer.
The only standard it bases on is the knowledge of fairness it learns at the pretraining phase.
As a result, the total loss reduction of the framework comes from the better content generated by the generator.
This design makes a bold difference in our system compared to naive copy/paste systems and other non-regulated systems.

\subsection{Experiments}

Table \ref{tab:ch5_bart2sdata} contains the information about the data to train the generator and discriminator.
Pretraining the generator, each input is transformed using Token Masking, Deletion and Infilling \cite{lewis2019bart}.
For the discriminator, we use the knowledge of fairness from the ToS;DR project, which provides the rating of legal experts over different Terms of Service from many different companies.

\begin{table}
\footnotesize
\center
\begin{minipage}{\textwidth}
\begin{tabular}{|l|p{4.8cm}|p{3.8cm}|r|}
\hline
\textbf{Component} & \textbf{Task}                    & \textbf{Data source}              & \textbf{Samples} \\ \hline
Generator          & Task 1: Next Sentence Generation & Crawled Terms of Service          & 5,323                 \\
          & Task 2: Title-based Generation   & LawInsider Contracts\footnote{https://lawinsider.com} & 901                   \\
          & Task 3: Paraphrasing             & MSRP dataset~\cite{dolan2004unsupervised}                      & 3,728                 \\ \hline
Discriminator      & Fairness Classification          & ToS;DR project\footnote{https://tosdr.org/}                    & 4,152                 \\ \hline
\end{tabular}
\end{minipage}
\caption{Data for training generator and discriminator.}\label{tab:ch5_bart2sdata}
\end{table}

To evaluate the system, we  use both automatic and human-based metrics.
For the generator, we use BLEU scores on the 1000 most popular contract terms according to Law Insider statistics.
For the discriminator, we use accuracy on the 10\% of data as the validation set.
For the entire framework, we manually evaluate 4 criteria \textit{Grammar}, \textit{Readability}, \textit{Relevance}, and \textit{Fairness}.
We select control systems from different robust configurations of BART Large \cite{lewis2019bart}.
Each model is passed in 30 short titles with an average length of 23 characters. Their output is evaluated by 10 independent evaluators according to the formula:

\begin{equation}
    score_a(\text {M})=\frac{1}{n}\sum_{i=1}^{n} \frac{p_a^i}{s}
    \label{eq:score}
\end{equation}

\noindent In which, $score_a(\text {M})$ is the evaluation score of model $M$ in aspect $a$, $s$ is the total of sentences, $n$ is the number of evaluators, $p_a^i$ is the number of sentences evaluated as possitive by $i^{th}$ evaluator in the aspect $a$.

\begin{table}
\center
\small
\begin{tabular}{|l|c|}
\hline
\textbf{Model / Approach} & \textbf{Performance} \\ \hline
Generator - All Tasks     & 60.1                \\
Generator - w/o Task 1    & 59.9                \\
Generator - w/o Task 2    & 57.3                \\
Generator - w/o Task 3    & 56.3                \\ \hline
Discriminator             & 66.0                \\ \hline
\end{tabular}
\caption{Performance of components with automatic metrics. Generator is evaluated in BLEU, Discriminator is evaluated in accuracy.}
\label{tab:automatic_eval}
\end{table}

Table \ref{tab:automatic_eval} shows the performance of components with automatic metrics, in which the generator is evaluated in BLEU and the discriminator is evaluated in accuracy. 
We can see that the generator trained with all proposed tasks achieve the best performance with a 60.1 BLEU score.
Eliminating any task causes results to drop.
The discriminator is trained with early stopping. This model achieves 66\% accuracy on the validation set, which indicates that the problem of fairness classification is not straightforward.

\begin{table*}[!htbp]
\center
\begin{tabular}{|l|c|c|c|c|}
\hline
\textbf{System}  & \textbf{Grammar} & \textbf{Readability} & \textbf{Relevance} & \textbf{Fairness} \\ \hline
BART Large w/o Ft & 0.34             & 0.31                 & 0.41               & 0.43              \\ 
BART Large MNLI           & 0.32             & 0.33                 & 0.37               & 0.37              \\ 
BART Large CNN            & 0.69             & 0.73                 & 0.72               & 0.86              \\ 
\underline{BART2S}                    & \underline{\textbf{0.80}}             & \underline{\textbf{0.82}}                 & \underline{\textbf{0.87}}               & \underline{\textbf{0.94}}              \\ \hline
\end{tabular}
\caption{Evaluation results on grammar, readability, relevance, and fairness of each system. The underlined line indicates our proposed system.}
\label{tab:human_eval}
\end{table*}

Table \ref{tab:human_eval} shows the performance of the proposed framework and control systems in the human evaluation as described.
Because of the technical limitation, the maximum length of the generated content of all systems is 512 subwords.
With the multi-task learning and novel few-shot tunning procedure, our BART2S Framework achieves state-of-the-art results in all metrics.
Although the results yield positive possibilities, note that all scores are measured in the local scope. 
For example, we can see that BART2S achieve a very high result in terms of fairness but this number only reflects that the evaluators find few fairness concerns in the generated text, which does not guarantee that the combination of them at the document level would remain the same fairness.

\subsection{Discussions}
The BART2S framework proposed in this section illustrates the possibilities in using knowledge of fairness to improve the output quality of generative models.
The framework is designed for terms-of-service generation problem with two components, generator and discriminator.
The generator is trained on multi-task to generate content from a short title.
The discriminator learns about knowledge of fairness, then plays the role of constraining the output quality.
The few-shot tunning framework proposed in this section can be widely applied in problems requiring output quality according to a standard of available knowledge.

\newpage
\section{Summary of Chapter}
The content of this chapter is about using different sources of knowledge to enhance the robustness and explainability of pretrained language models through a technique we call knowledge injection.
Instead of rigidly imposing human knowledge sources or letting deep learning models self-synthesize completely from data, this technique allows the models to update the weights based on the injected knowledge.
To this end, we introduce HYDRA, TRE and BART2S frameworks, which implement different ways to inject the knowledge into the models.

HYDRA is our novel framework using the knowledge of dependency relationships to guide the prediction of the model.
A new transformer layer is added to the existing transformer body and pretrained to imitate the syntactic dependency of interest matrix generated from an existing linguistic parser.
This architecture is interesting in that the pretraining process does not update the weights of the entire model but only the added lightweight layer. HYDRA is named after a mythical creature with many heads and it is also our goal in the future to add more attention heads with different types of knowledge to enhance the power of this framework.

For TREBERT, knowledge of the legal structure of the legal sentence is injected into the model through injection needles into different layers.
TRE in Vietnamese means bamboo, a tree with many sections, similarly, this framework allows knowledge to be injected into different layers of the transformer model.
Through experimentation, we found that injecting knowledge into the layers at the end of the model brings better results.
With legal knowledge injected, this model outperforms the control models in our experiment.

BART2S is designed as a GAN-like architecture containing a generator and a discriminator.
The generator's outputs are regulated by the discriminator, which is pretrained on the knowledge of fairness.
Our novel few-shot tuning framework is proved to enhance the quality of the generated terms-of-service content

With the proposed idea of knowledge injection, instead of letting them deal with raw data on their own, humans can engage these systems more efficiently and responsibly.
Despite the preliminary positive results, there is still work that needs to be done to fully prove the effectiveness of these approaches in particular and knowledge injection for legal attentive models in general.
In future work, we plan to conduct more experiments, ablation studies, and explore more knowledge source and their characteristic to produce more reliable and robust models in legal document processing.

\chapter{Conclusions}
\section{Conclusions}
Using individual studies as basic material, this dissertation proposes an overall solution toward improving attentive neural networks in legal text processing.
Legal text processing is a special subfield of natural language processing. 
The complexity of legal sentences and the high quality requirements of legal AI systems lead to a series of problems for models with attentive neural networks, which have been very successful and have brought many surprises to everyone when applying to problems in the general domain.
Within the scope of a doctoral dissertation, we raise and address four main problems of these models working with legal text.
To achieve this goal, the dissertation is presented in 6 chapters with the first two chapters providing introductory information, the main problems being handled in the next three chapters, and the last chapter is used to discuss, conclude and propose future directions from the initial findings of this dissertation.

The first problem is lacking of data for deep learning models.
Deep learning models in general and attentive models, in particular, are data-hungry computational models.
Unlike classical approaches based on rules, frames or heuristic algorithms, these models need a certain amount of data to achieve good results.
This is also the reason why neural networks were created in the early years of computing, but until recently, with the huge amount of data generated by the Internet, these models have received the attention they deserve.
The problem needs to be solved in order to be able to apply deep learning to legal text processing.
We examine the problem and introduce our solution in Chapter 3 of the dissertation. 
This approach has proven effective in our COLIEE entries in 2020 and 2021.
Besides, in Chapter 3, we also introduce the features of embeddings, model architecture in deep legal processing.
This is an important premise for us to propose solutions to the problems in the following chapters.

The second problem is domain difference for pretrained language models.
Pretrained language models are models that are pretrained on large amounts of text by unsupervised learning (\ie no labeled data is provided).
Tasks designed to pretrain these systems have the purpose of supporting them to model concepts in the language.
Law is a sublanguage with vocabulary and concepts different from everyday language.
Therefore, we base on this feature and the unique resources available in the legal field to obtain better-pretrained language models for problems in this domain.
Specifically, we introduce BERT Law and ParaLaw Nets, pretrained language models in the legal domain, with positive results.
Our contribution for BERT Law is to confirm the effectiveness of a simple solution in domain adaptation (\ie change the data) for the legal domain.
For ParaLaw Nets, we propose novel pretraining tasks to help these models enhance language comprehension through cross-lingual understanding tasks.

The third problem is lengthy content, a constant feature of legal text.
This is a serious problem for pretrained language models.
These models are limited to maximum length during pretraining.
This technical limitation means that these models can only obtain information at the beginning of a paragraph for a long text.
No matter how powerful a model is, it can't make accurate predictions if it doesn't have enough information.
This problem was discovered by us in Chapter 3 and solved in Chapter 4 with a novel architecture named Paraformer.
With the modeling of a paragraph as a set of sentences, we encode individual sentences and then synthesize the signal through the general attention mechanism.
This approach allows us to encode longer legal sentences compared to the standard approach, which brings better performance.

The last problem introduced in this dissertation is uncontrolled learning.
Although proven effective, leaving self-learning models entirely based on data precludes the possibility of human involvement in the process.
As a result, models may interpret data differently than humans do, leading to reduced explainability of the model.
Our proposed solution is to use knowledge sources to guide the learning of these models through different strategies introduced in Chapter 5.
HYDRA is a novel framework using a dependency structure as a linguistic knowledge injected into the attention matrix of the newly appended layer.
TRE is a newly proposed framework with logical structures of law sentences trained on different layers of a Transformer-based model. 
BART2S is a tunable framework that uses knowledge of fairness to constrain the generator to produce high-quality output.

\newpage

\section{Discussions and Future Works}
The studies in this dissertation can contribute to a revolution in a broader scope.
The findings in Chapter 3 are useful references for scientists and industry when approaching legal text processing problems with the use of deep learning models in general and attentive models in particular.
Issues such as data amount, data representation, and model architecture occur not only for legal domain adaptation but for all narrow domains.
Our approach in this dissertation can help them get ideas to develop suitable solutions to their own problems.
The solutions mentioned in Chapter 4 are topical solutions.
Pretrained language models are getting the most attention lately, and using them with possible enhanced solutions can help increase efficiency and save time.
The approaches proposed in Chapter 5 are innovatory ideas. 
Using knowledge as a resource to increase model strength and explainability is a compromise between rigid rules and uncontrolled self-learning.
This is an approach that requires the participation of experts in various fields, and much work remains to be done to achieve the goals set out in this chapter in a wider scope.

With the initial positive results in this dissertation, we plan to pursue more ambitious future works.
First and most straightforward, we intend to examine factors other than those introduced in this dissertation, namely possible issues of deep legal processing, enhancement techniques, and possible knowledge source for guiding deep legal systems. 
Second, we expand our exploration of unresolved issues with deep legal systems.
These problems are numerous and often very challenging.
For example, they could be the ability to recognize and understand relationships between entities that appear in a sentence, the ability of reasoning inside a single document and inter-document, or the ability to detect users' latent intent from a query.
To solve these problems, we need to make more use of data representations (\eg logical, semantic, AMR, etc.) and corresponding knowledge sources.
Last but not least, all scientific works are for only reference until it is actually applied to real life.
We plan to pursue this line of research for a long time, perfecting it, commercializing it, and improving people's quality of life through smart and reliable legal services.

\bibliographystyle{abbrv}
\bibliography{ref.bib}

\renewcommand{\bibname}{Related Publications}

\renewcommand{\bibname}{Other Publications}

\chapter*{Awards}

\begin{itemize}
    \item Research Grants for JAIST Students, 2019.
    \item COLIEE 2019 Winning Group: the best performance on the Legal Case Retrieval Task of the Competition on Legal Information Extraction/Entailment.
    \item COLIEE 2020 Winning Group: the best performance on the Legal Case Entailment Task and Statute Law Textual Entailment Task of the Competition on Legal Information Extraction/Entailment.
    \item COLIEE 2021 Winning Group: the best performance on the Statute Law Question Answering Task of the Competition on Legal Information Extraction/Entailment.
    \item KSE 2021: Runner-up student paper award.
\end{itemize}

\end{document}